\documentclass[final,5p,times,twocolumn]{elsarticle}
\usepackage[font=small,labelfont=bf]{caption}
\usepackage{graphicx,times,amsmath,amssymb,multirow,caption,subcaption,color}
\usepackage[LGR,T1]{fontenc}
\usepackage[latin9]{inputenc}
\usepackage{array,textcomp,stackrel,url,mathtools,enumerate}
\usepackage{multirow} 
\usepackage{multicol} 
\usepackage{mathrsfs, bm}
\usepackage{booktabs,helvet,courier,float,csquotes}
\usepackage{algorithm,algcompatible}
\usepackage{color,soul}
\usepackage[colorinlistoftodos]{todonotes}
\usepackage{blindtext}
\usepackage{csquotes}
\usepackage{longtable}
\usepackage{diagbox}
\usepackage{rotating}
\usepackage{autobreak}
\usepackage{setspace}
\usepackage{amsmath,balance}
\usepackage{graphicx}
\usepackage[normalem]{ulem}
\robustify\uline

\usepackage{tikz}  \usetikzlibrary{matrix,chains,positioning,decorations.pathreplacing,arrows}
\algnewcommand\INPUT{\item[\textbf{Input:}]}%
\algnewcommand\OUTPUT{\item[\textbf{Output:}]}%

\journal{}
\begin{document}
\begin{frontmatter}
\title{Planning-Assisted Context-Sensitive Autonomous Shepherding of Dispersed Robotic Swarms in Obstacle-Cluttered Environments}
\author[lable1]{Jing Liu\corref{cor1}}
\ead{jing.liu5@unsw.edu.au/ liujing2605@gmail.com}
\author[lable1]{Hemant~Singh}
\ead{h.singh@adfa.edu.au}
\author[lable1]{Saber~Elsayed}
\ead{s.elsayed@unsw.edu.au}
\author[lable1]{Robert~Hunjet}
\ead{r.hunjet@adfa.edu.au}
\author[lable1]{Hussein Abbass}
\ead{h.abbass@unsw.edu.au}

\cortext[cor1]{Corresponding author}

\cortext[]{Authors have no known competing financial interests or personal relationships that could have appeared to influence the work reported in this paper.}
\address[lable1]{School of Engineering and Information Technology, University of New South Wales, Canberra ACT, Australia}

\begin{abstract}
Robotic shepherding is a bio-inspired approach to autonomously guiding a swarm of agents towards a desired location. The research area has earned increasing research interest recently due to the efficacy of controlling a large number of agents in a swarm (sheep) using a smaller number of actuators (sheepdogs). However, shepherding a highly dispersed swarm in an obstacle-cluttered environment remains challenging for existing methods. To improve the efficacy of shepherding in complex environments with obstacles and dispersed sheep, this paper proposes a planning-assisted context-sensitive autonomous shepherding framework with collision avoidance abilities. The proposed approach models the swarm shepherding problem as a single Travelling Salesperson Problem (TSP), with two sheepdogs\textquoteright\ modes: no-interaction and interaction. An adaptive switching approach is integrated into the framework to guide real-time path planning for avoiding collisions with static and dynamic obstacles; the latter representing moving sheep swarms. We then propose an overarching hierarchical mission planning system, which is made of three sub-systems: a clustering approach to group and distinguish sheep sub-swarms, an Ant Colony Optimisation algorithm as a TSP solver for determining the optimal herding sequence of the sub-swarms, and an online path planner for calculating optimal paths for both sheepdogs and sheep. The experiments on various environments, both with and without obstacles, objectively demonstrate the effectiveness of the proposed shepherding framework and planning approaches.
\end{abstract}




\end{frontmatter}

\section{Introduction} \label{Intro}
\par As a bio-inspired swarm guidance approach, robotic shepherding seeks to guide a swarm of agents (e.g., sheep flock, crowd) to a goal area by controlling the movement of one or more outside robots (known as sheepdogs or shepherds)~\cite{long2020comprehensive}. Simulating the shepherding behaviour has attracted increasing attention of scholars due to the ability to map the level of abstraction in the shepherding problem to many real-world applications such as crowd control~\cite{lien2009interactive}, precision agriculture~\cite{evered2014investigation}, objects collection~\cite{strombom2018robot}, robotic manipulation~\cite{bat2017shepherding}, and preventing birds from entering an airspace in airports~\cite{paranjape2018robotic}.

\par One of the most challenging issues in shepherding is how to increase success rate while reducing mission\textquoteright s completion time when herding a large number of sheep that are highly dispersed in an  environment with obstacles (more on this challenge in Section~\ref{sec_relatedwork_shepherding}). 

Existing swarm shepherding methods can be roughly classified as rule-based methods~\cite{strombom2014solving}~\cite{singh2019modulation}, learning-based methods~\cite{nguyen2020perceptron}~\cite{zhi2021learning}, and planning-based methods~\cite{chipade2021multiagent}~\cite{song2021herding}. Rule-based algorithms lack the flexibility and adaptability required to manage a wide range of environments~\cite{el2020limits}. While learning-based methods have the potential to address adaptability, they rely heavily on training and require a large amount of data and/or significant computational time for training~\cite{hussein2022autonomous}. Planning-based methods integrate planning approaches~(e.g., optimisation techniques) into rule-based methods to guide sheepdog behaviour. However, current literature is limited to motion (e.g., path) planning algorithms, mostly for a single agent or multi-agents exhibiting self-control only (i.e. they do not need to exercise indirect control over groups). Therefore, existing methods face difficulty addressing the shepherding problem. The problem is compoinded when sheepdogs have limited influence ranges and need to herd a large dispersed flock into several sub-swarms in environments containing obstacles.

\par  Planning is an important research field in robotics and artificial intelligence~\cite{lavalle2006planning}~\cite{zhao2021path}~\cite{muller2022motion}. The field promises to improve the shepherding performance in terms of success rate and completion time~\cite{liu2021mission}. Aiming to address the multiple sub-swarm, obstacle-cluttered environment, and effective shepherding, this paper focuses on planning-based methods. We capitalise on the similarity between multiple sub-swarm shepherding and the Travelling Salesperson Problem~(TSP) to realise the benefits of path planning in obstacle-cluttered environments. TSP is a well-known route planning problem for determining the optimal visiting sequence of a list of cities~\cite{lenstra1975some} and has been extensively studied as described in Section~\ref{sec_relatedWorks_planning}. However, the similarities and application of TSP to shepherding problems has not been studied. Path planning using metaheuristics such as Evolutionary Computation Algorithms (EC)~\cite{elsayed2020path} and Rapidly Exploring Random Tree algorithm~\cite{song2021herding} have shown their promising early results in facilitating swarm shepherding.

\par This paper proposes a planning-assisted swarm shepherding framework by integrating the TSP  with path planning to improve the effectiveness of shepherding, especially for a highly dispersed sheep swarm in an obstacle-cluttered environment. In the proposed shepherding framework, the sheep swarm is firstly divided into sub-swarms to identify the set of virtual `cities'  and then the shepherding problem is transformed into a single TSP to determine the optimal push sequence of sheep sub-swarms. Path planning is integrated with TSP for finding the optimal path for the sheep sub-swarm to be herded towards the next `city' sequentially without collision with obstacles, and for the sheepdog to move towards the driving point of the sheep sub-swarm in real time. A primary difference between a classic TSP and the way it is adopted for shepherding is that the travelling salesperson actuates on itself, while in shepherding, it actuates on a group with unpredictable responses from the members of the groups. To handle this challenge effectively, we needed to combine the offline TSP with on-line heuristics to manage the emerging dynamics from these indirect interactions.

\par We consider two modes for a sheepdog to respond to context-sensitive information. First, a context-sensitive interaction mode where the sheepdog is forcing the sheep sub-swarm to move. Second, a no-interaction mode where the influence of sheepdogs on the sheep swarm is minimal to avoid undesired movements of sheep. An adaptive switching approach is proposed to assist sheepdogs in switching between the two interaction and no-interaction modes based on context during real-time opertions. Subsequently, we present a hierarchical mission planning algorithm, which combines the offline grouping and TSP solver, as well as the online path planner, to solve the optimisation problems involved in the framework. The grouping method divides the sheep swarm by evaluating if there is cohesion forces among the sheep and a well-known optimisation approach, Max-Min Ant System (MMAS)~\cite{stutzle2000max}, is introduced for addressing the TSP. Besides, a two-layer path planner, A*-Post Processing (A*-PP), is presented to optimise the path for both sheepdogs and sheep swarm. 

\par The contributions of this paper include the following:
\begin{itemize}
    \item A planning-assisted swarm shepherding model is proposed to effectively herd the highly dispersed sheep swarm to the goal in  obstacle-cluttered environments.

    \item The formulation of the swarm shepherding problem as a single TSP  to determine the optimal herding sequence of sub-swarms.

    \item  A context-sensitive response model where the sheepdog adaptively switches between two modes of operation during real-time path planning. 
    
    \item A hierarchical mission planning system, consisting of offline grouping and MMAS-based sequencing as well as online path planning, are designed. 

\end{itemize}

\par The remainder of the paper is organised as follows. Section~\ref{Sec_RelatedWork} provides a review of the works related to swarm shepherding and mission planning. Section~\ref{Sec_ProblemDefinition} presents the basic shepherding model while Section~\ref{sec_planning_shepherding} describes the proposed planning-assisted shepherding model. The planning algorithms used in the proposed shepherding model are presented in Section~\ref{sec_missionplanningalgorithms}, followed by the experimental results and analysis Section~\ref{Sec_experiments}. Last is the conclusion in Section~\ref{Sec_conclusion}.

\section{Related works} \label{Sec_RelatedWork}

This section covers an overview of related work to swarm shepherding.

\subsection{Swarm shepherding} \label{sec_relatedwork_shepherding}
 
The success of robotic swarm shepherding relies on the modelling of sheep flocking behaviour and the design of sheepdog control strategies. The rules of BOIDS~\cite{reynolds1987flocks}~\cite{miki2007effective}~\cite{fujioka2018effective} are the most common sheep modelling method, where separation, cohesion and alignment of sheep are considered.  To improve robustness when herding larger flocks, Harrison et al.\cite{harrison2010scalable} viewed the flock as an abstracted deformable shape,  while Hu et al.~\cite{hu2020occlusion}  used adaptive protocols and artificial potential filed methods to model the sheep flocking behaviour.

\par The shepherding field of research has focused more on the design of sheepdog control strategies. As a representative work of rule-based shepherding methods, Str{\"o}mbom et al\textquoteright s shepherding algorithm~\cite{strombom2014solving} laid the foundations for many other shepherding methods, such as the modulation model in~\cite{singh2019modulation} and the Reinforcement Learning (RL) approach in~\cite{go2016reinforcement}. Str{\"o}mbom et al.~\cite{strombom2014solving} (described in section~\ref{Sec_ProblemDefinition}) simulated two typical sheepdog behaviours (collecting the dispersed sheep, and driving the aggregated sheep swarm to a specific location). They found that the mission completion time increases and the success rate decreases as the number of agents in the swarm increases. 

A coordination algorithm was designed in~\cite{hu2020occlusion} to employ multiple robotic sheepdogs to herd two flocks of sheep, which consists of 20 and 30 sheep, respectively. It was observed that the proposed algorithm could not handle the shepherding of a large flock. 
El-Fiqi et al.~\cite{el2020limits} investigated the influence of some key factors (e.g., the density of obstacles and the initial spatial distribution of sheep) on the complexity of shepherding and identified the limitations of reactive shepherding. It was suggested that an increase in the density of obstacles and the sheep\textquoteright s initial level of dispersion in the environment escalate problem complexity and reduce mission success rate. 

\par Learning-based methods have also been studied~\cite{zhi2021learning}~\cite{nguyen2020perceptron}~\cite{nguyen2020continuous}. Go  et al.~\cite{go2016reinforcement} extended Str{\"o}mbom et al.\textquoteright s model by applying RL for learning the sheepdog\textquoteright s behaviour policy. Hussein et al.~\cite{hussein2022autonomous} decomposed the shepherding problem into two sub-problems: learning to push an agent from a location to the destination and selecting whether to collect scattered agents or drive the largest flock to the destination. They aimed to reduce the problem\textquoteright s complexity and proposed a curriculum-based RL to accelerate the learning process. However, the investigation of the swarm shepherding problem with multiple sub-swarms randomly dispersed in the obstacle-cluttered environment is still limited, and an efficient way to address this problem is lacking.

\subsection{Planning approaches} \label{sec_relatedWorks_planning}
Mission planning approaches such as path planning algorithms have been well investigated and applied for mobile robots.  
The planning sub-problems (e.g., path planning, route planning, task assignment) involved in mission planning are defined, and the promising applications for swarm shepherding are discussed in~\cite{liu2021mission}. Long et al.~\cite{long2020comprehensive} also suggested that the sheepdog should take charge of high-level planning, such as path planning and task allocation for completing complex shepherding tasks. 
Some research on the applications of planning approaches for swarm shepherding exist~\cite{chipade2021multiagent}~\cite{song2021herding}. For instance, Lien and Pratt~\cite{lien2009interactive} presented a computer-human interactive motion planning method to address the shepherding problem. They observed that the planner lacks efficiency when the flock separates into several sub-groups. 
To modify the shepherding model in environments with obstacles, Elsayed et al.~\cite{elsayed2020path} presented a 2-stage differential evolution-based path planning algorithm that optimises the path for the sheepdog and sheep. They demonstrated that the path planning algorithm could reduce the time to complete the shepherding task.  

\par TSP is a well-known NP-hard route planning problem that aims to find the route with the optimal cost for a salesperson to visit each city exactly once and returns to the initial city, given a set of cities and the travelling cost between each pair of cities~\cite{lenstra1975some}. TSP is a generalisation of or can be applied to many real-world problems, such as vehicle routing problem (VRP)~\cite{berczi2022efficient}, multi-robot task allocation~\cite{ayari2019acd3gpso}, multi-regional coverage path planning~\cite{xie2022multiregional}, transportation and delivery~\cite{baniasadi2020transformation}. 
Significant research has been conducted on TSP~\cite{khoufi2019survey}~\cite{xu2021precedence}. Some effective approaches for solving the TSP include EC~\cite{mavrovouniotis2016ant}~\cite{ali2020novel} and swarm-intelligence algorithm such as Ant Colony Optimisation (ACO)~\cite{dorigo1992optimization}, which has demonstrated its ability to solve TSP in multiple studies~\cite{stutzle2000max}~\cite{dorigo1997ant}~\cite{xiang2021pairwise}.
Many variants of TSP, such as Multiple TSP (MTSP) and Dynamic TSP (DTSP)~\cite{mavrovouniotis2016ant}, exist. For example, when there are multiple salespersons, the problem is called MTSP and can be further classified as single-depot and multi-depot based on where salespersons depart from~\cite{cheikhrouhou2021comprehensive}. Transformation methods have been used to convert a complex TSP problem to a classic single TSP where general and efficient TSP solvers exist~\cite{oberlin2010today}. Shepherding problems, especially with multiple sub-swarms, share some similarities with TSP. For example, there are some swarm locations (`cities') required to be visited by some agents (sheepdogs/salespersons) in both problems.  However, to the best of our knowledge, TSP has not been applied to the robotic shepherding problem before.

\section{Str{\"o}mbom model} \label{Sec_ProblemDefinition}

Before moving to the proposed approach, we briefly describe the model proposed by Str{\"o}mbom et al. to introduce the terminology associated with shepherding that will be used subsequently. Let the sheep swarm be $\Pi =\{\pi_1,...,\pi_i,...,\pi_N\}$ where $\pi_i$ denotes a sheep agent and $N$ is the number of sheep agents in the swarm.  $B=\{\beta_1,...,\beta_j,...,\beta_M\}$ is the set of sheepdog agents (UGVs) with $M$ sheepdogs denoted as $\beta_j$. The goal position which sheepdogs herd the sheep swarm towards is denoted as $P_G$. The position of $\pi_i$/$\beta_j$ at time step $t$ is denoted as $P_{\pi_i}^t$/$P_{\beta_j}^t$.
As per~\cite{singh2019modulation}~\cite{elsayed2020path}~\cite{el2020limits},  sheep $\pi_i$  total force $F_{\pi_i}^t$ and sheepdog $\beta_j$ total force $F_{\beta_j}^t$ are calculated as Equation~\eqref{Eq_SheepTotalForce} and  Equation~\eqref{Eq_SheepdogTotalForce} respectively.

\begin{equation} \label{Eq_SheepTotalForce}
\begin{split}
    F_{\pi_i}^t=W_{\pi_v}F_{\pi_i}^{t-1}+W_{\pi\Lambda}F_{\pi_i\Lambda_{\pi_i}^t}^{t}+W_{\pi\beta}F_{\pi_i\beta_j}^{t} \\
    + W_{\pi\pi}F_{\pi_i\pi_{i_1}}^{t} +W_{\pi o}F_{\pi_i o}^{t} +W_{e\pi_i}F_{\pi_i\epsilon}^{t}
\end{split}
\end{equation}
\begin{equation} \label{Eq_SheepdogTotalForce}
    F_{\beta_j}^t= F_{\beta_jcd}^t +W_{e\beta_j}F_{\beta_j \epsilon}^{t}
\end{equation}
where each $W$ is the weight of the corresponding force vector. Each force vector is described as follows:
\par For sheep $\pi_i$:
\begin{enumerate}
    \item $F_{\pi_i}^{t-1}$ is the previous total force vector;
    \item $F_{\pi_i\Lambda_{\pi_i}^t}^{t}$  represents the attraction force to its neighbours $\Lambda_{\pi_i}^t$ within the cohesion range $R_\Lambda$;
    \item $F_{\pi_i\beta_j}^{t}$ represents the repulsion force from sheepdog $\beta_j$ if $\pi_i$ is within the influence range of the sheepdog $R_{\pi\beta}$;
    \item $F_{\pi_i\pi_{i_1}}^{t}$ is the repulsion force from other sheep $\pi_{i_1}, i_1\ne i$ within the sheep avoidance radius $R_{\pi\pi}$;
    \item $F_{\pi_i o}^{t}$ is the repulsion force from the obstacles $o$  within the obstacles avoidance radius $R_{\pi o}$;
     \item $F_{\pi_i\epsilon}^{t}$ is the random forces added to sheep $\pi_i$.
\end{enumerate}

\par For sheepdog $\beta_j$:
\begin{enumerate}
    \item  $F_{\beta_jcd}^t$ represents the normalised force vector that makes the sheepdog move to the driving point $P_D^t$ or collection point $P_C^t$;
    \item $F_{\beta_j \epsilon}^{t}$ is the random forces added to Sheepdog $\beta_j$ to help avoid deadlocks.
\end{enumerate}

 To complete the shepherding mission, sheepdog agents switch between driving behaviour and collecting behaviour by evaluating if any sheep is further away from the sheep flock as shown in Algorithm~\ref{alg_reactive_shepherding}. Specifically, if the distance between any sheep and the Global Centre of Mass (GCM) of flock is further than the neighbourhood range $R_n$, the sheepdog moves to the collecting point $P_C^t$, which is located behind the furthest sheep $\pi_f^t$ in the direction of the GCM; otherwise, the sheep are clustered in the flock and  the sheepdog needs to execute a driving behaviour by moving to the driving point $P_D^t$, which is located behind the GCM relative to the final goal $P_G$. $P_D^t$ and $P_C^t$ are calculated as following:
\begin{equation} \label{equa_driving_point_global}
    P_D^t=GCM^t+ (R_n +R_s)\frac{P_G-GCM^t}{||P_G-GCM^t||}
\end{equation}
\begin{equation} \label{equa_collecting_point}
    P_C^t=P_{\pi_f^t}+ R_s \frac{GCM^t-P_{\pi_f}^t}{||GCM^t-P_{\pi_f}^t||}
\end{equation}
\begin{equation}
    R_n=R_{\pi\pi}\sqrt{2N}
\end{equation}
where $R_s$ is the safe operation distance between a sheepdog and a sheep. 

\begin{algorithm}[!th]
\small
\begin{spacing}{0.9}
  \caption{Herding($P_G$, $GCM^t$, $\Pi$)}\label{alg_reactive_shepherding}
  \begin{algorithmic}[1]
    \INPUT $P_G$, $GCM^t$, $\Pi =\{\pi_1,...,\pi_i,...,\pi_N\}$
    \STATE Locate the furthermost sheep $\pi_f$ 
    \IF{the distance between $\pi_f$ and $GCM> R_n$}
    \STATE  Calculate the driving point $P_D^t$ using Equation~\eqref{equa_driving_point_global}
    \ELSE
    \STATE  Calculate the collecting point $P_C^t$ using Equation~\eqref{equa_collecting_point}
    \ENDIF
    \OUTPUT $P_D^t$/$P_C^t$
  \end{algorithmic}
 \end{spacing}

\end{algorithm}

\par  Then sheepdog $\beta_j$ position $P_{\beta_i}^{t+1}$ and sheep $\pi_i$  position $P_{\pi_i}^{t+1}$ are updated according to Equation~\eqref{Eq_SheepdogPosition} and Equation~\eqref{Eq_SheepPosition}, respectively.
\begin{equation} \label{Eq_SheepdogPosition}
    P_{\beta_j}^{t+1}= P_{\beta_j}^{t} + S_{\beta_j}^{t} F_{\beta_j}^t
\end{equation}
\begin{equation} \label{Eq_SheepPosition}
    P_{\pi_i}^{t+1}= P_{\pi_i}^{t} + S_{\pi_i}^{t} F_{\pi_i}^t
\end{equation}  

where $S_{\beta_j}^{t}$ and $S_{\pi_i}^{t}$ represent the moving speed of sheepdog $\beta_j$ and sheep $\pi_i$.

\section{Planning-assisted swarm shepherding framework} \label{sec_planning_shepherding}
As discussed in Section~\ref{Sec_RelatedWork}, existing shepherding models are inefficient when the sheep agents are too dispersed and the density of obstacles is high. To address this issue, this section proposes a planning-assisted swarm shepherding framework to improve shepherding efficacy by integrating a grouping/clustering approach, a TSP solver, and a localised path planning and navigation into a planning-assisted shepherding framework. 

\subsection{Grouping of dispersed sheep in the environment}
Given a highly dispersed sheep swarm $\Pi$ in an environment, the first step in the planning-assisted shepherding framework is to group the dispersed sheep into some sub-swarms and locate the Local Centre of Mass (LCM) of each sub-swarm. The set of sheep sub-swarms is denoted as 
\begin{equation}
    \Phi=\{\phi_1,..., \phi_q,..., \phi_Q \}
\end{equation}
where $Q$ is the number of sub-swarms and the sub-swarm $\phi_q$ subjects to $\bigcup_{q=1}^Q \phi_q=\Pi$,\ $\bigcap_{q=1}^Q \phi_q= \emptyset$ and $\phi_q \neq \emptyset,\ \forall q\in \{1,..., Q\}$. A sheep is assigned to a sub-swarm $\phi_q$ if it is within the cohesion range $R_\Lambda$ from any sheep of this sub-swarm. 
The LCM of $\phi_q$ at time step $t$ is calculated as 
\begin{equation} \label{equa_lcm}
    LCM_q^t= \frac{1}{N_s^q} \sum_{l=1}^{N_s^q}P_{\pi_l^q}^t
\end{equation}
where $N_s^q$ is the number of sheep in the sub-swarm and $P_{\pi_l^q}^t$ is the position of the $l_{th}$ sheep grouped in $\phi_q$ at $t$. The LCM of each $\phi_q$ is regarded as a target location, which the sheepdog should visit.

\subsection{Transforming the swarm shepherding problem to the TSP for task sequencing}

\par After obtaining LCMs  $\{LCM_1^t,..., LCM_q^t,..., LCM_Q^t \}$ of sub-swarms, the swarm shepherding problem can be transformed into a variant of TSP. This section discusses how to transform the single-sheepdog shepherding and bi-sheepdog shepherding problems to a single TSP and presents the mathematical formulation of the shepherding-transformed single TSP. Subsequently, the general TSP solver (presented in Section~\ref{sec_aco_sequencing}) can be employed to find the optimal push sequence of sheep sub-swarms to guide the sheepdog(s)' behaviours.

\subsubsection{Transforming the single-sheepdog shepherding problem}
\begin{figure} [!t]
    \centering
    \subfloat[Swarm initialisation]{
    \includegraphics[width=0.22\textwidth,height=3.8cm]{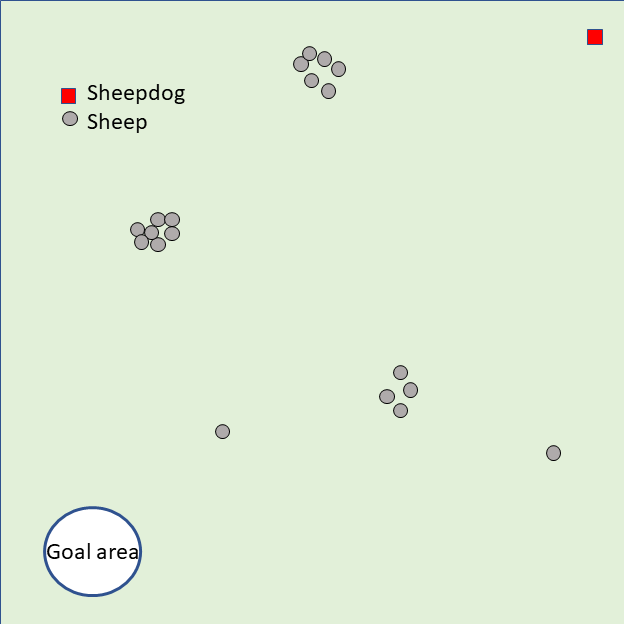}
    \label{fig_swarm_initialisation}
    }
    \subfloat[Single-sheepdog shepherding]{
    \includegraphics[width=0.22\textwidth,height=3.8cm]{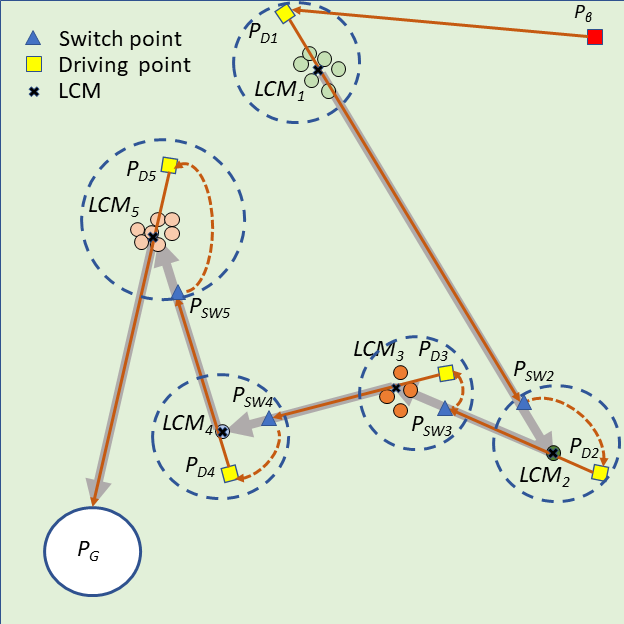}
    \label{fig_routing_illustration}
    }
  \caption{Illustration of a single-sheepdog swarm shepherding problem with multiple sub-swarms}
    \label{fig_swarm_shepherding_illustration}
 \end{figure}

To transform the single-sheepdog shepherding problem, we first describe how the shepherding mission is expected to be completed in our proposed model. 
For illustrative purposes, Fig.~\ref{fig_swarm_initialisation} presents a single-sheepdog swarm shepherding problem with sheep dispersed randomly in the obstacle-free environment and a sheepdog located in the top-right corner. The grouping result is indicated in Fig.~\ref{fig_routing_illustration} with 5 sub-swarms $\{\phi_1, \phi_2, \phi_3, \phi_4, \phi_5\}$ in different colours and the LCMs $\{LCM_1^t, LCM_2^t, LCM_3^t, LCM_4^t, LCM_5^t\}$ are represented as black crosses. Assuming the optimal push sequence of sub-swarms is $[1,2,3,4,5]$, Fig.~\ref{fig_routing_illustration} illustrates how the sheepdog is going to drive the sequenced sheep sub-swarms to reach the goal area. 

\par Similar to the description in Section~\ref{Sec_ProblemDefinition}, the driving point  of each sub-swarm is located behind the sub-swarm in the direction of the next target location, maintaining the distance of $R_n+R_s$ from the LCM of the sub-swarm. 
The driving point $P_{D_q}$ for each sub-swarm $\phi_q$ is represented as a yellow square in Fig.~\ref{fig_routing_illustration}. To control the sheep sub-swarms to move as indicated by the grey thick directed line segments in Fig.~\ref{fig_routing_illustration}, the sheepdog should follow the route represented by 
the red directed line segments and curves. To assist the sheepdog in switching to drive another sub-swarm, switch points $P_{SW_q}^t$ are introduced in the proposed model and are represented as blue triangles in Fig.~\ref{fig_routing_illustration}. Specifically, the sheepdog $\beta$ departs from its initial position $P_\beta^0$ for $P_{D_1}^t$, pushes $\phi_1$ to $LCM_2^t$ by travelling to $P_{SW_2}^t$, then switches to $P_{D_2}^t$ for pushing $\phi_2$, and repeats this process until all the sub-swarms reach $P_G$.

To define a TSP, two vital issues need to be addressed. These are 1) identifying the list of cities and 2) evaluating the travelling cost between each pair of cities. In the single-sheepdog swarm shepherding problem, $P_\beta^0$, $P_G$  and the areas where the sub-swarms $\phi_q$ are located in (blue dashed circles in Fig.~\ref{fig_routing_illustration}) constitute the set of cities. For convenience, let the sheepdog\textquoteright s initial position $P_\beta^0$ be $LCM_0^t$, $\phi_0=\emptyset$ and the final goal $P_G$ be $LCM_{Q+1}^t$. The route\textquoteright s start and end city are fixed to be $LCM_0^t$ and $LCM_{Q+1}^t$ to ensure that the sheepdog departs from $P_\beta^0$ and pushes all the sheep to $P_G$.
The travelling cost between each pair of cities should be evaluated by the cost $C_{qq'}$ for pushing $\phi_q$ from $LCM_q^t$  to $LCM_{q'}^t,\ q,\ q'\in \{0,1,2,...,Q+1 \},\ q\neq q'$. 
However, it is challenging to precisely evaluate $C_{qq'}$ as shepherding is a complex, interactive, dynamic process involving some uncontrollable factors. In this study, we simplify the evaluation of $C_{qq'}$ by calculating it as the distance between $LCM_q^t$  and $LCM_{q'}^t$ for the obstacle-free environment, and the cost of generated path between $LCM_q^t$  and $LCM_{q'}^t$ ($C_{qq'}=C\_Pa(q, q')$ as calculated in Equa~\eqref{equa_path_cost}) for the obstacle-laden environment.


\subsubsection{Transforming the bi-sheepdog shepherding problem}
Similarly, the bi-sheepdog shepherding problem can be regarded as a multiple TSP where multiple sheepdogs depart from their corresponding initial locations $P_{\beta_j}^0$ (depots) to visit each LCM (city) exactly once for collecting the dispersed sub-swarms $ \Phi=\{\phi_1,..., \phi_q,..., \phi_Q \}$ and finally drive them to reach the goal location $P_G$ (terminal). The sheepdogs are not required to return to their initial locations. In this section, we further convert the shepherding-transformed multiple TSP to a single TSP so that the general single TSP solver can be employed to address the problem. 

\par
\begin{figure} [!t]
    \centering
    \subfloat[A solution of the STSP]{
    \includegraphics[width=0.22\textwidth,height=3.8cm]{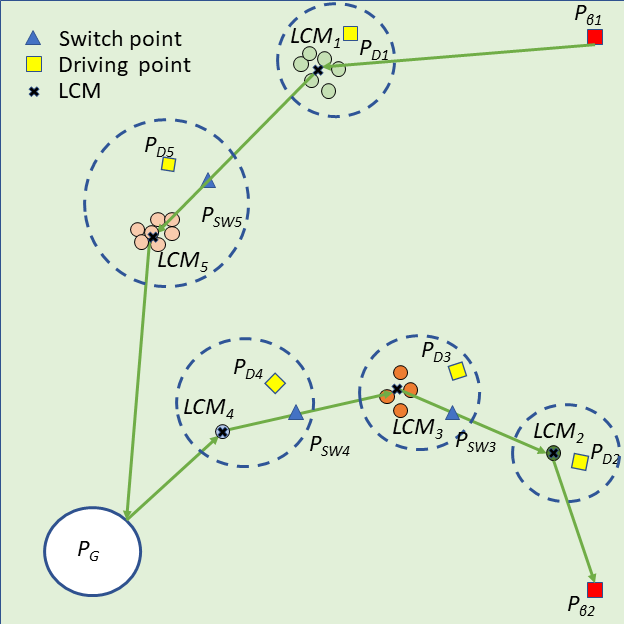}
    \label{fig_multidog_tsp}
    }
    \subfloat[Multi-sheepdog shepherding]{
    \includegraphics[width=0.22\textwidth,height=3.8cm]{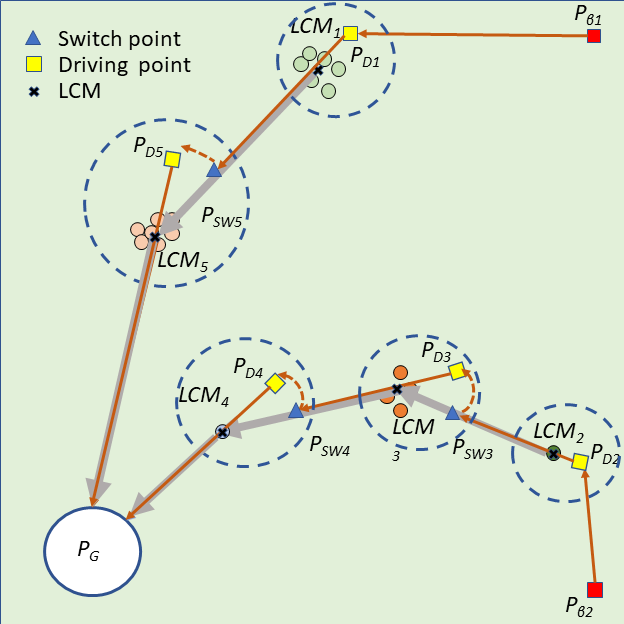}
    \label{fig_multidog}
    }
    
  \caption{Illustration of a multi-sheepdog swarm shepherding problem with scattered sub-swarms}
    \label{fig_multidog-swarm_shepherding_illustration}
 \end{figure}

 To solve the bi-sheepdog shepherding problem as a single TSP (STSP), we regard the initial position of a sheepdog $P_{\beta_1}^0$ as $LCM_0^t$, the goal location $P_G$ as $LCM_{Q+1}^t$ and another sheepdog\textquoteright s initial position $P_{\beta_2}^0$ as $LCM_{Q+2}^t$. $\{LCM_0^t,..., LCM_q^t,..., LCM_{Q+2}^t \}$ are the set of the cities' locations. The start and the end city of the route are fixed to be $LCM_0^t$ and $LCM_{Q+2}^t$. Then a solution of the STSP can be converted to the solution of MTSP by splitting it into two lists at $LCM_{Q+1}^t$ and reversing the order of the latter list. In this way, the first city of each list is the initial position of a sheepdog ($P_{\beta_1}^0$ or $P_{\beta_2}^0$) and the end city is the goal ($P_G$). Other cities on the list are the sub-swarms to be driven by the corresponding sheepdog located at the start of the list. Fig.~\ref{fig_multidog_tsp} shows a solution of the STSP, and Fig.~\ref{fig_multidog} illustrates the transformed solution of MTSP and the shepherding process guided by the MTSP solution.

\subsubsection{Mathematical formulation of the TSP}
Based on the abovementioned discussion, the solution of the shepherding-transformed TSP is formulated as follows:  
\begin{equation}
\psi_{qq'},\ q,\ q'\in \{0,1,2,...,Q+M \},\ q\neq q'
\end{equation}
where $q,\ q'$ are the indexes of LCMs. If the sheepdog pushes $\phi_q$ from $LCM_q^t$ to $LCM_{q'}^t$, $\psi_{qq'}=1$; otherwise, $\psi^u_{qq'}=0$. $M$ is the number of sheepdogs. $M$ is limited to $\{1, 2\}$ here. 

The optimisation objective of the TSP is:  \\
\par \textit{Minimize:}
\begin{flalign} \label{Equa_sequencing_obj}
\qquad & F =\sum_{q=0}^{Q+M} \sum_{q'=0}^{Q+M} C_{qq'}\cdot \psi_{qq'}  &&
\end{flalign}

\textit{Subject to: }
\begin{flalign}
\label{Equa_sequencing_c1}
\qquad  & \sum_{q=0, q\neq q'}^{Q+M} \psi^u_{qq'} =1, \forall q' \in \{1,2,...Q+M \} && \\
\label{Equa_sequencing_c2}
\qquad  & \sum_{q'=0, q'\neq q}^{Q+M} \psi^u_{qq'}=1, \forall q \in \{0,1,...Q+M-1\} && \\
\label{Equa_sequencing_c3}
\qquad & \sum_{q=1}^{Q+M} \psi^u_{q0}=0  && \\
\label{Equa_sequencing_c4}
\qquad & \sum_{q'=0}^{Q+M-1} \psi^u_{(Q+M)q'}=0 && 
\end{flalign}
Here, $F$ is the total cost and $C_{qq'}$ is the cost to push $\phi_q$ from $LCM_q^t$  to $LCM_{q'}^t$. Constraints~(\ref{Equa_sequencing_c1}) and (\ref{Equa_sequencing_c2}) ensure that each target location is visited exactly once. Constraints~(\ref{Equa_sequencing_c3}) and (\ref{Equa_sequencing_c4}) ensure that the sheepdog $\beta_j$ departs from $P_{\beta_j}$  and finally reaches $P_G$.
 
\subsection{Path planning for sheepdog(s) and sheep swarm} \label{sec_pathplanning}

Given the sequenced sub-swarms $ \{\phi_1',..., \phi_q',..., \phi_Q' \}$ and the corresponding LCMs $\{LCM_1',..., LCM_q',..., LCM_Q' \}$, the mission of the sheepdog can be regarded as a set of sequential sub-tasks, i.e., pushing $\phi_q'$ from $LCM_q'$ to $LCM_{q+1}'$, $\forall~q\in \{1,...,Q\}$.  Path planning is crucial for both sheepdog(s) and sheep swarm to reduce detours and mission completion time. Next we present the mathematical formulation of path planning and discuss how to integrate the path planning algorithm into shepherding based on a proposed classification of sheepdog moving mode.

\subsubsection{Mathematical formulation of path planning}
In this paper, the path is defined as a sequence of way-points that can be connected as a set
of path segments. Denoting the start and goal points as $W_0$ and $W_{D+1}$, respectively, the solution of path planning between the two points could be represented as:
\begin{equation}
    Path(W_0, W_{D+1})=\{W_0, W_1,...,W_d,..., W_{D+1}\}
\end{equation}
In the obstacle-cluttered environment, collision avoidance is a hard constraint which means that a path is infeasible once it intersects, i.e. collides, with any obstacle in the environment.

Referring to~\cite{roberge2012comparison}~\cite{yu2018aco}, the cost evaluation function of a feasible path, which is also the optimisation objective of path planning, is defined as:
\begin{equation} \label{equa_path_cost}
   C\_{Pa}(W_0, W_{D+1})=\alpha_1 \cdot C\_L(D)+
   \alpha_2 \cdot C\_Th(D)
\end{equation}
where $\alpha_1$ and $\alpha_2$ are the weights of costs.
\par $C\_L$ is the path length cost and is calculated as: 
\begin{equation}
    C\_L(D)=\sum_{d=0}^{D} \vert\vert  W_d-W_{d+1} \vert\vert  
\end{equation}

\par $C\_Th$ is the threat cost evaluating the unwanted disturbance of sheepdogs on sheep and is calculated as:
\begin{equation}
    C\_Th(D)=\sum_{d=0}^D Threat_d 
\end{equation}
$Threat_d=1$ if the path segment from $W_d$ to $W_{d+1}$ collides with the threat area which is  defined as a set of circles with the centre points $P^t_{\pi_i}, \forall i\in \{1,...N \}$ and the radius $R_{th}$. $R_{th}$ is the threat range, representing the distance that the sheepdog should keep from the sheep to avoid the unwanted influence. A large $R_{th}$ will increase the path length cost of the sheepdog to reach the target point while a small $R_{th}$ might disturb the sheep and cause unexpected movements.  

\subsubsection{Path planning for shepherding} \label{sec_pathplanning_shepherding}
To integrate path planning into context-sensitive shepherding, we design a two-mode sheepdog operations, a no-interaction and an interaction modes based on whether the influence of sheepdogs on sheep swarm $C\_Th$ should be minimised or not. 
\par \textbf{No-interaction mode:} The no-interaction mode is usually triggered when the sheepdog departs from its initial location $P_{\beta_j}^0$ for the driving point of the sub-swarm $\phi_1'$, or when the sheepdog just finished a sub-task of pushing $\phi_{q-1}'$ and switches to the next sub-task by moving towards the driving point of the next sub-swarm $\phi_{q}'$. During the no-interaction mode, the influence of sheepdogs on the sheep swarm should be minimised to avoid unwanted movements of the sheep swarm. Sheep are considered obstacles that should be avoided to avoid disturbing the flocks while the sheepdog is positioning itself for a driving position. The path planning algorithm, A*-PP (presented in Section~\ref{sec_A*PP}), is used to find the optimal path $Path(P_{\beta_j}^t, P_{D_q}^t)$, which is obstacle-free and has the lowest cost, for the sheepdog $\beta_j$ to follow from its current location $P_{\beta_j}^t$ to the driving point $P_{D_q}^t$ of the target sub-swarm. The path cost $C\_{Pa}(P_{\beta_j}^t, P_{D_q}^t)$ is evaluated according to Equation~\eqref{equa_path_cost} by setting $\alpha_1, \alpha_2$ as positive numbers.

\par \textbf{Interaction mode:} The interaction mode is activated when the sheepdog is driving the sheep sub-swarm $\phi_q'$ from $LCM_q^{'t}$ towards a sub-goal $P_{SG}^t$. During this phase, the sheepdog continues to influence the sheep sub-swarm by witching between collecting and driving behaviours as it evaluates the furthest distance of the sheep to $LCM_q^{'t}$. 
The path $Path(P_{\beta_j}^t, P_{D_q}^t/P_{C_q}^t)$ from the sheepdog\textquoteright s current position $P_{\beta_j}^t$ to the driving/collecting point $P_{D_q}^t$/$P_{C_q}^t$ of the target sub-swarm is also optimised using A*-PP. Different from the no-interaction mode, only the path length $C_L$ is considered for the path cost evaluation in the interaction mode. Therefore, $ \alpha_2$ in Equation~\eqref{equa_path_cost} should be set to 0. 

\par The driving point $P_{D_q}^t$ is calculated as follows:
\begin{equation} \label{equa_driving_point}
    P_{D_q}^t=LCM_q^{'t} + (R_n +R_s)\frac{P_{SG}^t-LCM_q^{'t}}{||P_{SG}^t-LCM_q^{'t}||}
\end{equation}
where $LCM_q^{'t}$ is the LCM of the sheep sub-swarm $\phi_q'$ that the sheepdog will be driving or is currently driving; $P_{SG}^t$ is the sub-goal that the sheepdog is driving $\phi_q'$ towards. $P_{SG}^t$ is set as the waypoint in the optimised path $Path(LCM_q^{'t}, LCM_{q+1}^{'t})$ of $\phi_q'$ from $LCM_q^{'t}$ to $LCM_{q+1}^{'t}$, which is obtained by A*-PP as well. In this way, the sheepdog will push $\phi_q'$ to move towards the optimal path so as to reduce the detours of both sheepdogs and sheep sub-swarms. 

\subsection{Planning-assisted shepherding framework} \label{sec_framework}
Based on the discussion above, the overall planning-assisted shepherding framework is presented in Algorithm~\ref{alg_shepherding}, where $Mode=0$ represents the no-interaction mode while $Mode=1$ represents the interaction mode. Before the real-time shepherding, the offline planner obtains the grouping and sequencing results~(lines 2 and 3). Then the sheepdog starts with the no-interaction mode for reaching the driving point of $\phi_1'$. 
 \begin{algorithm}[!ht]
 \small
  \begin{spacing}{0.9}
  \caption{Planning-assisted shepherding model}\label{alg_shepherding}
  \begin{algorithmic}[1]
    \INPUT $\Pi =\{\pi_1,...,\pi_i,...,\pi_N\}$, $\beta$,  $P_G$
    \STATE \textbf{Initialise}: $q=1, t=0, Mode=0$, $T$
    \STATE Get $\Phi=\{\phi_1,..., \phi_q,..., \phi_Q \}$ via Algorithm~\ref{alg_grouping}
    \STATE Get sequenced sub-swarms $ \Phi'=\{\phi_0',..., \phi_q',..., \phi_Q' \}$  and corresponding LCMs $\{LCM_0',..., LCM_q',..., LCM_{Q+1}' \}$ using MMAS
    \WHILE{$t<T$ \&\& sheep have not all reached $LCM_{Q+1}'$}
         \STATE $t=t+1$
        \IF{$\phi_q'$ encounters $\phi_{q+1}'$ }
            \STATE $\phi_{q+1}'=\phi_{q+1}' \cup  \phi_q'$;
            \STATE $q=q+1$; $Mode= 0$;
        \ENDIF
        \STATE Update LCMs;
        \IF{$Mode=0$ || $mod(t,10)=1$}
        \STATE Calculate the optimal path of $\phi_q'$:
        $Path_{\phi'_q}=\{LCM_q', W_1, W_2,...,W_D,  LCM_{q+1}'\}$
        \STATE Let $W_1$ be the sub-goal $P_{SG}^t$
        \ELSE
        \STATE Locate the nearest waypoint on $Path_{\phi'_q}$ ahead of $\beta$ as the sub-goal  $P_{SG}^t$
        \ENDIF
        \IF{$Mode= 0$}
            \STATE Calculate $P_{D_q}^t$ based on Equation~\eqref{equa_driving_point}
            \STATE Optimise $Path_{\beta_j}(P_{\beta_j}^t, P_{D_q}^t)$ as per no-interaction mode
        \ELSE
            \STATE Calculate $P_{D_q}^t$ or $P_{C_q}^t$ based on Algorithm~\ref{alg_reactive_shepherding}:
        $ P_{D_q}^t/P_{C_q}^t$= Herding($P_{SG}^t$, $LCM_q^{'t}$, $\phi_q'$)
            \STATE Optimise $Path_{\beta_j}(P_{\beta_j}^t, P_{D_q}^t/P_{C_q}^t)$ as per interaction mode
        \ENDIF
        \STATE Sheepdog moves following $Path_{\beta_j}$ with the limitation of maximum speed
        \STATE Update the sheep position according to Equation~\eqref{Eq_SheepPosition}
        \IF{$Mode=0$ \& $\beta$ reaches $P_{D_q}^t$}
        \STATE $Mode=1$ 
        \ENDIF
    \ENDWHILE
    \OUTPUT $t$
  \end{algorithmic}
  \end{spacing}
\end{algorithm}

\par
\begin{figure*}[!ht]
    \centering
    \includegraphics[width=0.95\textwidth]{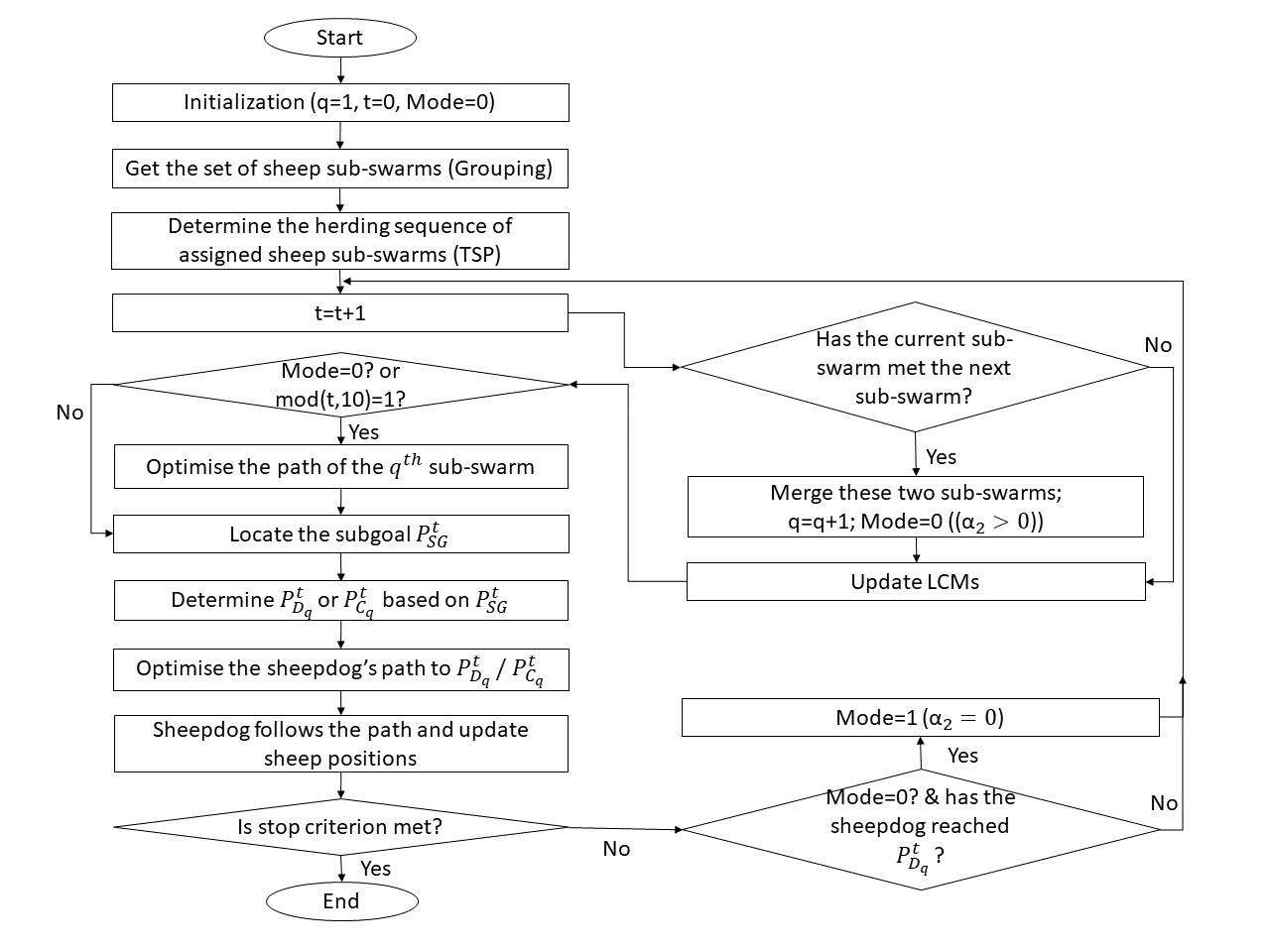}
    \caption{The flowchart of the planning-assisted swarm shepherding model}
    \label{fig_flowchart}
\end{figure*}

 During the shepherding process, the sheepdog switches between the no-interaction and the interaction mode based on context and to complete the set of sequenced sub-tasks, i.e., pushing the sheep sub-swarms $\phi_q'$ to $LCM_{q+1}'$, $\forall~q\in \{1,...,Q\}$. An adaptive switching approach is designed based on the real-time evaluation of the shepherding progress and is integrated into the shepherding framework. As presented in line 5-9 of Algorithm~\ref{alg_shepherding}, the switching from the interaction mode to the no-interaction mode happens when  $\phi_q'$ encounters $\phi_{q+1}'$, indicating that the sheepdog just successfully pushed $\phi_q'$ to $LCM_{q+1}'$ to merge with $\phi_{q+1}'$ and is in preparation for the next sub-task by moving to the driving point of $\phi_{q+1}'$. Then once the driving point of $\phi_{q+1}'$ is reached in the no-interaction mode, the sheepdog switches to the interaction mode~(line 27-29), which means that the sheepdog starts pushing the sub-swarm $\phi_{q+1}'$. This process continues until all the sheep reach the goal area $LCM_{Q+1}'$ or the pre-defined maximum time steps $T$ is reached.  Figure~\ref{fig_flowchart} shows the corresponding flowchart.

\section{Hierarchical mission planning for shepherding} \label{sec_missionplanningalgorithms}
To assist the proposed swarm shepherding framework, a  hierarchical mission planning system is proposed in this section by combining the approach for grouping, MMAS for TSP, and A*-PP for online path planning.

\subsection{Grouping and MMAS: offline task planner} \label{sec_aco_sequencing}
Given the sheep swarm $\Pi$, a cohesion range-based grouping method as presented in Algorithm~\ref{alg_grouping} is used to obtain the set of sheep sub-swarms and calculate the LCMs. Then MMAS~\cite{stutzle2000max}, a well-known ACO algorithm, is introduced to address the shepherding-transformed TSP for getting the optimal push sequence of sub-swarms due to its outstanding performance in addressing TSP. 
ACO is inspired from the real ant colonies\textquoteright  \  foraging behaviour. During the foraging process, ants deposited pheromone trails on the return routes if they find food sources. It enables other ants to find optimal paths to food sources by getting information from pheromone trails. 
\begin{algorithm}[!ht]
\small
\begin{spacing}{0.9}
  \caption{Grouping of dispersed sheep}\label{alg_grouping}
  \begin{algorithmic}[1]
    \INPUT a sheep swarm $\Pi =\{\pi_1,...,\pi_i,...,\pi_N\}$ 
    \STATE \textbf{Initialise}: $Q=0$, $\phi_q=Null$
    \STATE Calculate the distance $D_{\pi_1 \pi}$ between $\pi_1$ and all other sheep, and sort $D_{\pi_1 \pi}$ in ascending order to get the sorted index $Idx$ of sheep
    \FOR{i=1:N}
    \STATE $k=Idx(i)$
    \STATE Find the neighborhood sheep $\Lambda_{\pi_k}$ of $\pi_k$ within $R_\Lambda$
    \IF{$\Lambda_{\pi_k}\neq \emptyset$ \& any of $\Lambda_{\pi_k}$ or $\pi_k$ belongs to an exiting sub-swarm $\phi_q$} 
    \STATE $\phi_q=\phi_q \cup \pi_k \cup \Lambda_{\pi_k}^t $
    \ELSE
    \STATE $Q=Q+1$; $\phi_Q= \pi_k \cup \Lambda_{\pi_k}$
    \ENDIF
    \ENDFOR
    \STATE Calculate the LCM of each $\phi_q$ as Equation~\eqref{equa_lcm}
    \OUTPUT the set of sheep sub-swarms $ \Phi=\{\phi_1,..., \phi_q,..., \phi_Q \}$ and the corresponding LCMs $\{LCM_1^t,..., LCM_q^t,..., LCM_Q^t \}$
  \end{algorithmic}
\end{spacing}
\end{algorithm}
\par The core components of MMAS are solution construction and pheromone update. To find the optimal visiting sequence of sheep sub-swarms, the solution in MMAS is constructed by selecting sub-goals (regarded as a solution component $c_q^{q'}$) one by one based on pheromone values $\tau_{qq'}$ and the values $\eta(c_q^{q'})$ of edges to form the complete solution, which indicates the visiting sequencing. Pheromone values $\tau_{qq'}$ are updated every generation based on the quality of constructed solutions and the evaporation of existing pheromones. The values of $\eta(c_q^{q'})$ are defined as $\eta(c_q^{q'})=1/C_{qq'}$ where $C_{qq'}$ is the travelling cost from $LCM_q^t$ to $LCM_q'^t$. In MMAS, only the best ant is used to update the pheromone and the pheromone values are limited in the predefined ranges $[\tau_{min}, \tau_{max}]$.
The implementation of MMAS for finding the optimal push sequence of sub-swarms is described as follows:
\par  \textbf{Step 1:}  Initialise the parameters of MMAS, including the ant colony size $N_s$, two parameters $\alpha$ and $\gamma$, evaporation rate $\rho$, pheromone values of each edge $\tau_{qq'}=\tau_{max}$, the values of each edge $\eta(c_q^{q'})=1/C_{qq'}$,  the best solution $s^*=$ NULL;
\par \textbf{Step 2:}  Construct new ant solutions:
\par ${\rm{ }}$  \textbf{Step 2.1:} Start with a partial solution $s_p=LCM_0^t$;
\par ${\rm{ }}$  \textbf{Step 2.2:} For each $LCM_{q'}^t, q'\in \{1, 2, ...,Q \}$, calculate the probability $p(c_q^{q'}\vert s_p)$ of moving from the current location $LCM_q^t$ to $LCM_{q'}^t$ based on the following Equation;
\begin{equation} \label{equa_prob_aco_discrete}
    p(c_q^{q'}\vert s_p)=\frac{\tau^\alpha _{qq'}\cdot [\eta(c_q^{q'})]^\gamma}{\sum_{c_q^l\in N(s_p)}\tau^\alpha _{ql}\cdot [\eta(c_q^l)]^\gamma},\ \forall c_q^{q'}\in N(s_p)
\end{equation}
where $N(s_p)$ is a set of available solution components for the current partial solution $s_p$;
\par ${\rm{ }}$ \textbf{Step 2.3:} Select the next solution component based on the probability $p(c_q^{q'}\vert s_p)$ and add the selected $LCM_{q'}^t$ to the current partial solution $s_p$;
\par ${\rm{ }}$ \textbf{Step 2.4:} If all $LCM_{q'}^t, q'\in \{1, 2, ...,Q \}$ are visited, add $LCM_{Q+1}^t$ to $s_p$ and go to Step 2.5; otherwise, go to Step 2.2;
\par ${\rm{ }}$ \textbf{Step 2.5:} If $N$ new solutions are constructed, go to Step 3; otherwise, go to Step 2.1;
\par \textbf{Step 3:} Calculate the cost $F(s)$ and record the best solution $s*$ found with the lowest cost;
\par \textbf{Step 4:}  Update the pheromone values according to:
\begin{flalign}
\qquad \tau_{qq'}& =(1-\rho)\tau_{qq'}+ \Delta\tau_{qq'}^{best} & \\ 
\qquad \tau_{qq'}& =min(max(\tau_{qq'}, \tau_{min}), \tau_{max}) & 
\end{flalign}
where $\Delta\tau_{qq'}^{best}=1/F(s*)$; 
\par \textbf{Step 5:} If the termination condition is met, output the best solution which represents the optimal travelling sequence; otherwise, go to Step 2.

\subsection{A*-PP: online path planner} \label{sec_A*PP}

\par As discussed in Sections~\ref{sec_pathplanning} and~\ref{sec_framework}, path planning is crucial for reducing detours of both sheepdogs and sheep swarm, and is invoked during the online shepherding process. This section presents a two-layer path planning algorithm A*-PP, where the first layer, A*, finds the path with optimal cost, and the second layer, post-processing, eliminates the redundant waypoints in the path. 

\par A*~\cite{hart1968formal} is a well-known node-based path search algorithm that searches in a landscape represented by graphs. A* starts from the specific start node of the search graph and expands the nodes on candidate paths by adding one node at a time until it reaches the goal node in the graph. To decide which node on the candidate paths to be extended next, A* employs an evaluation function $f(n)$ which can be calculated as Equation~\eqref{equa_astar_cost} to estimate the cost of the path going through node $n$.
\begin{equation} \label{equa_astar_cost}
    f(n)=g(n)+h(n)
\end{equation}
where $g(n)$ is the cost of the optimal path from the start node to the current node $n$ and $h(n)$ is the heuristic function for estimating the cost from the current node $n$ to the goal node. 
In this paper, $g(n)$ is calculated as following:
\begin{equation}
    g(n)=C\_L(n)+C\_Th(n),
\end{equation}
where $C\_L(n)$ and $C\_Th(n)$ are the length cost and threat cost. $h(n)$ is calculated as the straight line distance from the current node $n$ to the goal node, which is permissible to guarantee A* returns the optimal path. 

\par The pseudo-code of A* is given in Algorithm \ref{alg_A*}. $Open$ is the set of nodes that can be considered for expansion. $Closed$ is the set of nodes that have been expanded, which makes sure that each node can be travelled at most once. $c(n,m)$ is the cost from the node $n$ to node $m$. $Parent(m)$ is to record the path with the lowest cost. A* starts the search from the initial point $W_{init}$. At each iteration of the main loop, A* selects the node $n$ with the lowest $f(n)$ from $Open$ and removes it from $Open$ to $Closed$. Then A* checks the neighbours of $n$ to insert feasible neighbour nodes $m$ into $Open$ if $m$ is not in $Open$, or update $f(m)$ if $m$ is already in $Open$ and $g(m)+h(m)$ is better than the old $f(m)$. The loop continues until the node with the lowest $f(n)$ is the goal point $W_{goal}$ or $Open$is empty, meaning no feasible path exists. 

\begin{algorithm}[!ht]
\begin{spacing}{0.9}
  \caption{The pseudo code of A*} \label{alg_A*}
  \begin{algorithmic}[1]
  \INPUT $g(W_{init})=0;\ Open= \emptyset;\ Closed= \emptyset;\ Parent(W_{init})=W_{init}$ 
  \STATE Inset $W_{init}$ into $Open$ with $g(W_{init})+h(W_{init})$;
  \WHILE{$Open \neq \emptyset $}
  \STATE Select the node $n$ in $Open$ with the lowest value of $f(n)$ ;
  \IF{$n=W_{goal}$} 
  \STATE Extract $Path$ from $Parent$
  \STATE \textbf{Return} $Path$;
  \ENDIF
  \STATE Remove the node $n$ from $Open$;
  \STATE Add the node $n$ to $Closed$;
  \FOR{each neighbour $m$ of $n$}
  \IF{$i\notin Closed$}
  \IF{$i\notin Open$}
  \STATE $g(m)=\infty$;
  \STATE $Parent(m)=NULL;$
  \ENDIF
  \IF{$g(n)+c(n,m)<g(m)$}
  \STATE $g(m)=g(n)+c(n,m)$;
  \STATE $Parent(m)=n$;
  \STATE Inset $m$ into $Open$ or update $f(m)$ in $Open$ with $g(m)+h(m)$;
  \ENDIF
  \ENDIF
  \ENDFOR
  \ENDWHILE
  \OUTPUT $Path$
  \end{algorithmic}
\end{spacing}
\end{algorithm}

\par However, the original path obtained by A* usually contains many waypoints and the sub-path between two waypoints might be taking an unnecessary detour in some cases where a straight line can connect these two waypoints with no obstacle collision. Therefore, a path post-processing method, line of sight path pruning~\cite{yang2011anytime}, is introduced to remove some redundant waypoints on the path to further reduce the path cost. The pseudo-code of the path post-processing is presented in Algorithm~\ref{alg_post_proces}.  The core of this process is to replace the original sub-path between two waypoints with a straight line if the straight line does not collide with any obstacles, meaning that the waypoints on the original sub-path, except the start point and end point, will be removed.

\begin{algorithm}[!ht]
\begin{spacing}{0.9}
  \caption{The pseudo code of path post-processing} \label{alg_post_proces}
  \begin{algorithmic}[1]
  \INPUT $Path$, $N_{nodes}=size(Path,  1)$
  \WHILE{$i<=N_{nodes}$}
  \FOR{$j=2:N_{nodes}-1$}
  \STATE Check the collision between the $i^{th}$ node and the $(i+j)^{th}$ node 
  \IF{No collision exists}
  \IF{$j>=N_{nodes}-1$}
  \STATE Add the last node on $Path$ to $Processed\_path$ and \textbf{Break}
  \ELSE
  \STATE \textbf{Continue}
  \ENDIF
  \ENDIF
  \STATE Add the $(i+j-1)^{th}$ node to $Processed\_path$ and \textbf{Break}
  \ENDFOR
  \STATE $i=i+j-1$
  \ENDWHILE
  \STATE Add the last node on $Path$ to $Processed\_path$ if it is not 
  \OUTPUT $Processed\_path$
\end{algorithmic}
\end{spacing}
\end{algorithm}

\section{Numerical Experiments} \label{Sec_experiments}
\subsection{Experimental setting}
\par
\begin{figure*} [!ht]
    \centering
    \subfloat[Case1 (50*50, $N=20$)]{
    \includegraphics[width=0.23\textwidth,height=3.5cm]{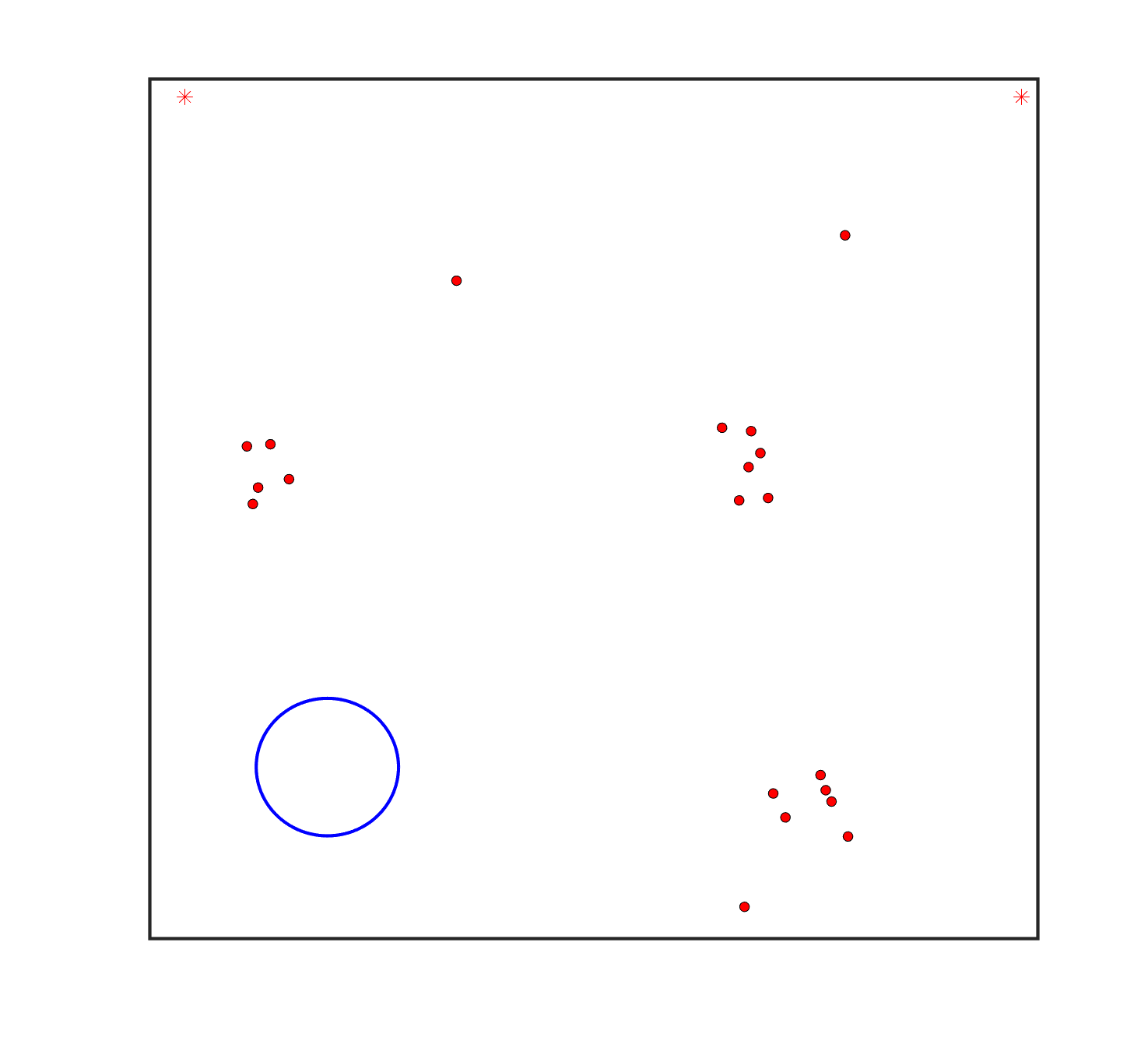}
    \label{Case1_Initailisation}
    }
    \subfloat[Case2 (100*100, $N=20$)]{
    \includegraphics[width=0.23\textwidth,height=3.5cm]{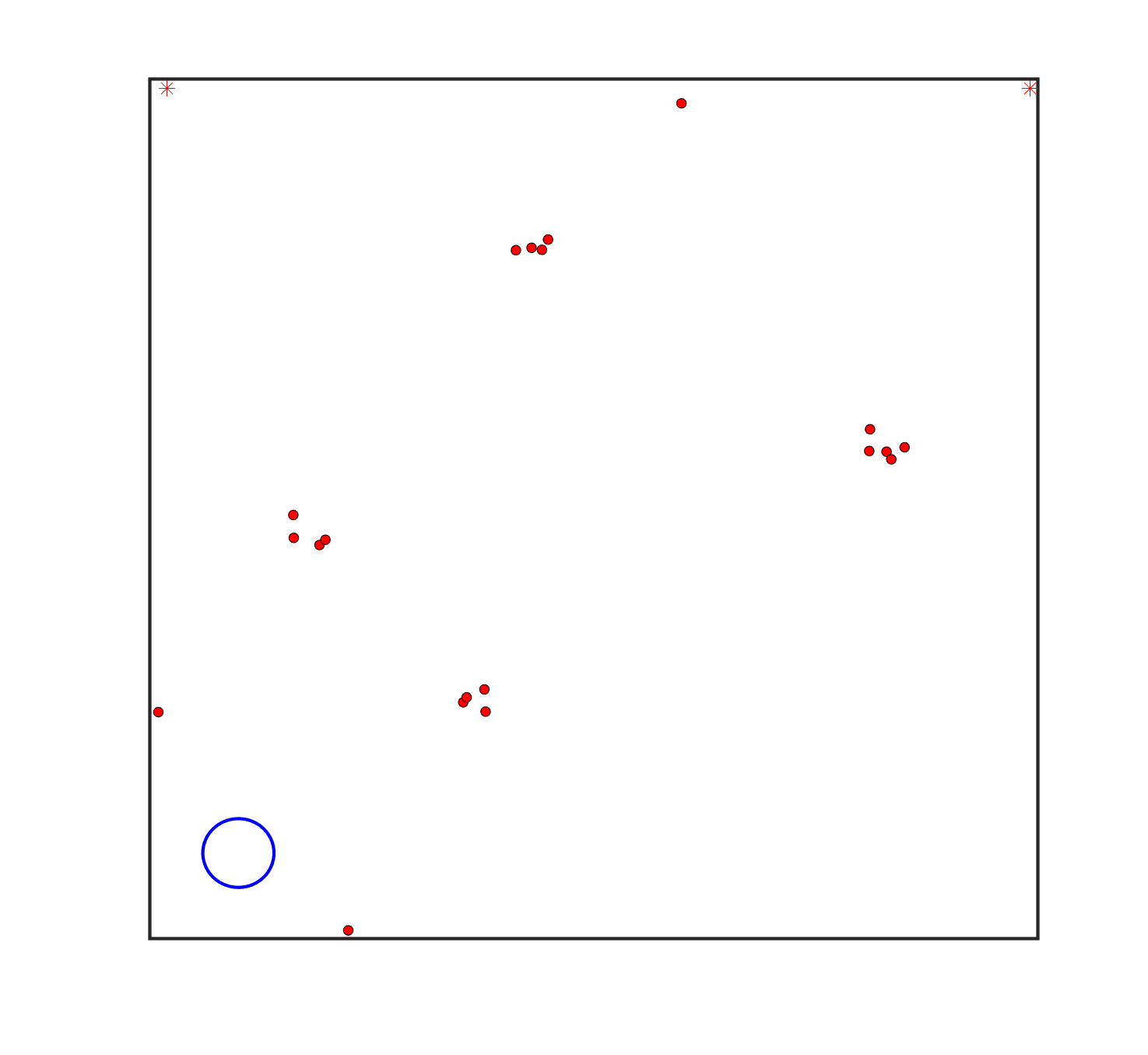}
    \label{Case2_Initailisation}
    }
     \subfloat[Case3 (100*100, $N=50$)]{
    \includegraphics[width=0.23\textwidth,height=3.5cm]{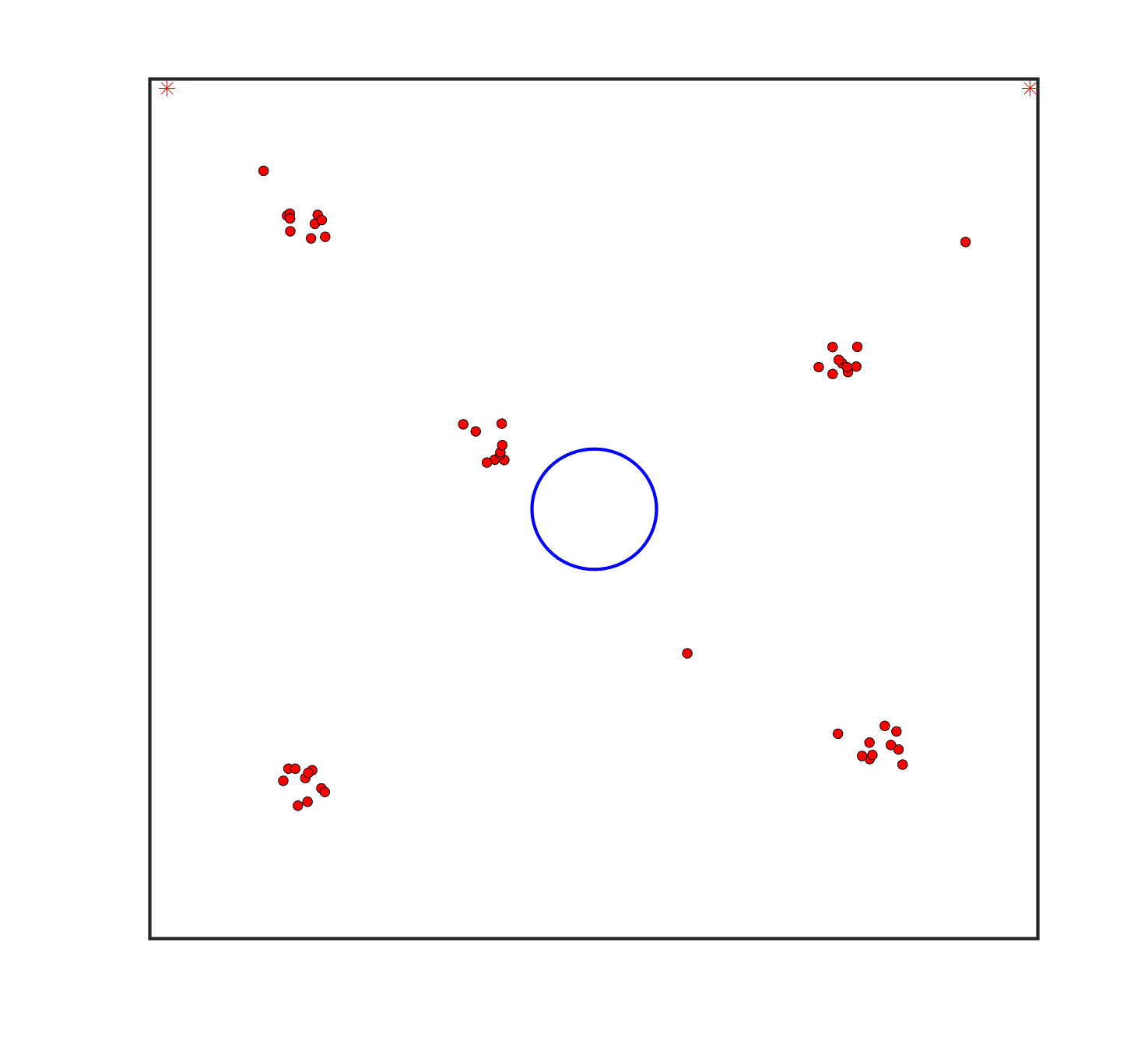}
    \label{Case3_Initailisation}
    }
   \subfloat[Case4 (100*100, $N=50$)]{
    \includegraphics[width=0.23\textwidth,height=3.5cm]{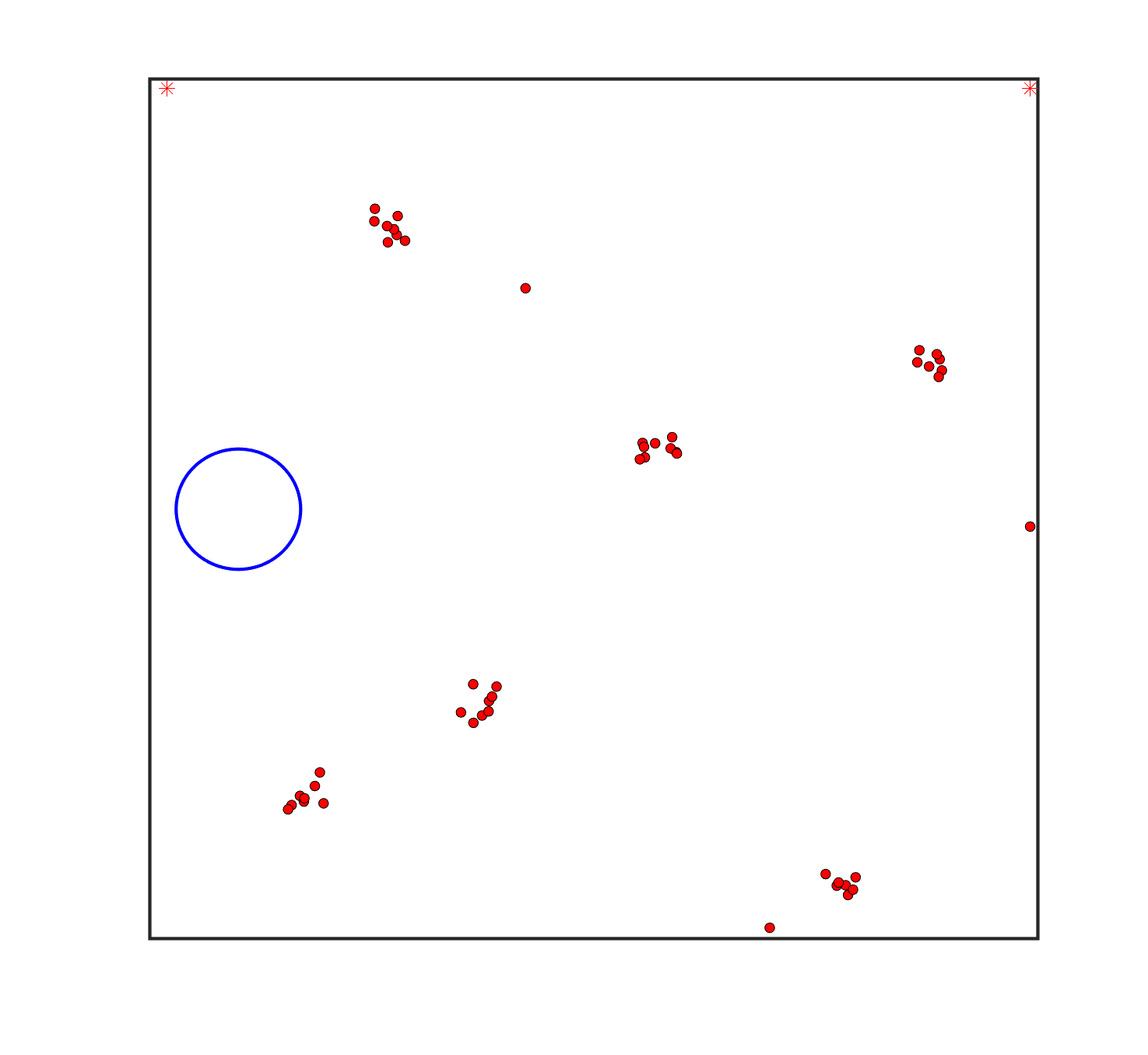}
    \label{Case4_Initailisation}
    }
    
    \subfloat[Case5 (100*100, $N=100$)]{
    \includegraphics[width=0.23\textwidth,height=3.5cm]{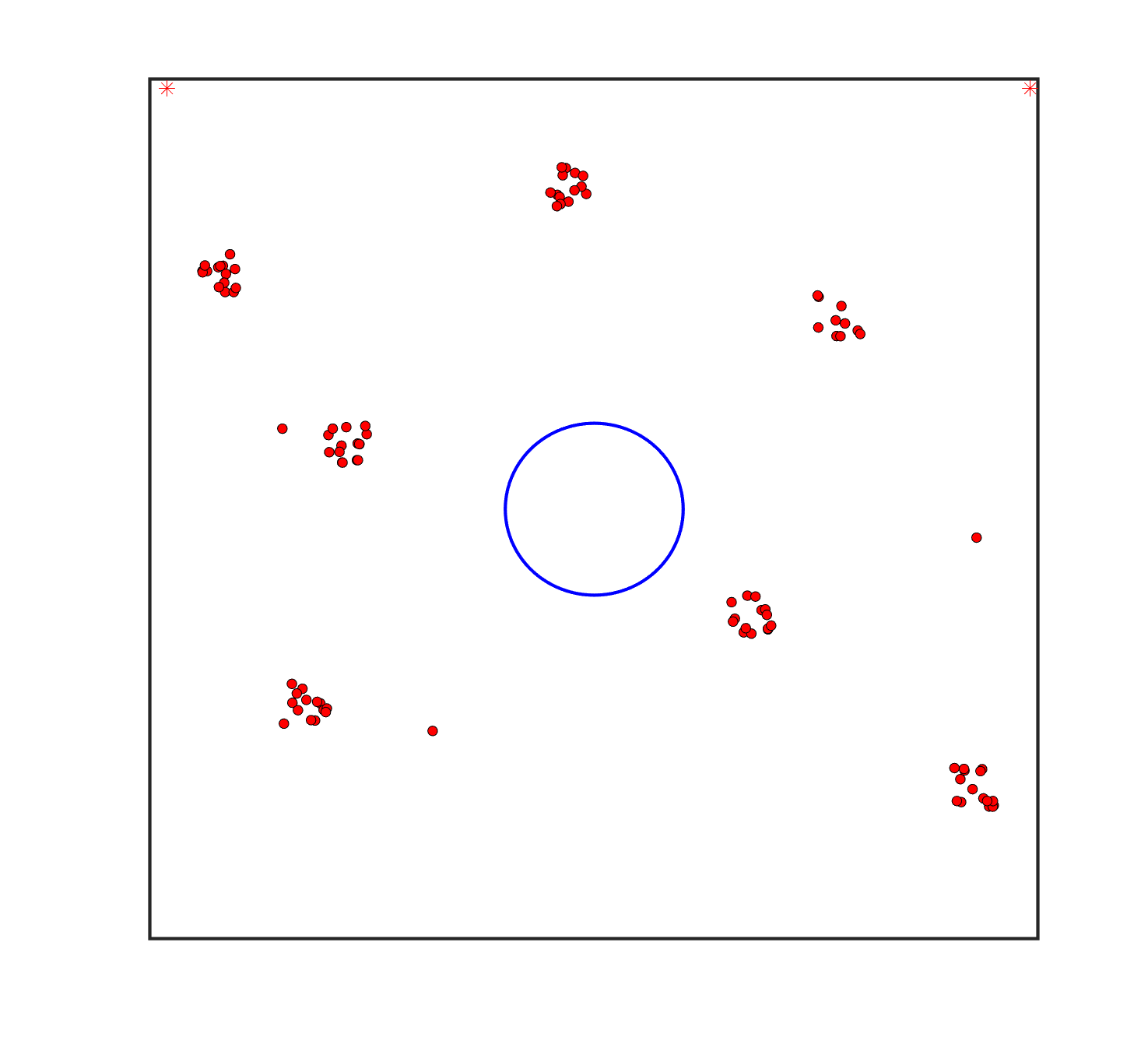}
    \label{Case5_Initailisation}
    }
    \subfloat[Case6 (100*100, $N=100$)]{
    \includegraphics[width=0.23\textwidth,height=3.5cm]{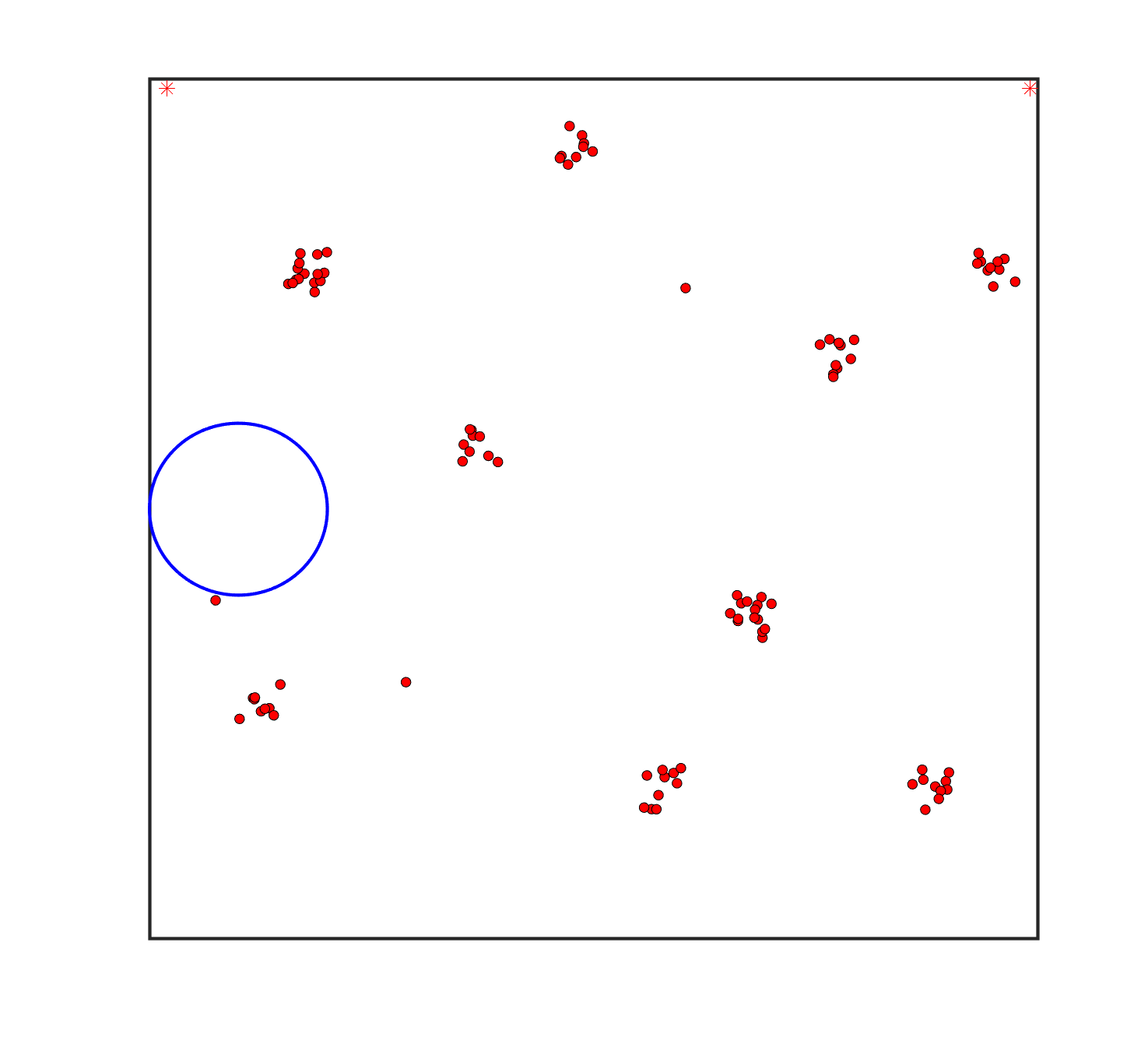}
    \label{Case6_Initailisation}
    }
    \subfloat[Case7 (50*50, $N=20$)]{
    \includegraphics[width=0.23\textwidth,height=3.5cm]{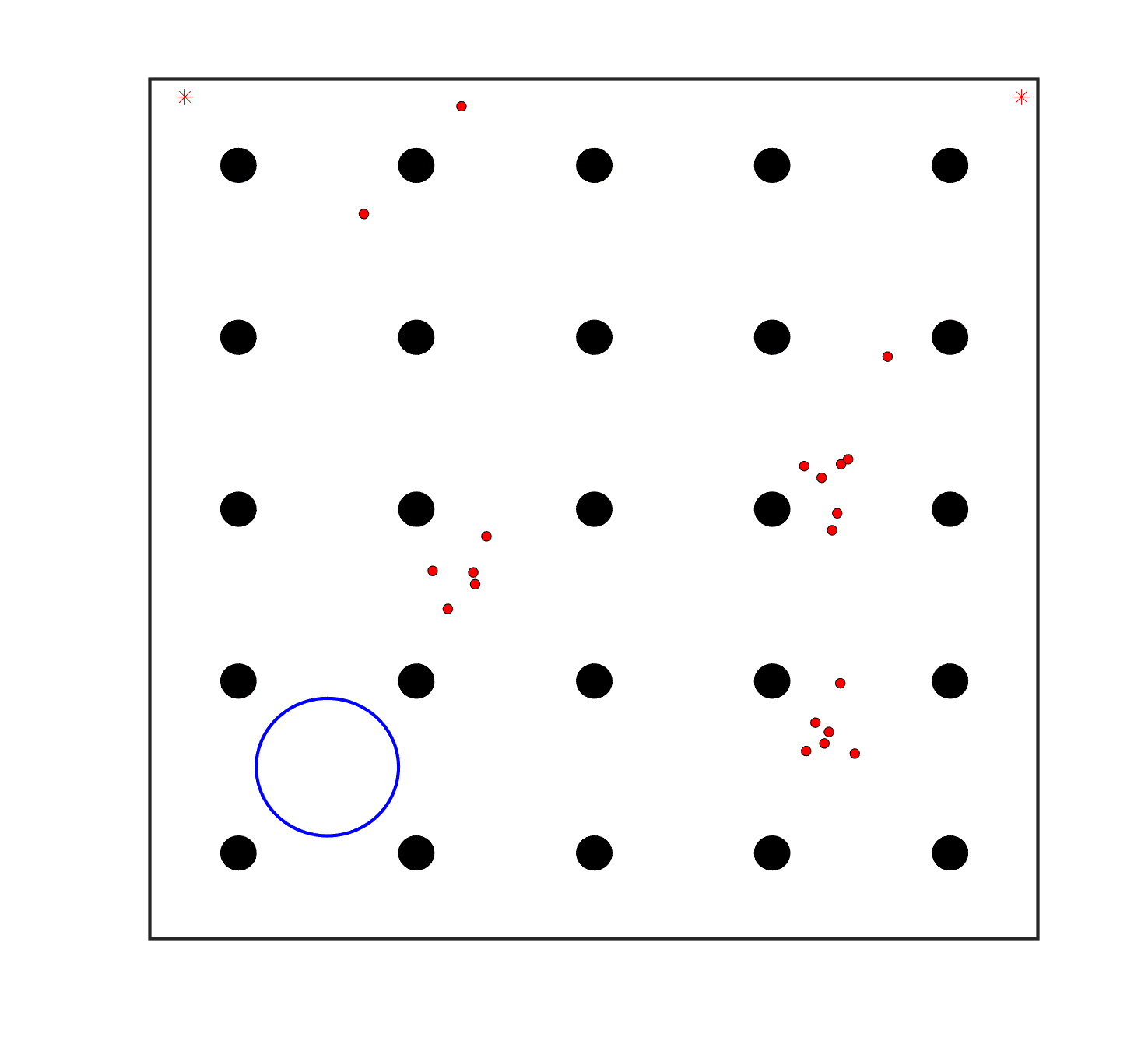}
    \label{Case7_Initailisation}
    }
    \subfloat[Case8 (100*100, $N=20$)]{
    \includegraphics[width=0.23\textwidth,height=3.5cm]{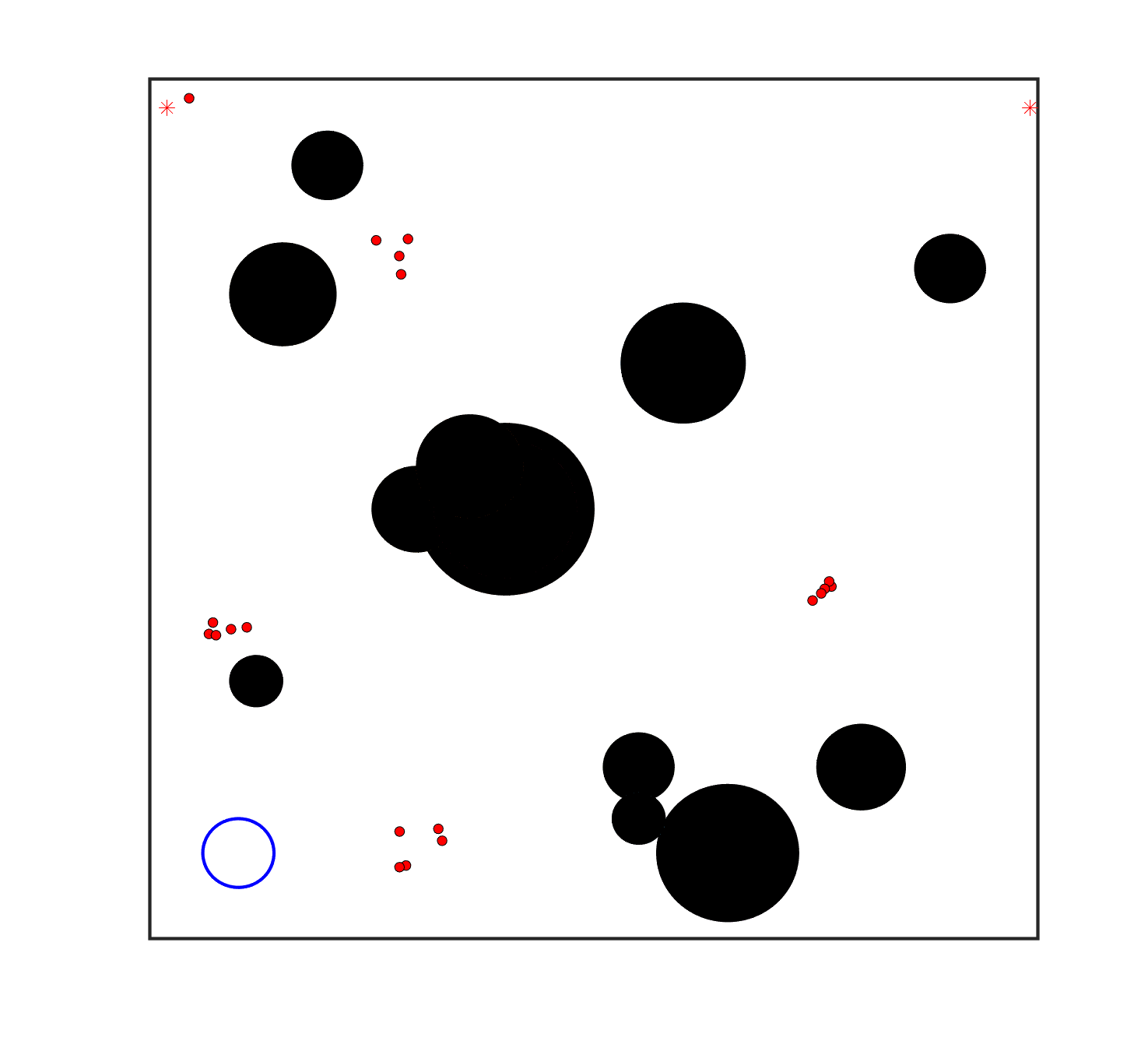}
    \label{Case8_Initailisation}
    }
    
    \subfloat[Case9 (100*100, $N=20$)]{
    \includegraphics[width=0.23\textwidth,height=3.5cm]{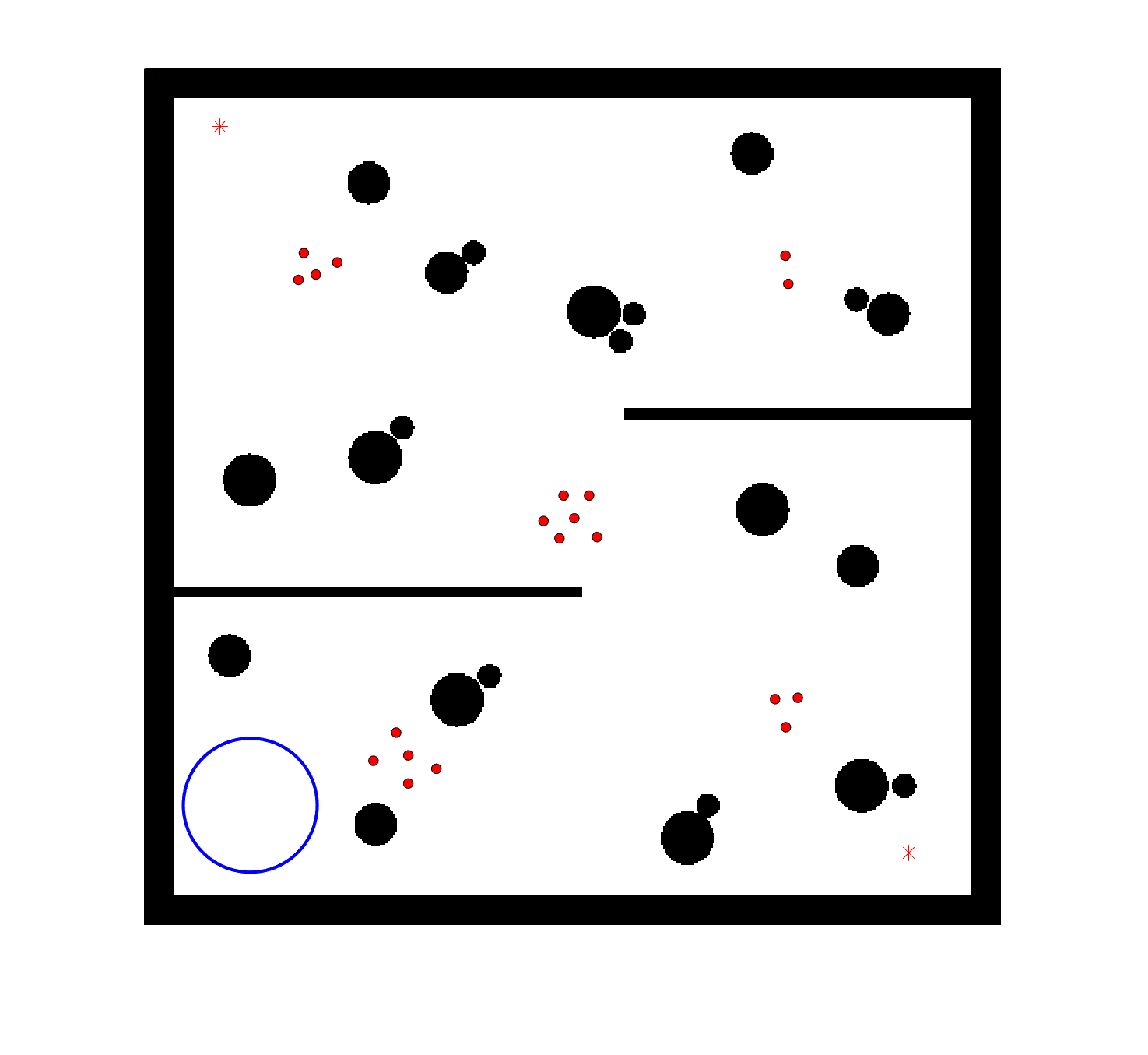}
    \label{Case9_Initailisation}
    }
     \subfloat[Case10 (100*100, $N=50$)]{
    \includegraphics[width=0.23\textwidth,height=3.5cm]{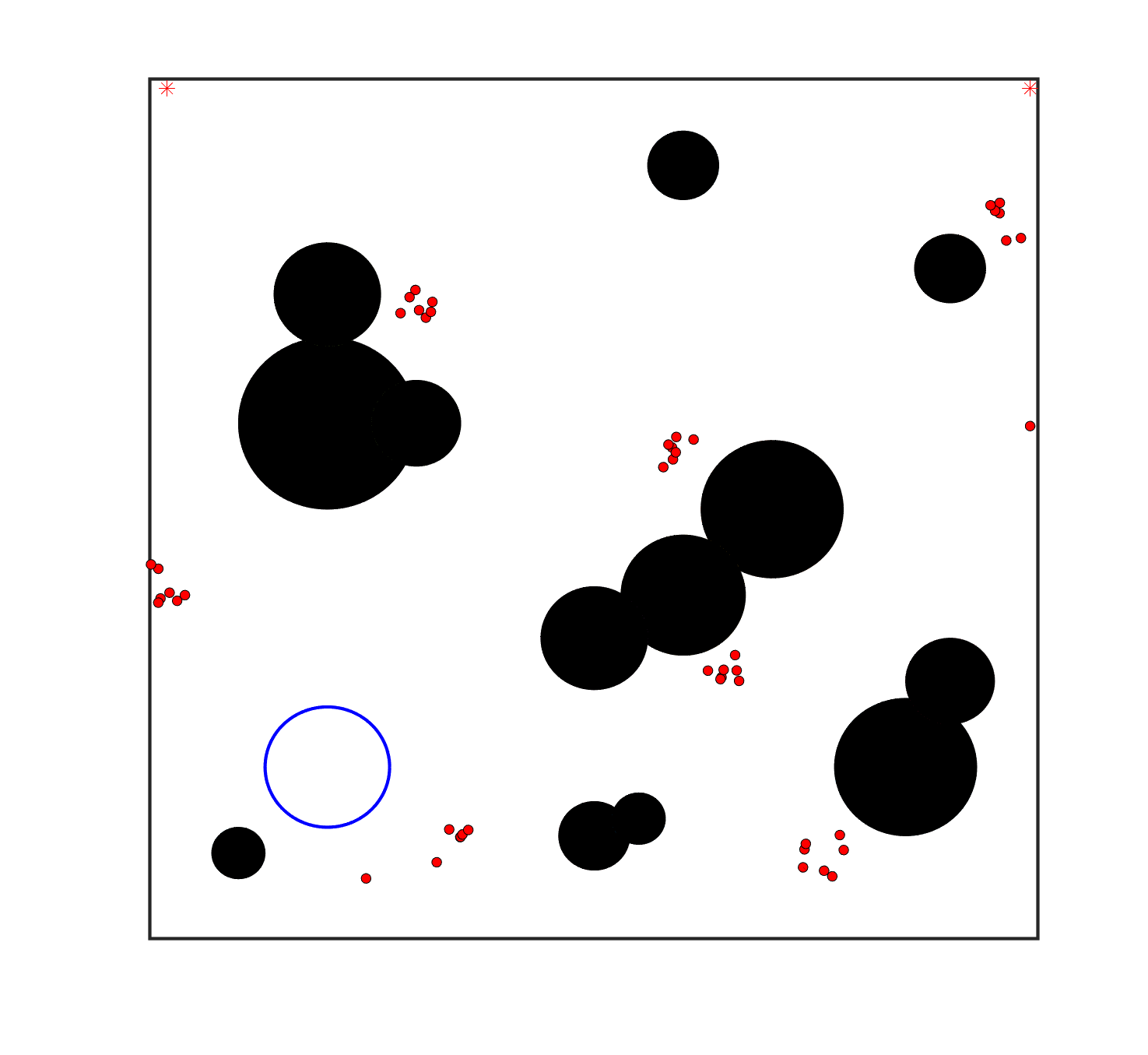}
    \label{Case10_Initailisation}
    }
     \subfloat[Case11 (100*100, $N=50$)]{
    \includegraphics[width=0.23\textwidth,height=3.5cm]{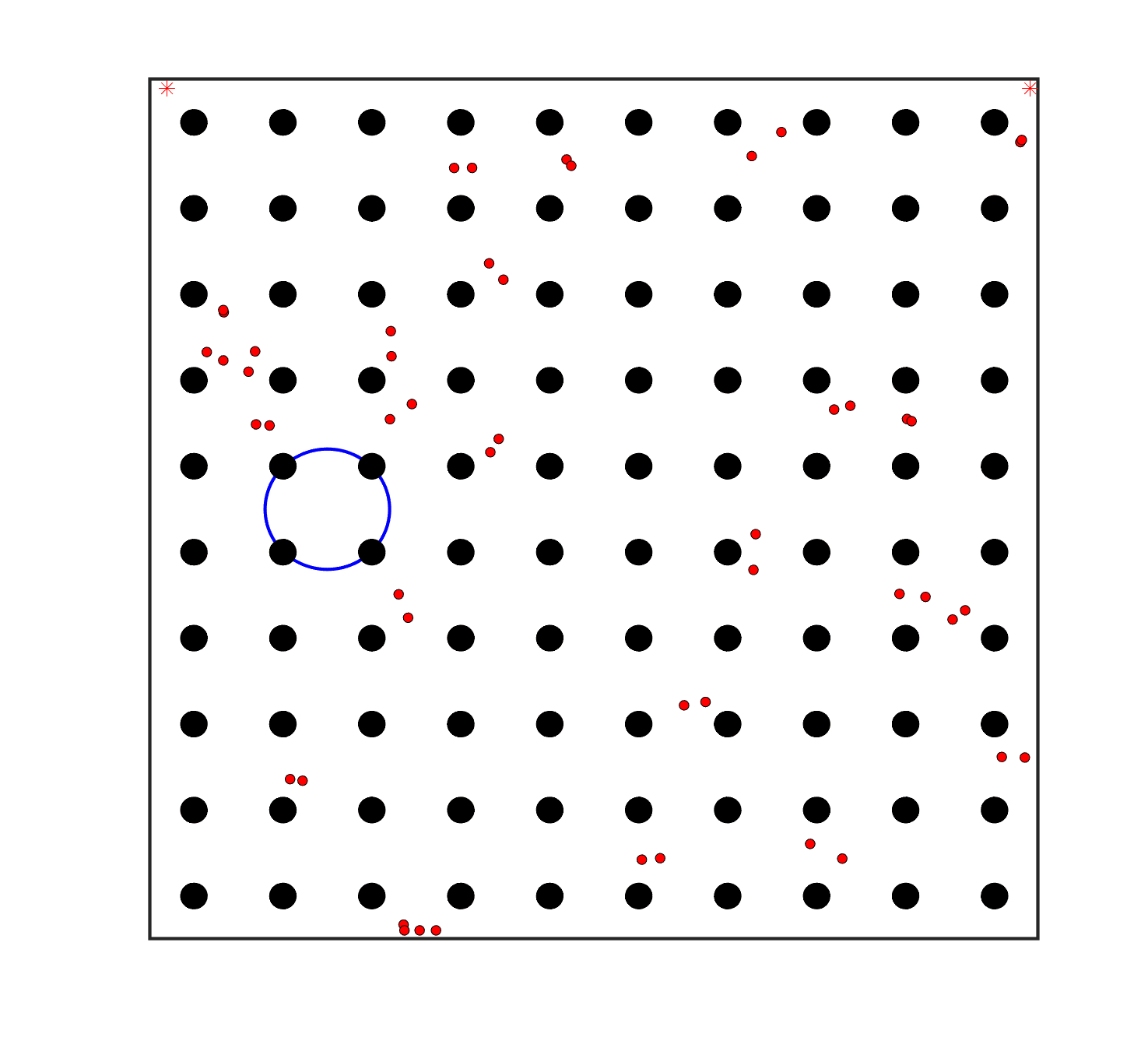}
    \label{Case11_Initailisation}
    }
     \subfloat[Case12 (100*100, $N=50$)]{
    \includegraphics[width=0.23\textwidth,height=3.5cm]{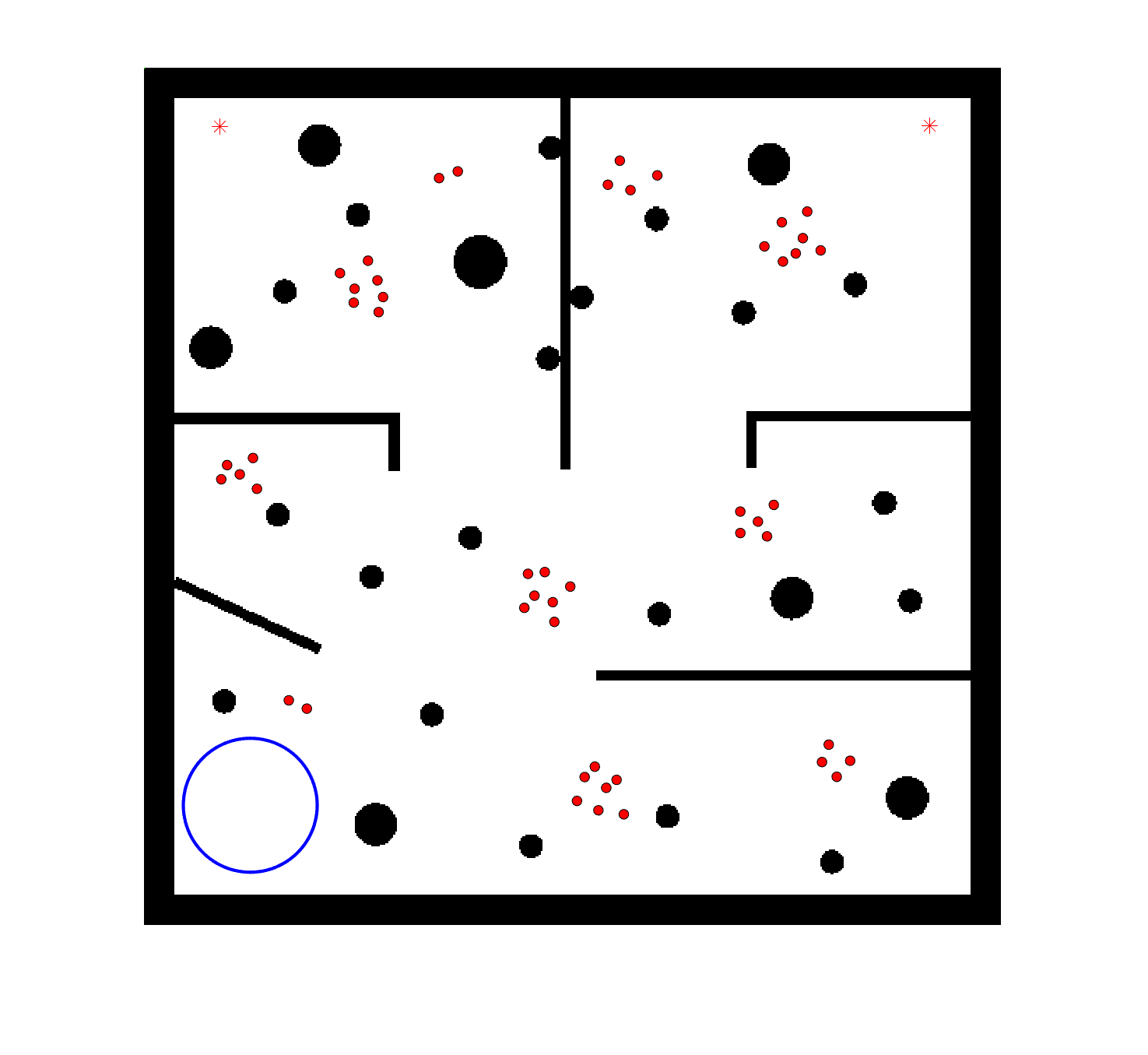}
    \label{Case12_Initailisation}
    }
    
    \subfloat[Case13 (100*100, $N=50$)]{
    \includegraphics[width=0.23\textwidth,height=3.5cm]{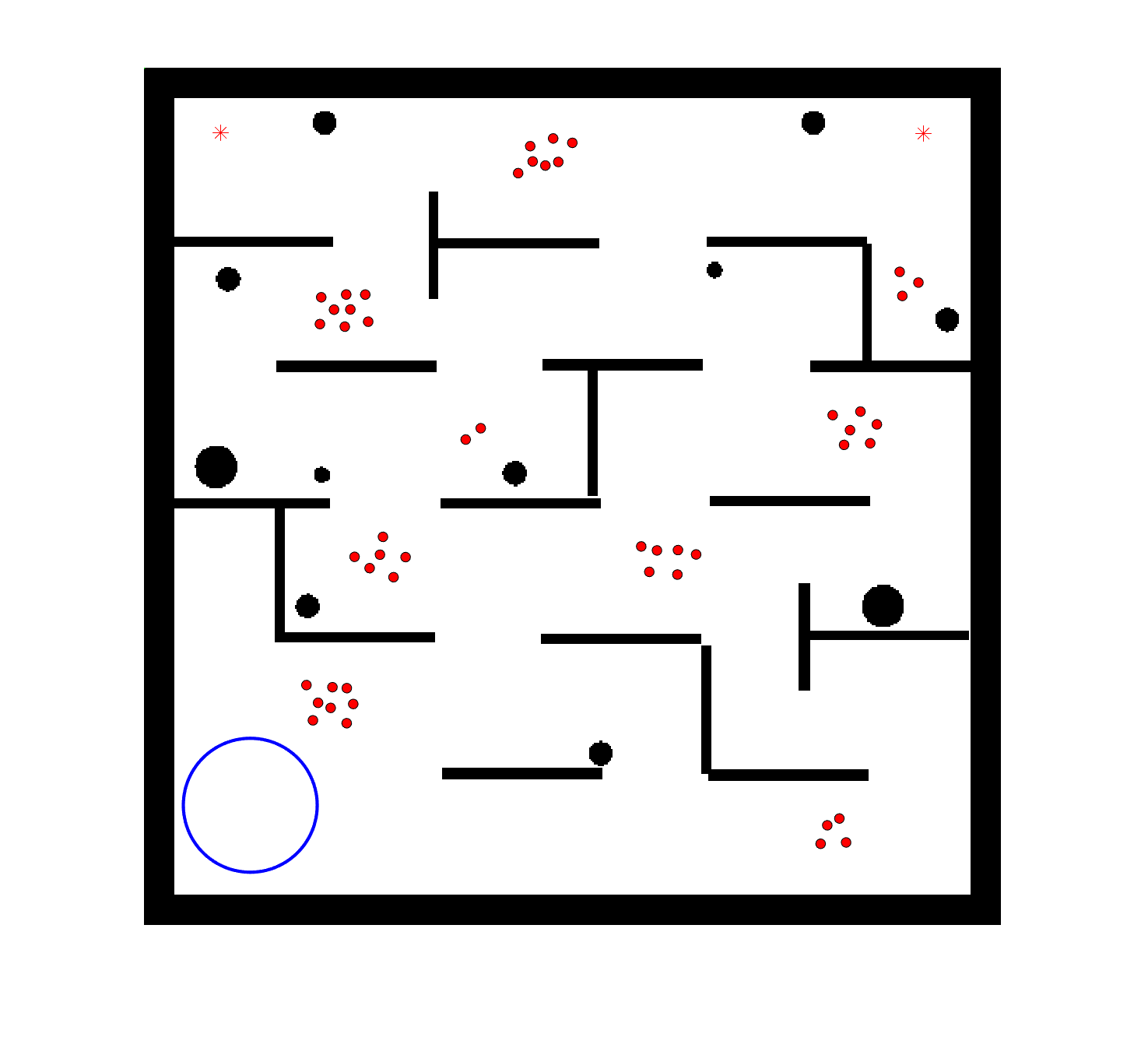}
    \label{Case13_Initailisation}
    }
    \subfloat[Case14 (100*100, $N=100$)]{
    \includegraphics[width=0.23\textwidth,height=3.5cm]{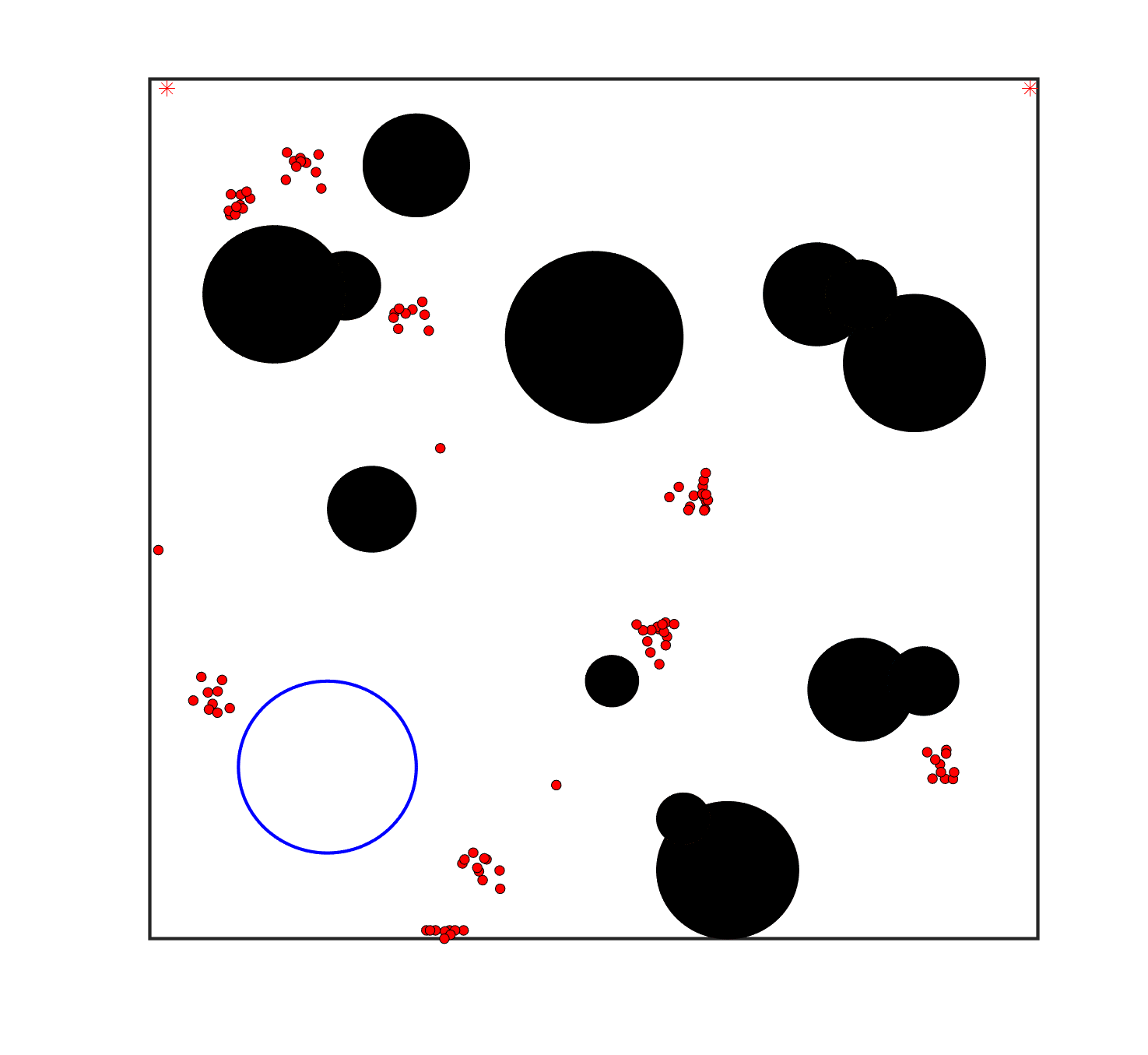}
    \label{Case14_Initailisation}
    }
     \subfloat[Case15 (100*100, $N=100$)]{
    \includegraphics[width=0.23\textwidth,height=3.5cm]{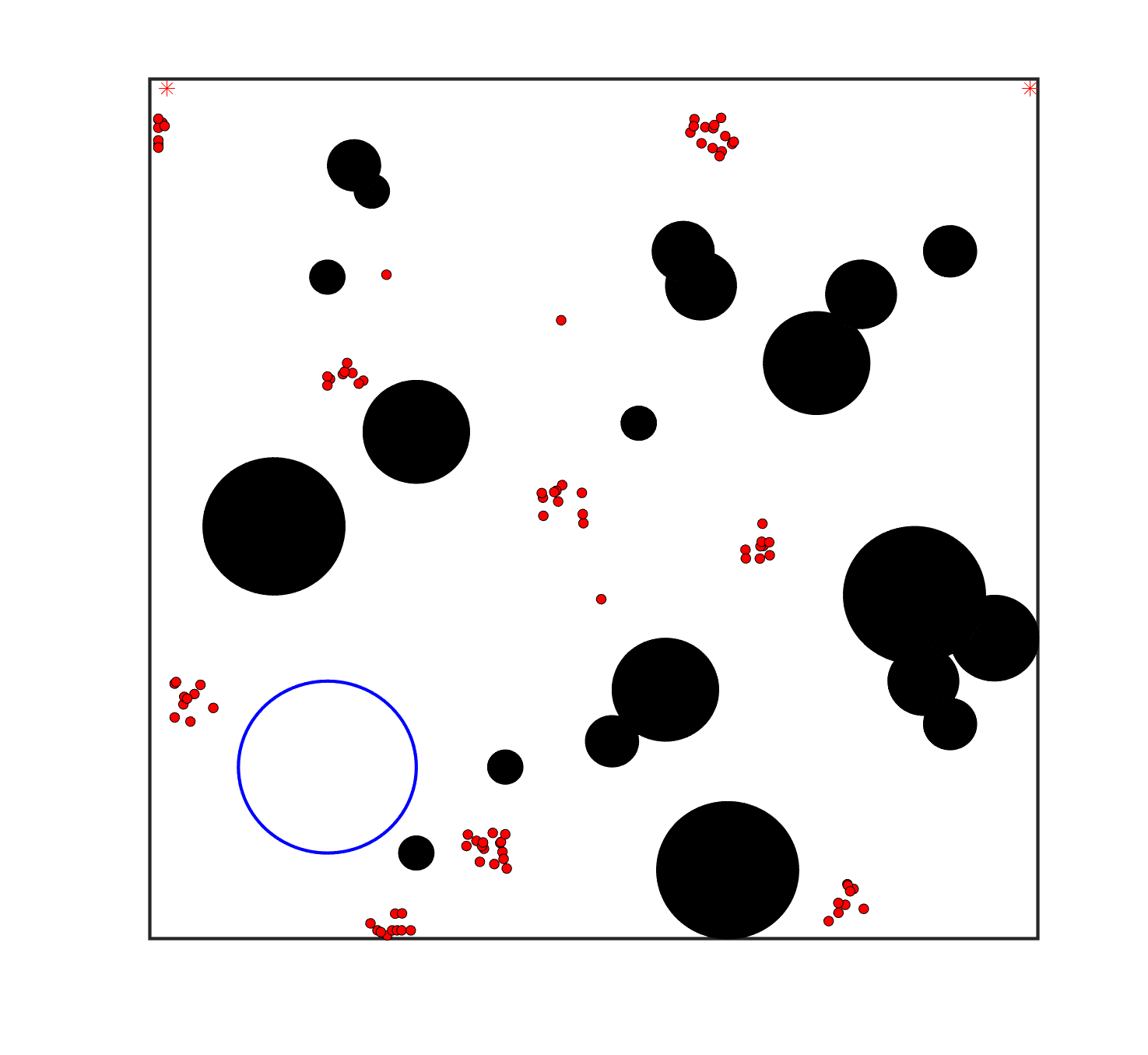}
    \label{Case15_Initailisation}
    }
   \subfloat[Case16 (100*100, $N=100$)]{
    \includegraphics[width=0.23\textwidth,height=3.5cm]{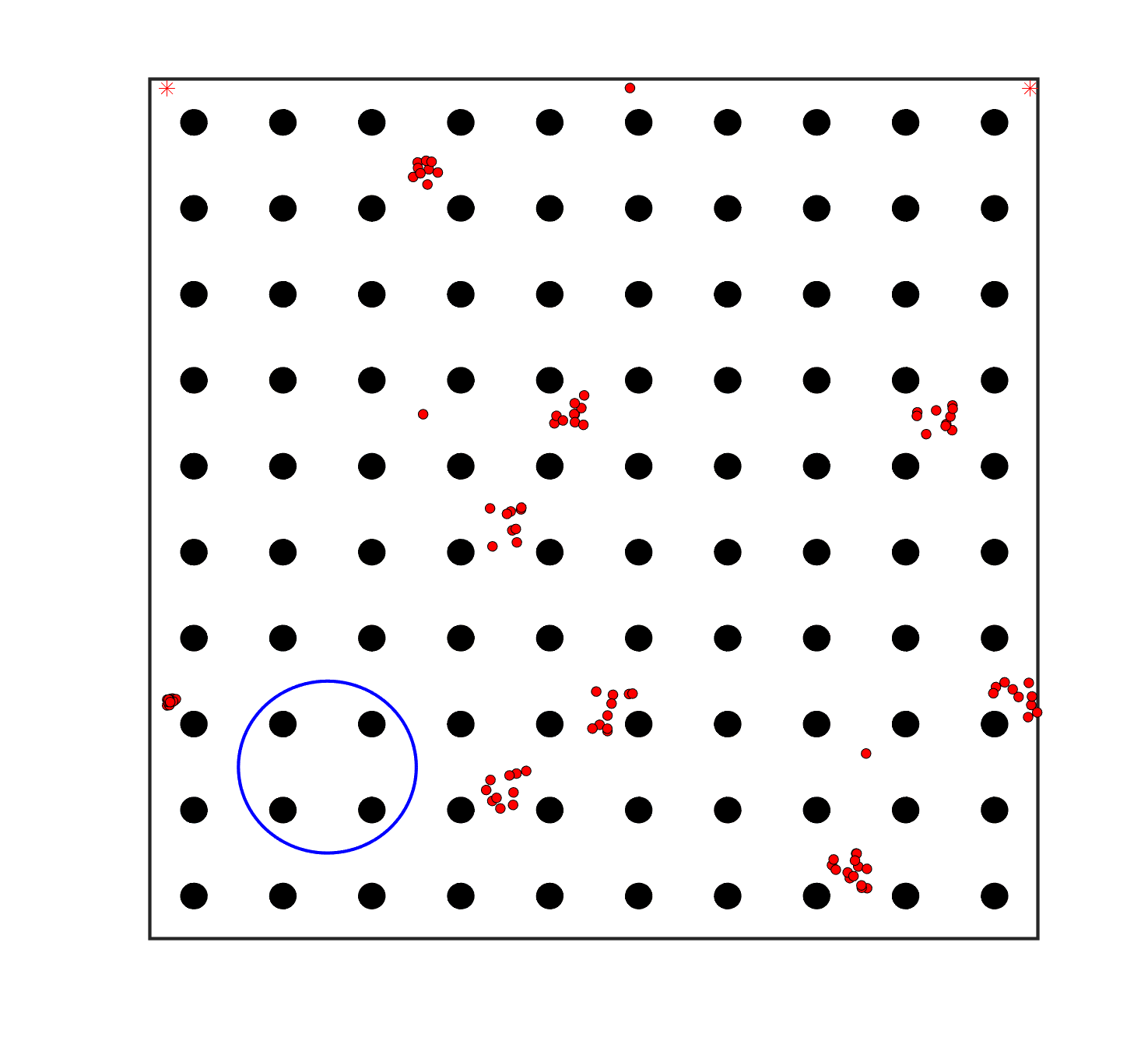}
    \label{Case16_Initailisation}
    }
   
    \subfloat[Case17 (100*100, $N=100$)]{
    \includegraphics[width=0.23\textwidth,height=3.5cm]{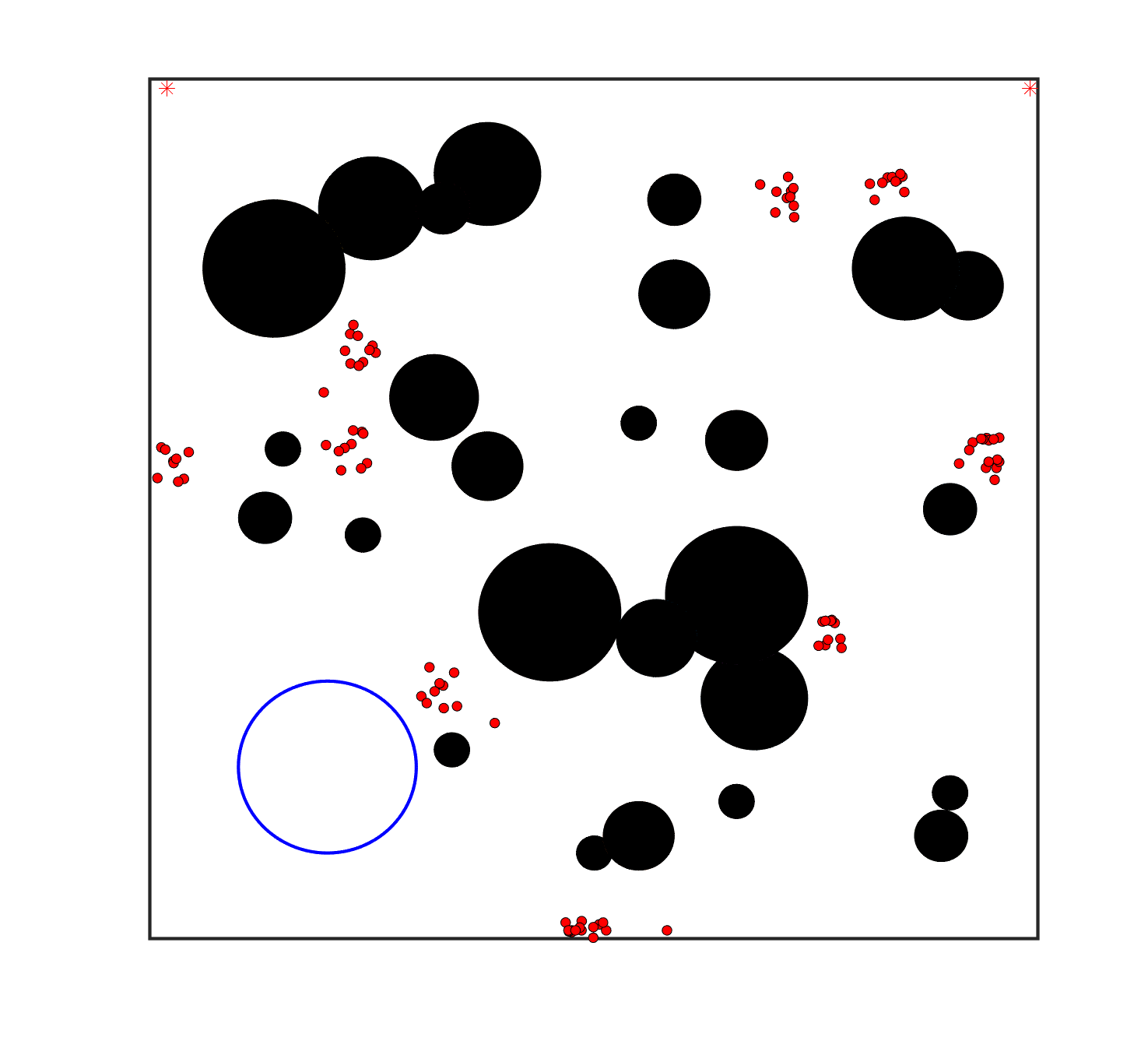}
    \label{Case17_Initailisation}
    }
    \subfloat[Case18 (100*100, $N=100$)]{
    \includegraphics[width=0.23\textwidth,height=3.5cm]{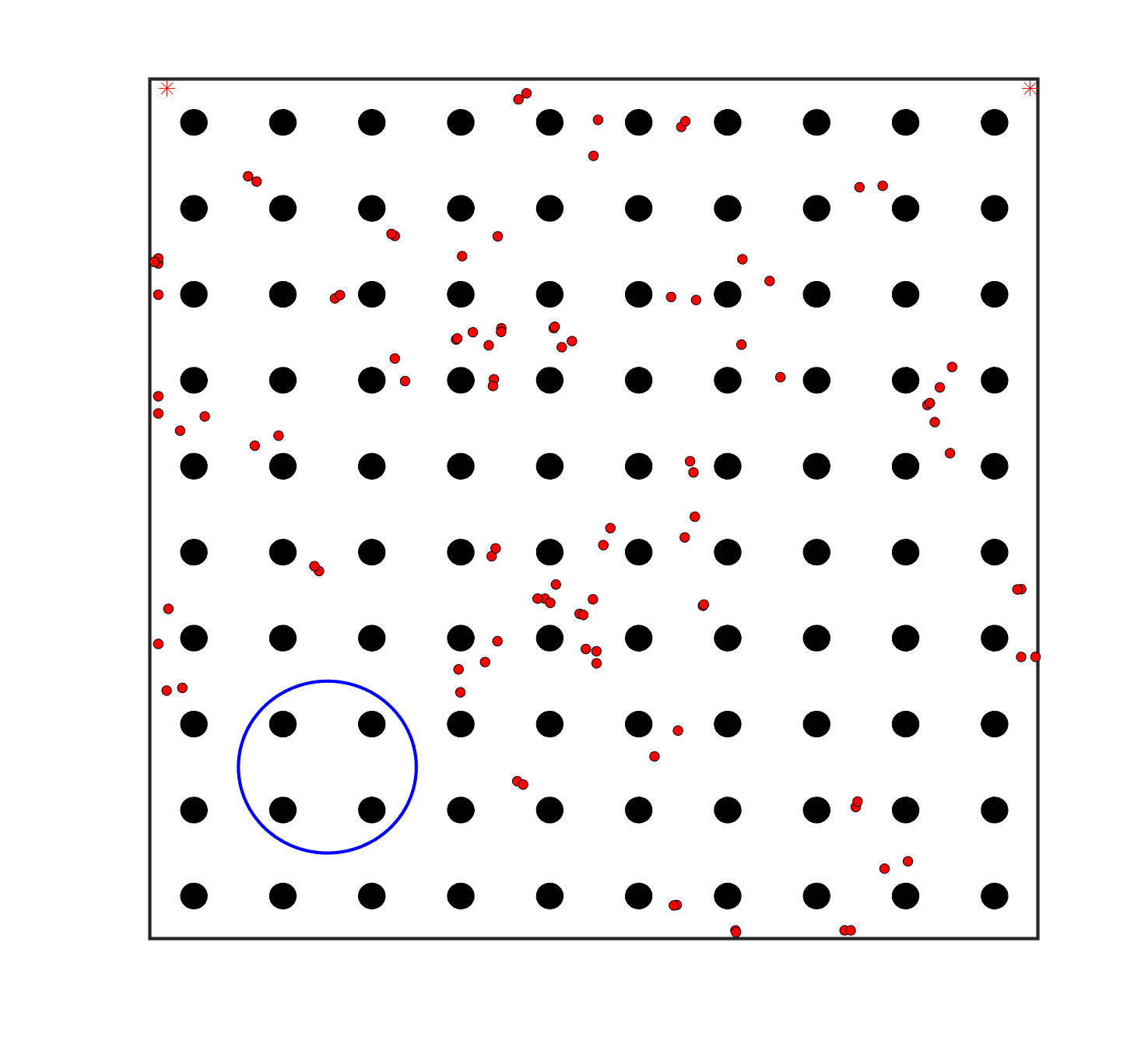}
    \label{Case18_Initailisation}
    }
    \subfloat[Case19 (100*100, $N=100$)]{
    \includegraphics[width=0.23\textwidth,height=3.5cm]{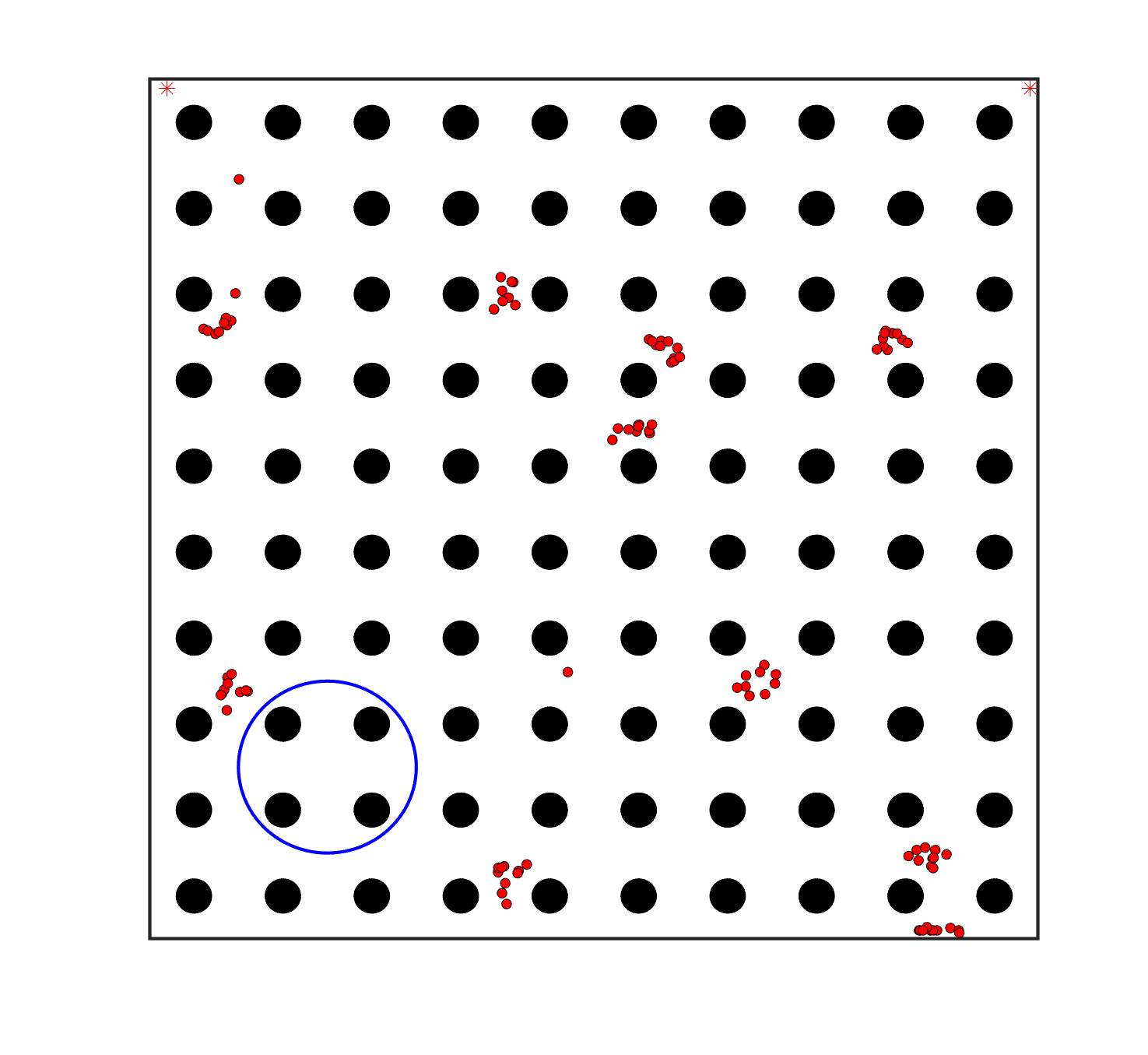}
    \label{Case19_Initailisation}
    }
    \subfloat[Case20 (100*100, $N=100$)]{
    \includegraphics[width=0.23\textwidth,height=3.5cm]{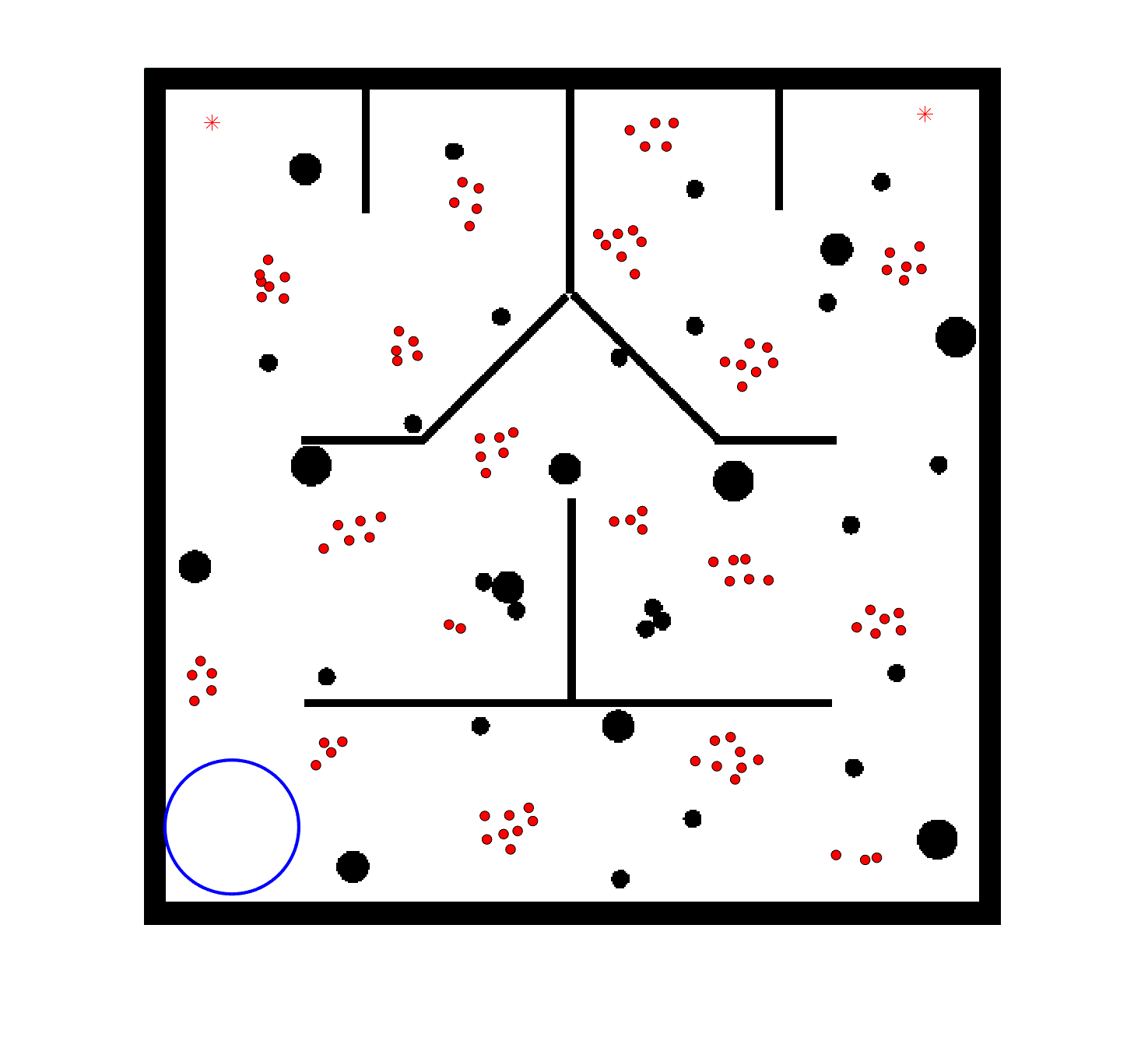}
    \label{Case20_Initailisation}
    }
    
    \caption{The visualised initialisation of Case 1-Case 20}
    \label{fig_VisualisedInitialisation}
 \end{figure*}

 To evaluate the planning-assisted shepherding model and the hierarchical mission planning algorithm, experiments are conducted on a set of synthetic shepherding problems with different levels of complexity. Table~\ref{tab_benchmark} presents the details of the 20 benchmark problems, showing the environment size (mostly $100\times100$), the number of sheep $N$ (20, 50, 100) and if obstacles are contained in each case. The benchmark set consists of three groups, the obstacle-free group, the obstacle-contained group with small swarms and the obstacle-contained group with large swarms. The cases in each group have an increasing level of complexity. Fig.~\ref{fig_VisualisedInitialisation} shows the visualised initialisation of each case with red dots representing the sheep, red asterisks representing sheepdogs, black areas denoting obstacles and a blue circle representing the goal area. Cases 1-6 are obstacle-free environments with an increasing level of complexity  by varying the environment size, $N$, the goal location and the swarm initialisation. Cases 7-20 are obstacle-contained environments where the density of obstacles further impacts the problem\textquoteright s complexity. The initialisation in cases 11 and 18 is based on randomly distributed sheep individuals, while the initialisation in other cases is based on randomly distributed sub-swarms. 

\par For each problem instance, the experiments are conducted 20 times to capture the statistical behaviour. The number of the maximum time steps for each run is set to $T=300 + 20*N$. Three metrics are recorded to evaluate shepherding performance, including 1) SR: the success rate, i.e., number of times the shepherding mission was completed out of 20 runs; 2) No. of steps: the number of time steps consumed to complete the shepherding mission, and 3) path length: the total moving distance of the sheepdog. The mean and standard deviation of only the successful runs are presented for the no. of steps and path length. 

\begin{table*}[!ht]
\begin{spacing}{0.9}
    \centering
    \caption{Basic features of the benchmark problems}  
    \resizebox{0.8\textwidth}{!}{
    \begin{tabular}{|c|c|c|c|c|c|c|c|c|c|c|c|c|c|}
    \hline
    Group & \multicolumn{6}{c|}{Group 1: Obstacle-free cases }  & \multicolumn{4}{c|}{Group 2: Obstacle-contained cases... } \\
    \hline
    Case & 1 & 2 & 3 & 4 & 5 & 6 & 7 & 8 & 9& 10   \\
     \hline
    Environment size & 50*50 & 100*100 & 100*100  &100*100 &100*100 &100*100 & 50*50 &100*100 &100*100 &100*100 \\
    Number of sheep $N$ & 20  & 20 &50 &50 & 100  & 100 & 20 & 20 & 20 & 50\\
    Obstacles&N &N & N&N& N&N  & Y& Y&Y & Y \\
    \hline
     Group & \multicolumn{3}{c|}{...with small sheep swarms}  & \multicolumn{7}{c|}{Group 3: Obstacle-contained cases with large sheep swarms } \\
     \hline
    Case & 11 & 12 & 13 & 14 & 15 & 16 & 17 & 18 & 19& 20 \\
     \hline
    Environment size & 100*100 & 100*100 & 100*100  &100*100 &100*100 &100*100 &100*100 &100*100  &100*100 &100*100 \\
    Number of sheep  & 50&50 &50 &100 & 100 &100 & 100 &100 & 100& 100\\
    Obstacles &Y & Y&Y& Y&Y &Y & Y& Y&Y& Y\\
     \hline
    \end{tabular}
    }
    \label{tab_benchmark}
    \end{spacing}
\end{table*}

\begin{table*}[!ht]
    \centering
    \caption{Parameters setting of the shepherding model}  
    \resizebox{0.6\textwidth}{!}{
    \begin{tabular}{|c|c|c|c|c|c|c|c|c|c|c|c|c|c|}
    \hline
    $W_{\pi_v}$ & $W_{\pi\Lambda}$ & $W_{\pi\beta}$ & $W_{\pi\pi}$ & $W_{\pi o}$ & $W_{e\pi_i}$ & $W_{e\beta_j}$  & $R_\Lambda$ & $R_{\pi\beta}$ & $R_{\pi\pi}$ & $R_{\pi o}$ & $R_s$ \\
    \hline
    0.5 & 1.05 &1 & 2 & 3 &0.3&0.3 &4 & 8 & 0.4 & 2 & 4\\
     \hline
    \end{tabular}
    }
    \label{tab_parameters}
\end{table*}

\subsection{Parameters values and the effects of the adaptively-switch}
\par The parameters in MMAS are set according to \cite{stutzle2000max}. To be specific, the maximum number of iteration is set to 600; the population size is set to the problem dimension; $\alpha=1$, $\beta=2$, $\rho=0.98$, $\tau_{min}=1/D$, $\tau_{max}=1$.  The number of neighbours in the A* search algorithm is set to 8, and the scaling factor is 1.  Table~\ref{tab_parameters} presents the setting of most parameters involved in the shepherding model by referring to~\cite{strombom2014solving}.  The newly introduced parameter $R_{th}$ and the cost weights $\alpha_1,\ \alpha_2$ are analysed in the following. 

\par $R_{th}$ determines the threat area size that the sheepdog should try to avoid in the no-interaction mode. It should be no less than $R_s=4$ to keep the safe operating distance from the sheep and no more than the sheepdog\textquoteright s influence range $R_{\pi \beta}=8$.  Therefore, we tested the effects of $R_{th}$ on shepherding performance of 6 representative cases by setting $R_{th}$ to  4, 5, 6, 7 and 8. 
Table~\ref{tab_threat_range} presents the experimental results. Without the explicit declaration, the best results are shown in bold in all the following tables, and Wilcoxon rank-sum tests are conducted between the best results and other results for each case to test if their performances are statistically different, with a significance level of 0.05. `*' indicates the significant difference. 

 \begin{table*}[!ht]
    \centering
    \caption{The effects of $R_{th}$ on shepherding performance } 
   \resizebox{0.4\textwidth}{!}{ 
    \begin{tabular}{|c|c|c|c|}
    \hline
      Case& SR   & No. of steps &Path length    \\
    \hline
    \multicolumn{4}{|c|}{$R_{th}=4$ }  \\
     \hline
Case1&1.00 & \textbf{131.00$\pm$12.14} & \textbf{214.70$\pm$20.69} \\
     	Case3&1.00 & \textbf{396.05$\pm$18.85} & \textbf{510.17$\pm$38.77} \\
     	Case7&1.00 & \textbf{232.50$\pm$47.04} & \textbf{409.36$\pm$94.24} \\
     	Case11&0.95 & 996.21$\pm$89.63& 1401.94$\pm$166.63\\
     	Case16&1.00 & \textbf{1191.75$\pm$137.19} & \textbf{955.65$\pm$143.42} \\
     	Case18&0.60 & 2071.17$\pm$156.11& 1858.70$\pm$235.51\\
     	\hline
     \multicolumn{4}{|c|}{$R_{th}=5$  }  \\
     \hline
     	Case1&1.00 & 133.85$\pm$13.57& 220.77$\pm$24.12\\
     	Case3&1.00 & 432.25$\pm$33.11*& 567.76$\pm$53.05*\\
     	Case7&0.95 & 249.05$\pm$70.08& 439.90$\pm$130.25\\
     	Case11&0.80 & \textbf{991.00$\pm$113.56} & \textbf{1391.99$\pm$238.58} \\
     	Case16&1.00 & 1304.50$\pm$136.34*& 1152.18$\pm$257.64*\\
     	Case18&0.25 & 2088.00$\pm$108.46& 1892.80$\pm$293.40\\
     		\hline
     \multicolumn{4}{|c|}{$R_{th}=6$ }  \\
     \hline
     	Case1&1.00 & 146.05$\pm$15.13*& 239.70$\pm$29.18*\\
     	Case3&1.00 & 433.80$\pm$25.89*& 577.58$\pm$46.55*\\
     	Case7&1.00 & 258.85$\pm$110.51& 451.99$\pm$212.87\\
     	Case11&0.90 & 1091.89$\pm$78.26& 1560.16$\pm$194.04\\
     	Case16&1.00 & 1345.30$\pm$155.60*& 1156.08$\pm$250.54*\\
     	Case18&0.35 & 2038.29$\pm$172.45& 1875.59$\pm$215.09\\
     	\hline
     \multicolumn{4}{|c|}{$R_{th}=7$  }  \\
     \hline
     	Case1&1.00 & 156.70$\pm$14.75*& 262.80$\pm$30.15*\\
     	Case3&1.00 & 449.55$\pm$25.79*& 595.32$\pm$49.50*\\
     	Case7&1.00 & 254.65$\pm$74.52& 442.00$\pm$140.60\\
     		Case11&0.85 & 1023.06$\pm$132.27& 1442.46$\pm$245.77\\
     	Case16&0.95 & 1337.58$\pm$167.01*& 1127.46$\pm$242.16*\\
     	Case18&0.25 & 2052.00$\pm$129.39& 1806.39$\pm$71.94\\
     		\hline
     \multicolumn{4}{|c|}{$R_{th}=8$  }  \\
     \hline
     	Case1&1.00 & 156.50$\pm$16.94*& 257.28$\pm$28.99*\\
     	Case3&1.00 & 453.10$\pm$45.54*& 605.75$\pm$110.63*\\
     	Case7&1.00 & 274.60$\pm$123.49& 485.03$\pm$232.67\\
     	Case11&0.95 & 1109.37$\pm$104.44& 1605.64$\pm$173.35\\
     	Case16&1.00 & 1326.00$\pm$131.83*& 1146.23$\pm$247.51*\\
     	Case18&0.40 & \textbf{1999.12$\pm$225.17} & \textbf{1717.57$\pm$234.66} \\
    \hline
    \multicolumn{4}{l}{ * represents the statistical significance} \\
    \end{tabular}
    }
    \label{tab_threat_range}
\end{table*}

\par We can observe from Table~\ref{tab_threat_range} that
$R_{th}=4$ performs better than other values, achieving the highest SR on 6 cases and minimum time steps and path length on 4 cases.  It should also be noted that the differences in the results by setting different $R_{th}$ values are not always significant. This is due to the high randomness of sheep behaviours which are impacted by many factors such as the obstacles and the neighbourhood.  But only $R_{th}=4$ is not significantly worse than other values in these cases. This is probably because although $R_{th}=4$ can not minimise the influence of sheepdogs on some sheep on the edge of the swarm, it can avoid most of the poor disturbance behaviours, e.g., crossing the swarm. On the contrary, a large $R_{th}$ might cause unnecessary detours for sheepdogs. Therefore, $R_{th}$ is set to 4 in the following experiments. 

\par The parameters $\alpha_1$ and $\alpha_2$ are parameters for determining the weights of the path length cost and the threat cost. We examined the effects of the ratio of $\alpha_2$ to $\alpha_1$ on shepherding performance of some representative cases by fixing $\alpha_1$ to 1 and varying $\alpha_2$ to 0, 20, 40, 60, 80 and 100. In particular, $\alpha_2=0$ means that the no-interaction mode turns into the interaction mode, and the adaptively switch between these two modes is disabled. Table~\ref{tab_alpha2} shows that, similar to the effects of $R_{th}$, the influence of different $\alpha_2$ values is not significant in some cases. Particularly, the shepherding performance is not very sensitive to the change of $\alpha_2$ if it is non-zero. This is because the change of non-zero $\alpha_2$ values only slightly impacts the path planning results in no-interaction mode, which does not cause a significant difference in the shepherding. However, when $\alpha_2 =0$, which disables the adaptively switch, the shepherding performance of more cases is impacted. Therefore,  $\alpha_1=1,\ \alpha_2=100$ are set in the following experiments.

 \begin{table*}[!ht]
    \centering
    \caption{ The effects of $\alpha_2$ on shepherding performance ($\alpha_1=1$)} 
    \resizebox{\textwidth}{!}{ 
    \begin{tabular}{|c|c|c|c|c|c|c|c|c|c|c|c|c|c|c|c|c|c|c|c|}
    \hline
       & \multicolumn{3}{c|}{$\alpha_2=0$ }  & \multicolumn{3}{c|}{$\alpha_2=20$}  & \multicolumn{3}{c|}{$\alpha_2=40$}  \\
     \hline
      Case& SR   & No. of steps &Path length  &  SR  & No. of steps &Path length & SR  & No. of steps &Path length  \\
    \hline
 C1&1.00 & 140.95$\pm$14.53*& 232.33$\pm$23.90*& 1.00 & 131.45$\pm$14.80& 218.62$\pm$26.35& 1.00 & \textbf{128.85$\pm$13.18} & \textbf{213.01$\pm$23.70} \\
     	C3&1.00 & 425.35$\pm$36.43*& 565.53$\pm$72.38*& 1.00 & 410.20$\pm$25.86& 530.70$\pm$41.55& 1.00 & 405.35$\pm$25.48& 521.85$\pm$38.94\\
     	C7&1.00 & 234.95$\pm$73.25& 413.66$\pm$135.80& 1.00 & \textbf{208.25$\pm$53.61} & \textbf{364.71$\pm$104.67} & 1.00 & 225.50$\pm$50.62& 399.21$\pm$95.84\\
     	C11&1.00 & 976.25$\pm$130.14& 1396.81$\pm$247.64& 1.00 & 978.00$\pm$120.45& 1380.84$\pm$228.90& 0.95 & 978.00$\pm$130.36& 1376.36$\pm$254.97\\
     		C16&1.00 & 1249.50$\pm$124.46& 1002.57$\pm$170.85& 1.00 & 1202.60$\pm$184.64& 1009.26$\pm$233.03& 1.00 & 1282.00$\pm$136.54& 1091.85$\pm$245.40\\
     	C18&0.20 & 2171.25$\pm$138.81*& 1768.19$\pm$398.51& 0.20 & 2095.25$\pm$189.02*& 1594.14$\pm$271.94& 0.50 & 2198.40$\pm$33.07*& 1606.22$\pm$145.12\\
     	
     		\hline
       & \multicolumn{3}{c|}{$\alpha_2=60$ }  & \multicolumn{3}{c|}{$\alpha_2=80$}  & \multicolumn{3}{c|}{$\alpha_2=100$}  \\
     \hline
      Case& SR   & No. of steps &Path length  &  SR  & No. of steps &Path length & SR  & No. of steps &Path length  \\
    \hline
     C1&1.00 & 134.25$\pm$11.57& 214.39$\pm$18.91& 1.00 & 133.90$\pm$11.24& 221.59$\pm$20.24& 1.00 & 131.00$\pm$12.14& 214.70$\pm$20.69\\
     	C3&1.00 & 407.45$\pm$15.67*& 523.76$\pm$31.46& 1.00 & 406.85$\pm$28.58& 521.03$\pm$44.73& 1.00 & \textbf{396.05$\pm$18.85} & \textbf{510.17$\pm$38.77} \\
     	C7&1.00 & 239.90$\pm$74.86& 421.64$\pm$144.35& 1.00 & 238.35$\pm$59.09& 424.55$\pm$111.83& 1.00 & 232.50$\pm$47.04& 409.36$\pm$94.24\\
     	C11&1.00 & 1004.20$\pm$140.56& 1420.79$\pm$232.61& 0.90 & \textbf{950.06$\pm$139.47} & \textbf{1336.35$\pm$265.27} & 0.95 & 996.21$\pm$89.63& 1401.94$\pm$166.63\\
     	C16&1.00 & 1245.05$\pm$147.89& 1012.75$\pm$206.21& 1.00 & 1225.45$\pm$135.50& 969.26$\pm$132.54& 1.00 & \textbf{1191.75$\pm$137.19} & \textbf{955.65$\pm$143.42} \\
     	C18&0.20 & 2104.25$\pm$127.41*& \textbf{1572.75$\pm$263.46} & 0.30 & 2201.33$\pm$106.89*& 1594.05$\pm$213.48& 0.60 & \textbf{2071.17$\pm$156.11} & 1858.70$\pm$235.51*\\
    \hline
    \multicolumn{4}{l}{ * represents the statistical significance} \\
    \end{tabular}
    }
    \label{tab_alpha2}
\end{table*}

\subsection{Performance of the planning-assisted shepherding}
The proposed method is compared to the reactive shepherding from Str{\"o}mbom et al.~\cite{strombom2014solving}, referred to as Method 1 for convenience, to validate the effectiveness of the proposed shepherding model. As the proposed model consists of offline task planning (grouping and TSP-based sequencing) and online path planning, we further add the shepherding method with only task planning assisted, referred to as Method 2 as a comparative method to evaluate the impact of task planning and path planning separately. The proposed planning-assisted shepherding method is referred to as Method 3 in the comparisons. 

\subsubsection{Planning-assisted shepherding with single-sheepdog}

\begin{figure*} [!ht]
    \centering
    \subfloat[Case3\_planning results]{
    \includegraphics[width=0.23\textwidth,height=3.2cm]{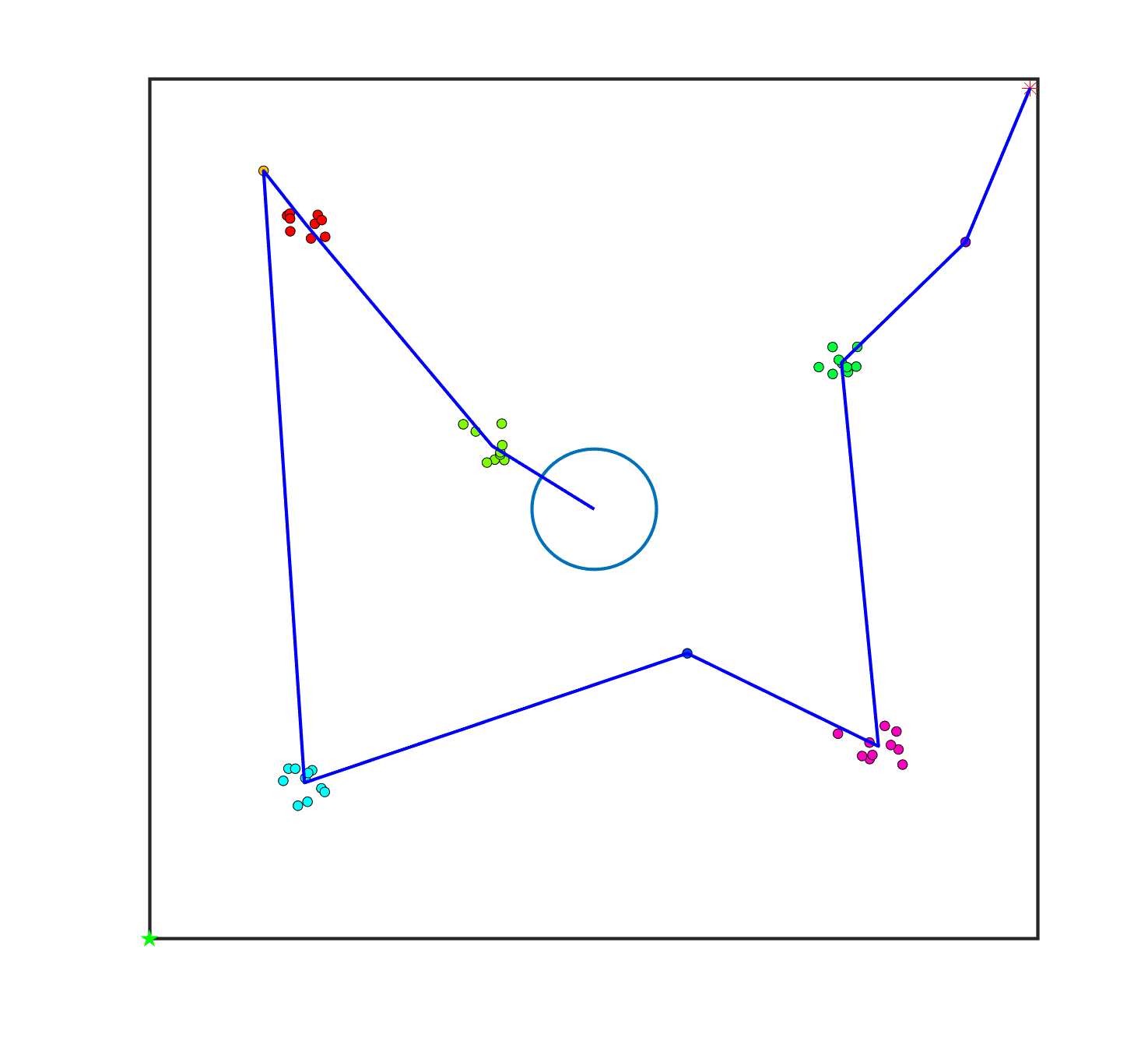}
    \label{fig_singlesheepdog_planningresults_case3}
    }
    \subfloat[Case3\_trajectories\_Method1]{
    \includegraphics[width=0.23\textwidth,height=3.2cm]{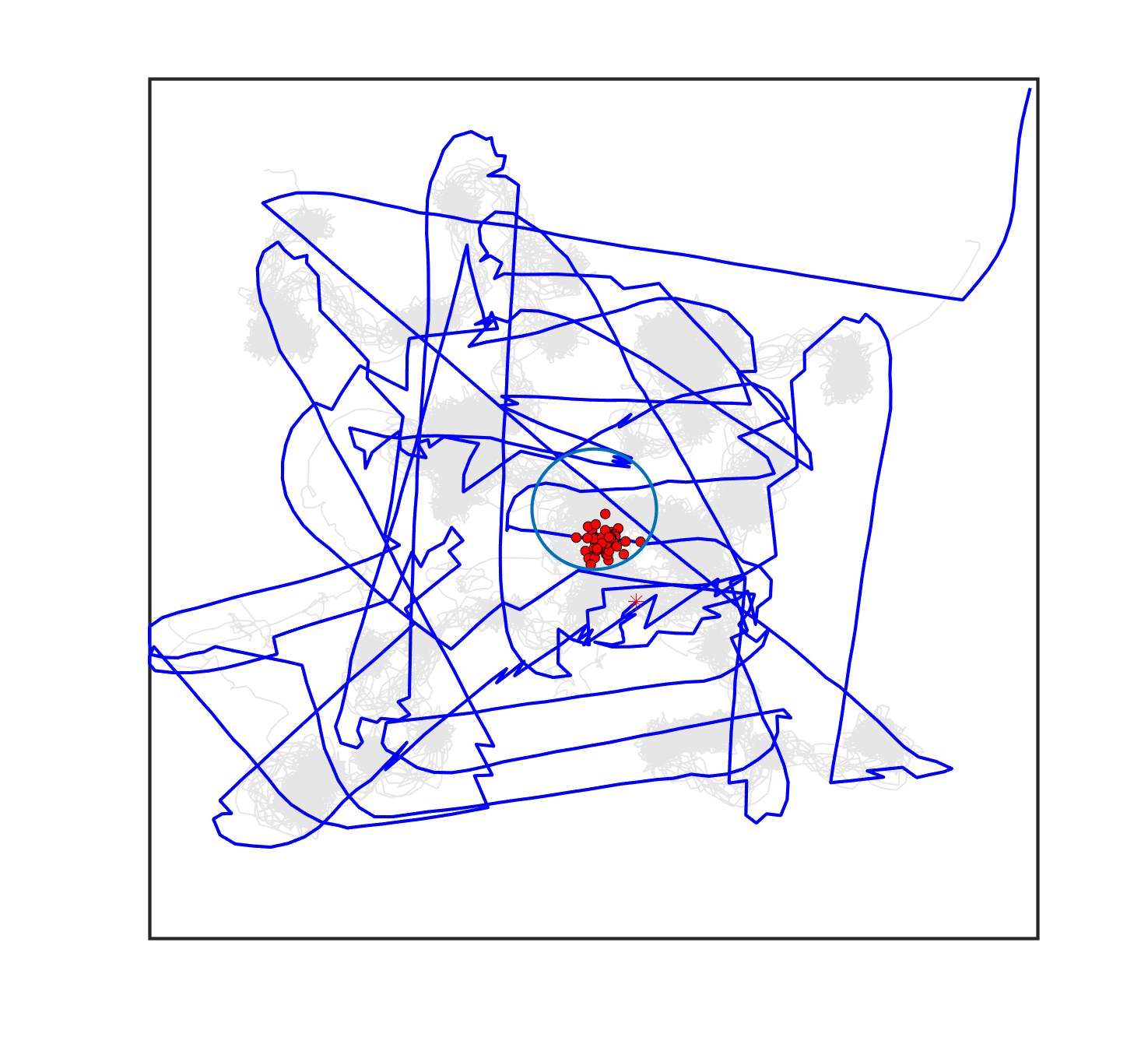}
    \label{fig_trajectory_case3_method1}
    }
    \subfloat[Case3\_trajectories\_Method2]{
    \includegraphics[width=0.23\textwidth,height=3.2cm]{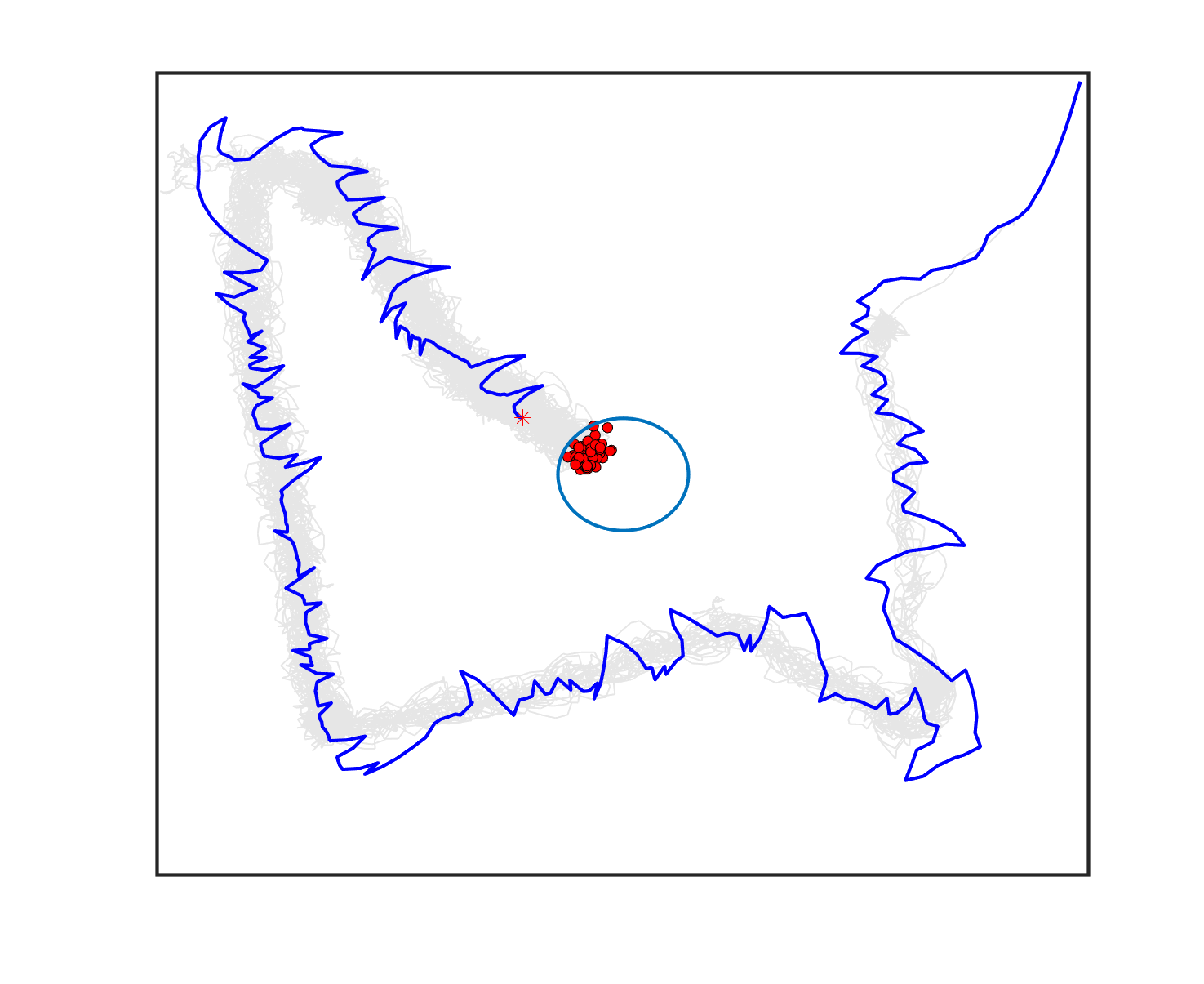}
    \label{fig_trajectory_case3_method2}
    }
    \subfloat[Case3\_trajectories\_Method3]{
    \includegraphics[width=0.23\textwidth,height=3.2cm]{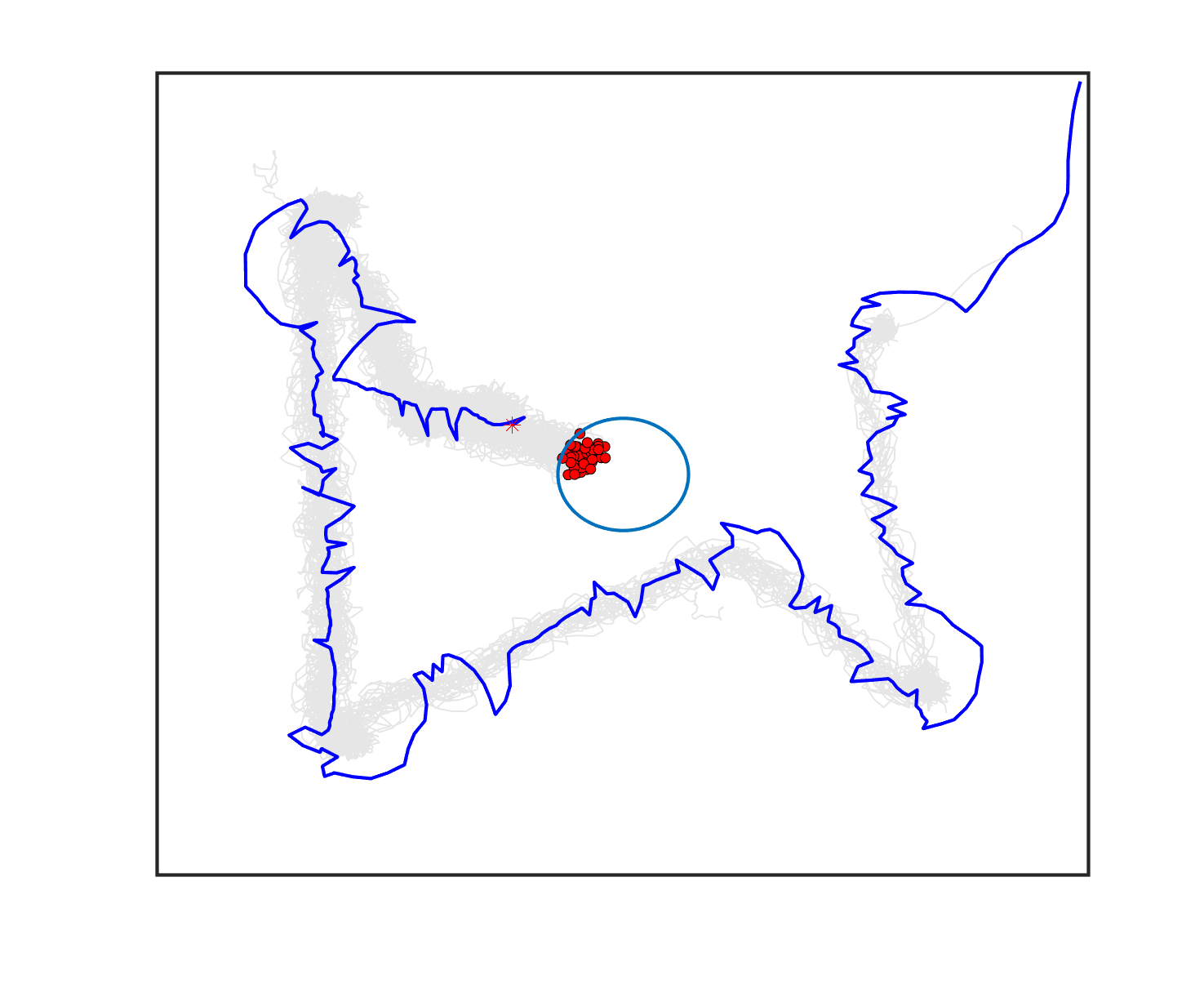}
    \label{fig_trajectory_case3_method3}
    }
    
    \subfloat[Case13\_planning results]{
    \includegraphics[width=0.23\textwidth,height=3.2cm]{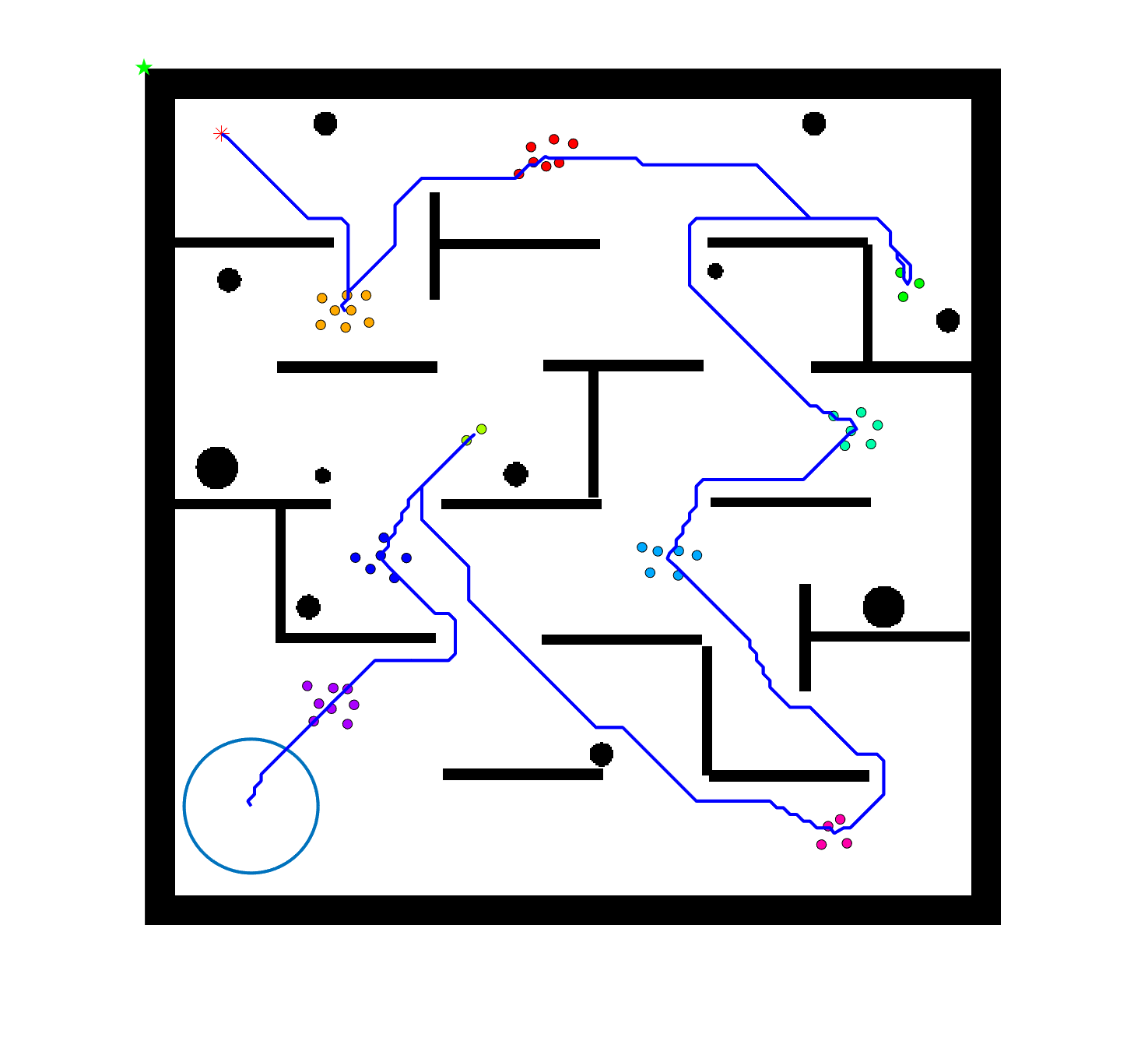}
    \label{fig_singlesheepdog_planningresults_case13}
    }
     \subfloat[Case13\_trajectories\_Method1]{
    \includegraphics[width=0.23\textwidth,height=3.2cm]{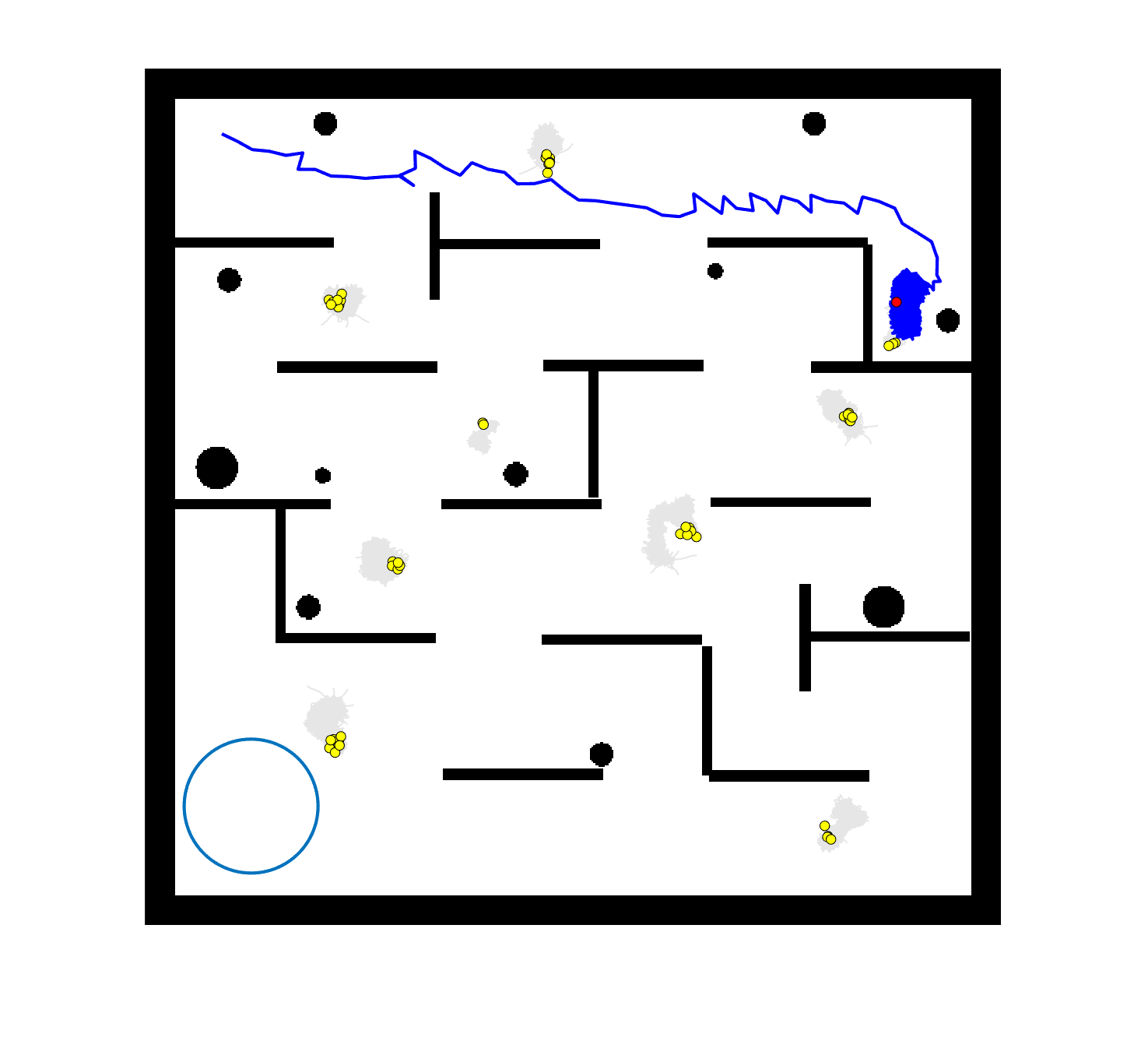}
    \label{fig_trajectory_case13_method1}
    }
    \subfloat[Case13\_trajectories\_Method2]{
    \includegraphics[width=0.23\textwidth,height=3.2cm]{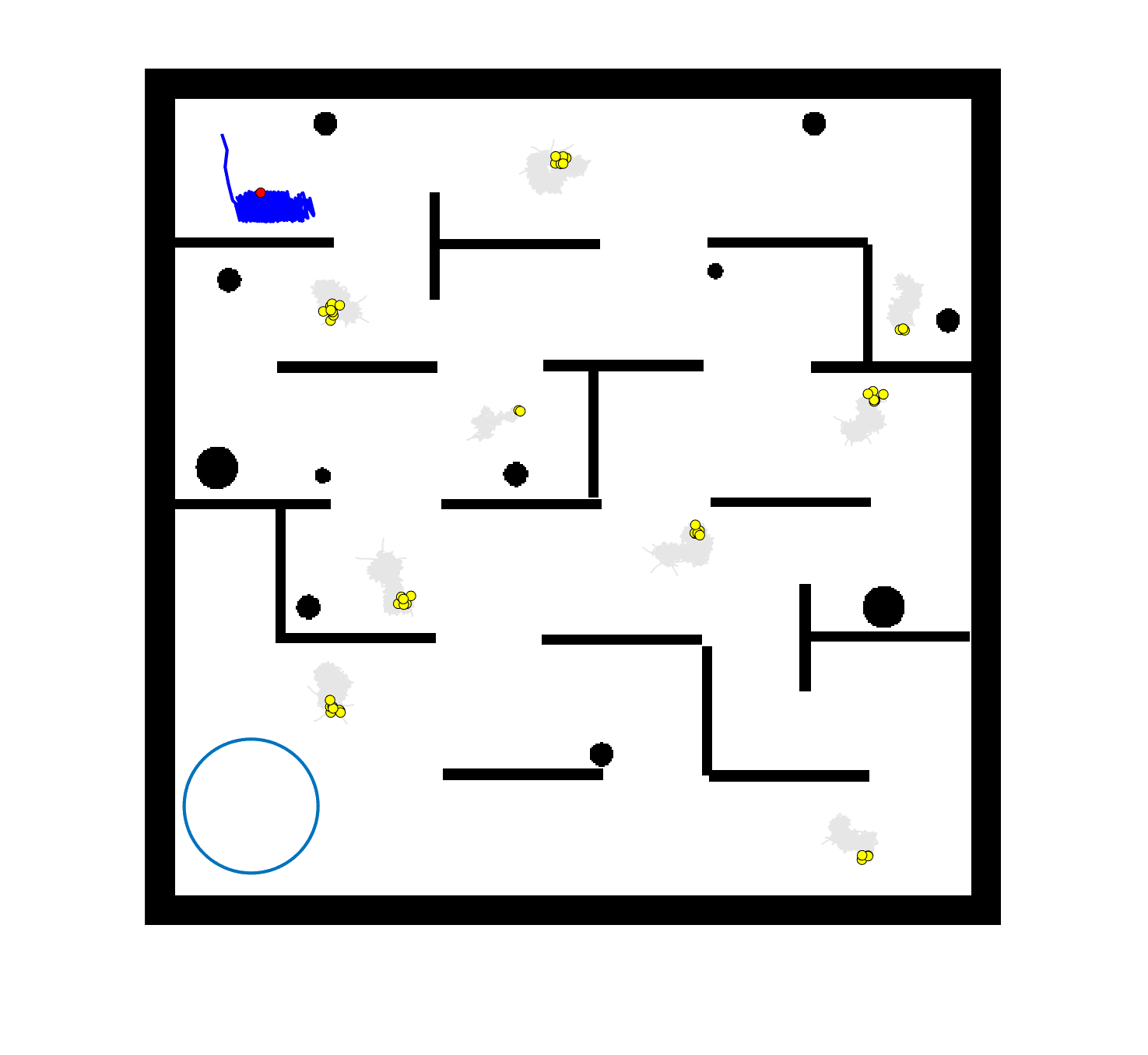}
    \label{fig_trajectory_case13_method2}
    }
    \subfloat[Case13\_trajectories\_Method3]{
    \includegraphics[width=0.23\textwidth,height=3.2cm]{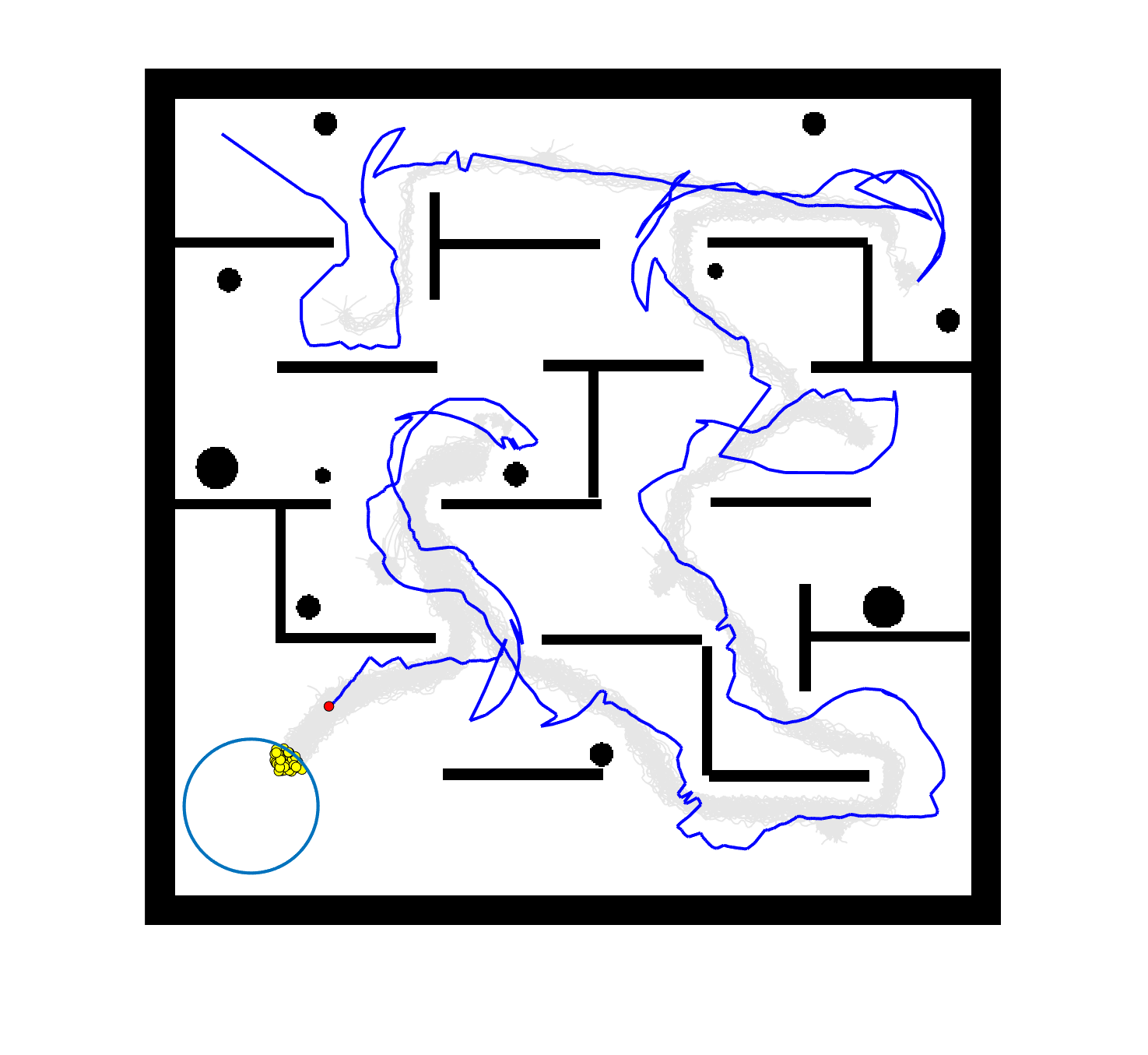}
    \label{fig_trajectory_case13_method3}
    }
    
    \subfloat[Case18\_planning results]{
    \includegraphics[width=0.23\textwidth,height=3.2cm]{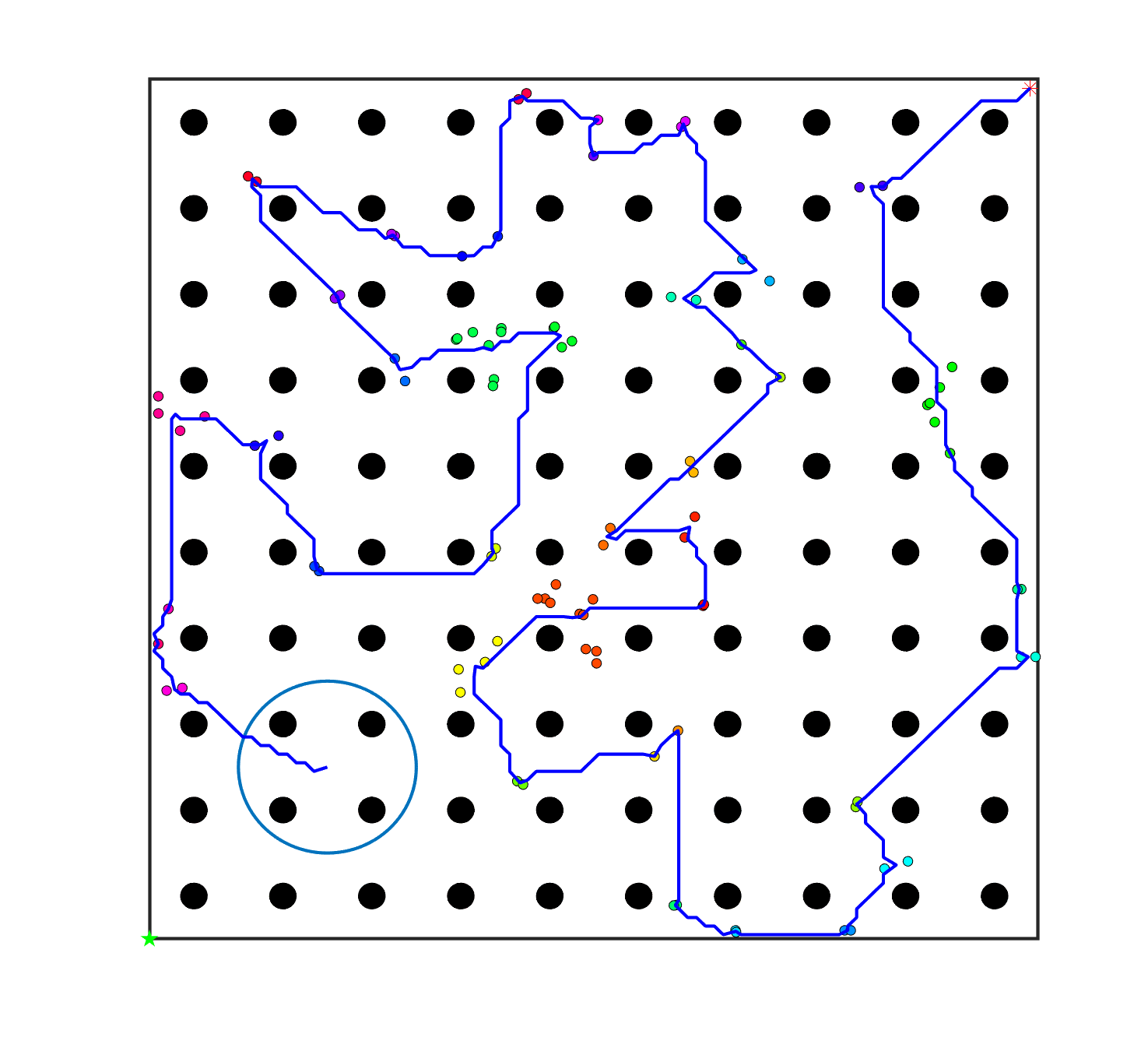}
    \label{fig_singlesheepdog_planningresults_case18}
    }
    \subfloat[Case18\_trajectories\_Method1]{
    \includegraphics[width=0.23\textwidth,height=3.2cm]{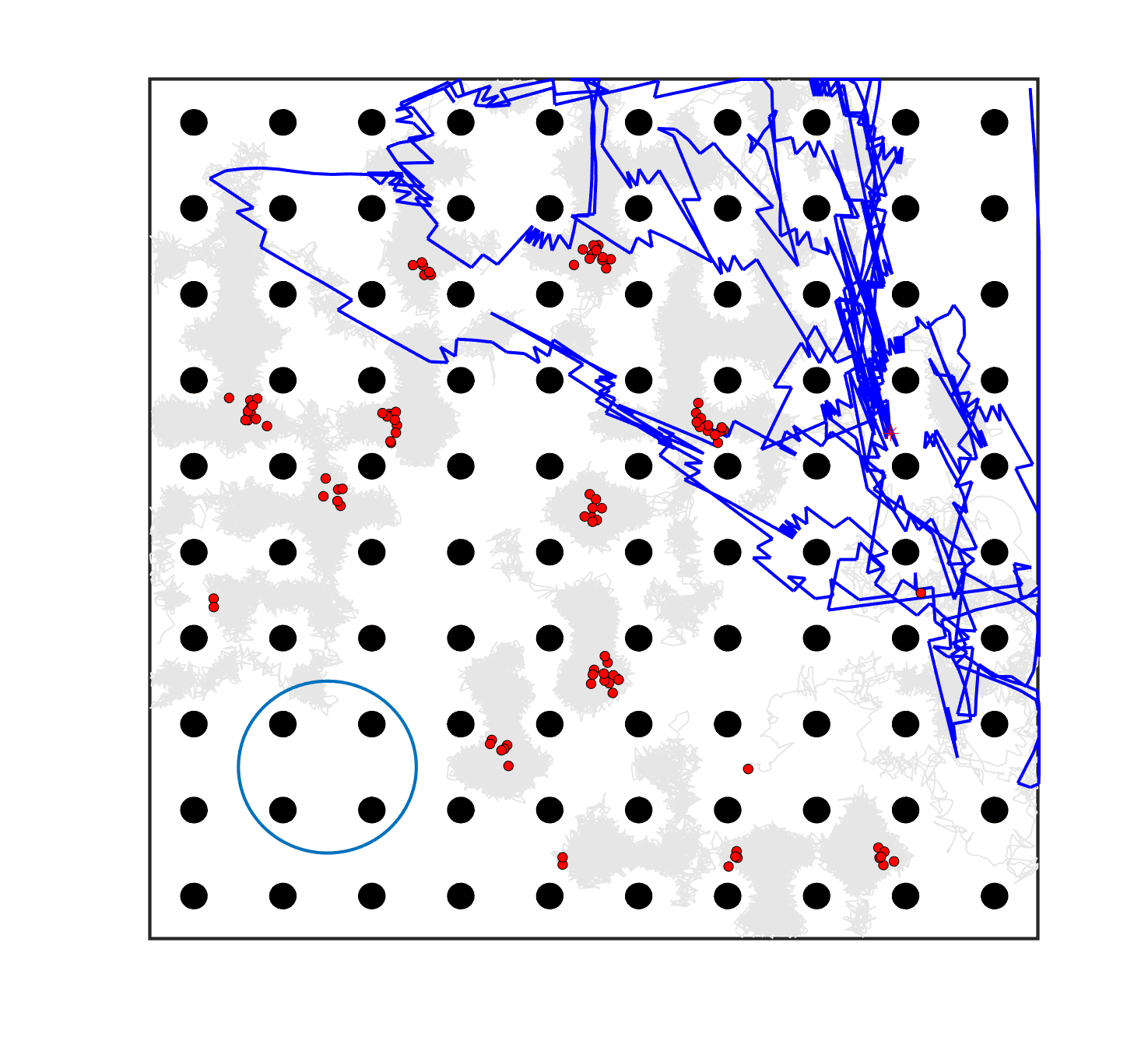}
    \label{fig_trajectory_case18_method1}
    }
    \subfloat[Case18\_trajectories\_Method2]{
    \includegraphics[width=0.23\textwidth,height=3.2cm]{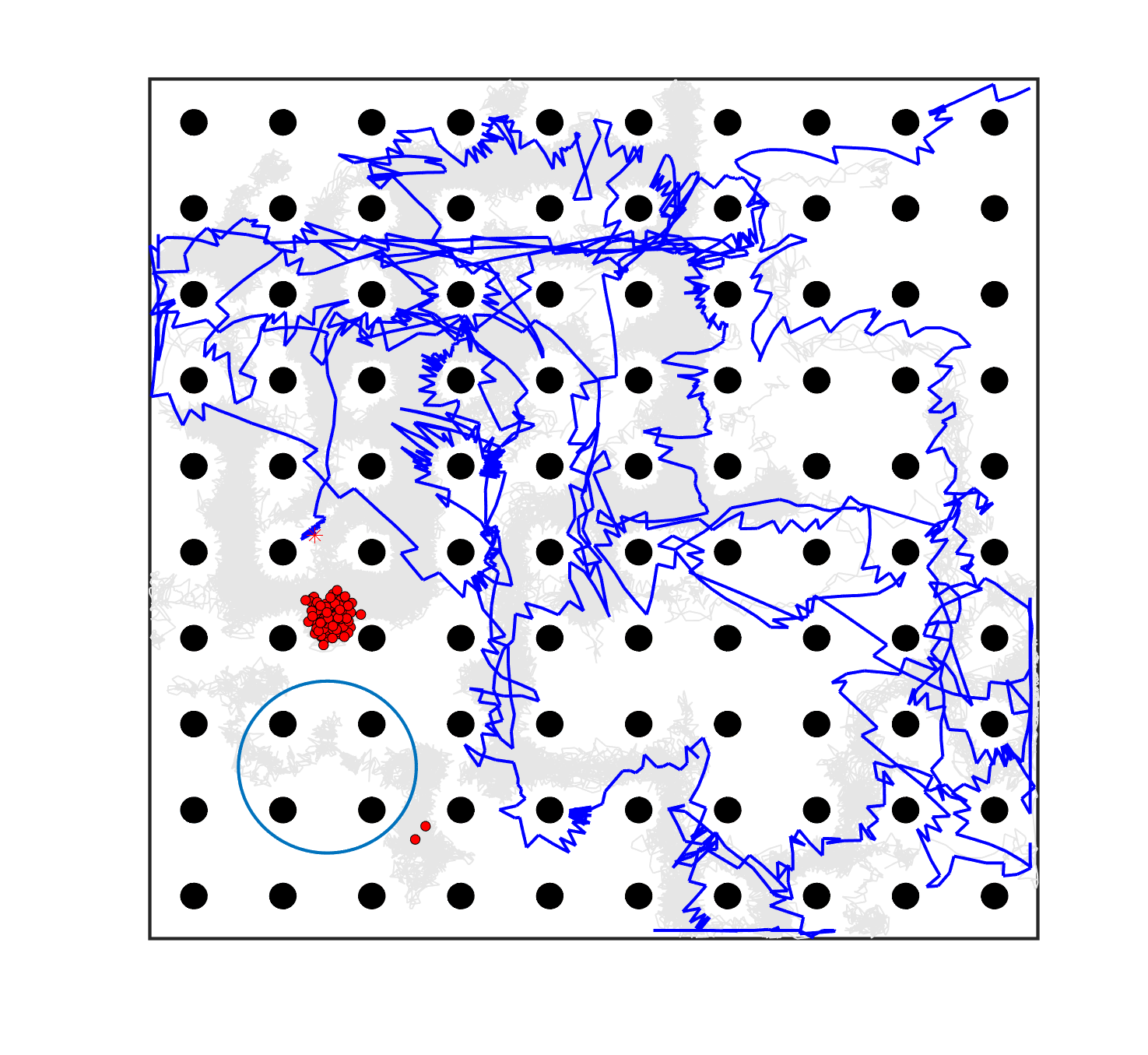}
    \label{fig_trajectory_case18_method2}
    }
    \subfloat[Case18\_trajectories\_Method3]{
    \includegraphics[width=0.23\textwidth,height=3.2cm]{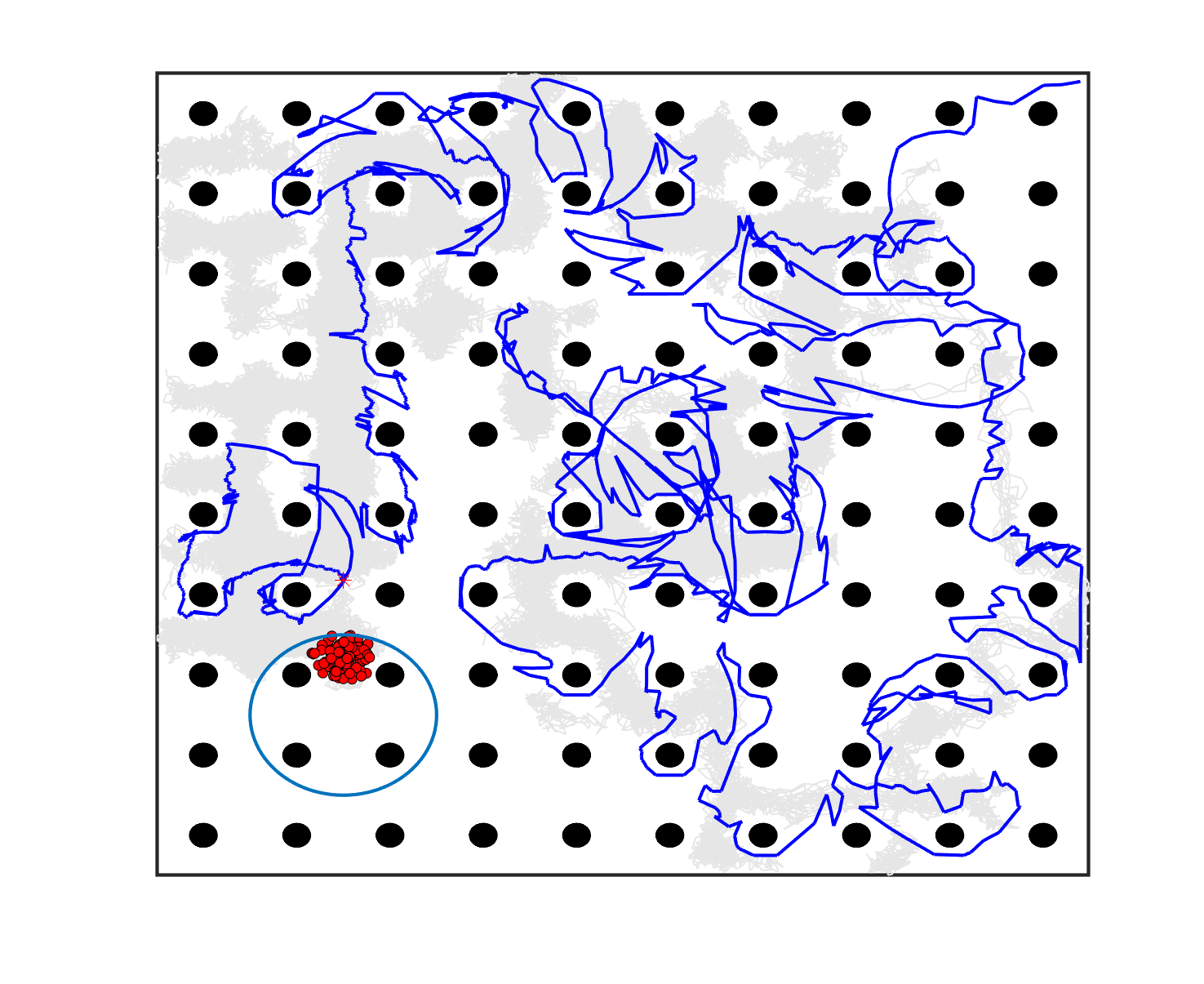}
    \label{fig_trajectory_case18_method3}
    }
     
    \caption{The visualised planning results and generated trajectories of Case 3, 13 and 18 with single-sheepdog}
    \label{fig_VisualisedShepherdingResults_single}
 \end{figure*}

The comparative results of shepherding methods using single-sheepdog are presented in Table~\ref{tab_results_single_sheepdog}. The planning results and generated trajectories by each method during shepherding for three representative cases (one for a group) are visualised in Fig.~\ref{fig_VisualisedShepherdingResults_single}. The blue lines in Fig.~\ref{fig_singlesheepdog_planningresults_case3},~\ref{fig_singlesheepdog_planningresults_case13},~\ref{fig_singlesheepdog_planningresults_case18} represent the planning results. In other figures, the grey lines represent the sheep trajectories, and the blue lines represent the sheepdog trajectories. 

\par As we can see from Table~\ref{tab_results_single_sheepdog}, the proposed planning-assisted swarm shepherding method performed the best overall among the three methods in almost all cases. In terms of the SR, with the increase of shepherding complexity, it becomes increasingly untenable for reactive shepherding to successfully complete the mission within the limited number of time steps, and the SR drops from 1 to 0. While reactive shepherding only obtained 100\% SR on 3 cases of Group 1 (Cases 1, 5 and 6) and failed in 10 cases, task planning-assisted shepherding achieved 100\% SR on 6 cases (Cases 1-5 of Group 1 and 7 of Group 2 ), which include most obstacle-free cases and a few obstacle-contained cases.  But task planning-assisted shepherding failed in 7 cases (Cases 10, 12-13, 17-20) and had low SR (less than 50\%) on 4 cases (Cases 8, 11, 15-16). This indicated that task planning, which divides the sheep swarm and determines the optimal pushing sequence of sub-swarms, could significantly increase the SR of shepherding in the environment without obstacles. It could also address some relatively simple shepherding missions in the environment with obstacles, but is unable to deal with the complex situations with cluttered obstacles and a large sheep swarm size. On the contrary, the proposed planning-assisted shepherding with both task planning and path planning integrated succeeded in all cases of Group 1 and more than half of Group 2 and 3, and achieved higher SR of Cases 8-13, 14-18 compared to Method 2. This demonstrates the effectiveness of path planning in improving the shepherding SR in obstacle-cluttered environments. Fig.~\ref{fig_trajectory_case13_method1}-~\ref{fig_trajectory_case13_method3} also validate that the sheepdog easily reached a deadlock without path planning~(Methods 1 and 2), while Method 3 was able to effectively avoid this situation.
 
\begin{table*}[!ht]
\begin{spacing}{0.9}
    \centering
    \caption{Comparative results of shepherding methods with single-sheepdog }   \label{tab_results_single_sheepdog}
    \resizebox{\textwidth}{!}{ 
    \begin{tabular}{|c|c|c|c|c|c|c|c|c|c|c|}
    \hline
     Case  & \multicolumn{3}{c|}{Method 1: Reactive shepherding }  & \multicolumn{3}{c|}{Method 2: Task planning-assisted shepherding}  & \multicolumn{3}{c|}{Method 3: Planning-assisted shepherding}  \\
     \hline
      & SR   & No. of steps &Path length  &  SR  & No. of steps &Path length & SR  & No. of steps &Path length  \\
    \hline
  	C1&1.00 & 219.45$\pm$46.38*& 427.43$\pm$87.90*& 1.00 & 143.35$\pm$12.09*& 257.31$\pm$24.19*& 1.00 & \textbf{131.00$\pm$12.14} & \textbf{214.70$\pm$20.69} \\
     	C2&0.80 & 540.94$\pm$95.78*& 1058.34$\pm$184.48*& 1.00 & 281.55$\pm$35.88& 494.55$\pm$67.65*& 1.00 & \textbf{272.80$\pm$24.27} & \textbf{457.75$\pm$43.86} \\
     	C3&0.95 & 898.53$\pm$186.88*& 1801.14$\pm$363.64*& 1.00 & 410.70$\pm$32.80& 604.55$\pm$41.13*& 1.00 & \textbf{396.05$\pm$18.85} & \textbf{510.17$\pm$38.77} \\
     	C4&0.75 & 968.07$\pm$145.03*& 1898.39$\pm$279.91*& 1.00 & 464.10$\pm$36.62& 633.36$\pm$37.50*& 1.00 & \textbf{456.75$\pm$24.57} & \textbf{563.06$\pm$35.75} \\
     	C5&1.00 & \textbf{1082.45$\pm$192.37} & 2132.50$\pm$362.61*& 1.00 & 1098.10$\pm$73.27& 645.07$\pm$26.22*& 1.00 & 1118.30$\pm$95.97& \textbf{560.12$\pm$30.96} \\
     	C6&1.00 & 1584.65$\pm$182.53*& 3083.65$\pm$373.67*& 0.95 & \textbf{1410.84$\pm$210.77} & 677.38$\pm$56.06*& 1.00 & 1434.05$\pm$255.51& \textbf{616.90$\pm$56.48} \\
     	\hline
     	C7&0.85 & 438.53$\pm$80.00*& 807.11$\pm$154.04*& 1.00 & 251.85$\pm$52.26& 455.44$\pm$100.63& 1.00 & \textbf{232.50$\pm$47.04} & \textbf{409.36$\pm$94.24} \\
     	C8&0.05 & 685.00$\pm$0.00*& 1082.91$\pm$0.00*& 0.05 & 695.00$\pm$0.00*& 614.87$\pm$0.00*& 1.00 & \textbf{293.70$\pm$22.20} & \textbf{490.47$\pm$37.76} \\
     	C9&0.00 &----&----& 0.80 & 544.75$\pm$37.62*& 2292.09$\pm$129.15*& 1.00 & \textbf{484.30$\pm$13.09} & \textbf{283.20$\pm$18.14} \\
     	C10&0.00 & ----& ----& 0.00 & ----& ----& 1.00 & \textbf{567.45$\pm$56.21} & \textbf{725.97$\pm$88.96} \\
     	C11&0.00 &----&----& 0.20 & 1212.00$\pm$68.00*& 1653.71$\pm$259.48*& 0.95 & \textbf{996.21$\pm$89.63} & \textbf{1401.94$\pm$166.63} \\
     	C12&0.00 &----&----& 0.00 &----&----& 0.75 & \textbf{1230.60$\pm$31.84} & \textbf{749.71$\pm$57.93} \\
     	C13&0.00 &----&----& 0.00 &----&----& 0.75 & \textbf{1216.40$\pm$43.09} & \textbf{773.66$\pm$45.49} \\
     	\hline
     	C14&0.10 & 1808.00$\pm$48.08*& 2652.05$\pm$63.06*& 0.80 & 1528.75$\pm$145.03*& 804.04$\pm$92.00*& 1.00 & \textbf{1441.85$\pm$109.28} & \textbf{714.76$\pm$58.81} \\
     	C15&0.10 & 1763.00$\pm$206.48*& 2701.55$\pm$285.92*& 0.50 & 1223.30$\pm$203.73*& 794.82$\pm$94.96*& 1.00 & \textbf{1129.40$\pm$96.65} & \textbf{722.48$\pm$64.41} \\	
     	C16&0.00 &----&----& 0.30 & 1837.83$\pm$311.60*& 1652.02$\pm$285.06*& 1.00 & \textbf{1191.75$\pm$137.19} & \textbf{955.65$\pm$143.42} \\
      C17&0.00 & ----& ----& 0.00 & ----& ----& 0.15 & \textbf{1984.00$\pm$52.12} & \textbf{967.57$\pm$138.94} \\
     	C18&0.00 &----&----& 0.00 &----&----& 0.45 & \textbf{2071.17$\pm$156.11} & \textbf{1858.70$\pm$235.51}* \\
     	C19&0.00 &----&----& 0.00 &----&----& 0.00 &----&----\\
     	C20&0.00 &----&----& 0.00 &----&----& 0.00 &----&----\\
    \hline
    \end{tabular}
    }
    \label{tab_results}
\end{spacing}
\end{table*}

\par The integration of task planning and path planning also significantly reduces the number of steps and the path length required to herd the sheep swarm to the goal in almost all cases. This proves that the planning-assisted shepherding method can save time and reduce the energy consumption of robots to complete the shepherding mission, which results in significant benefit in the real-world shepherding applications. In detail, we can observe from Table~\ref{tab_results_single_sheepdog} that, in most of the obstacle-free environments~(Cases 2-6) and the relatively simple obstacle-cluttered environments~(Case 7), the reduction of the number of steps is mainly caused by the employment of task planning as the difference between the number of steps obtained by task planning-assisted shepherding and planning assisted shepherding are not significant. However, when the shepherding complexity increases in Cases 8-18,  path planning plays an important role in further reducing the number of time steps so that the planning-assisted shepherding achieves the minimum number of steps. In terms of the path length, the task planning-assisted shepherding performed better than the reactive shepherding on all data-applicable cases, while the planning-assisted shepherding obtained the best performance as presented in Table~\ref{tab_results_single_sheepdog} and Fig.~\ref{fig_VisualisedShepherdingResults_single}. This demonstrates that both tasking planning and path planning are very effective in reducing the detours of the sheepdog. However, the single-sheepdog planning-assisted shepherding still has difficulty in addressing the most complex Cases 19 and 20 within the limited time and cannot guarantee 100\% SR for a few cases. 

\subsubsection{Planning-assisted shepherding with bi-sheepdog shepherding}
\begin{figure*} [!ht]
    \centering
    \subfloat[Case3\_planning results]{
    \includegraphics[width=0.23\textwidth,height=3.2cm]{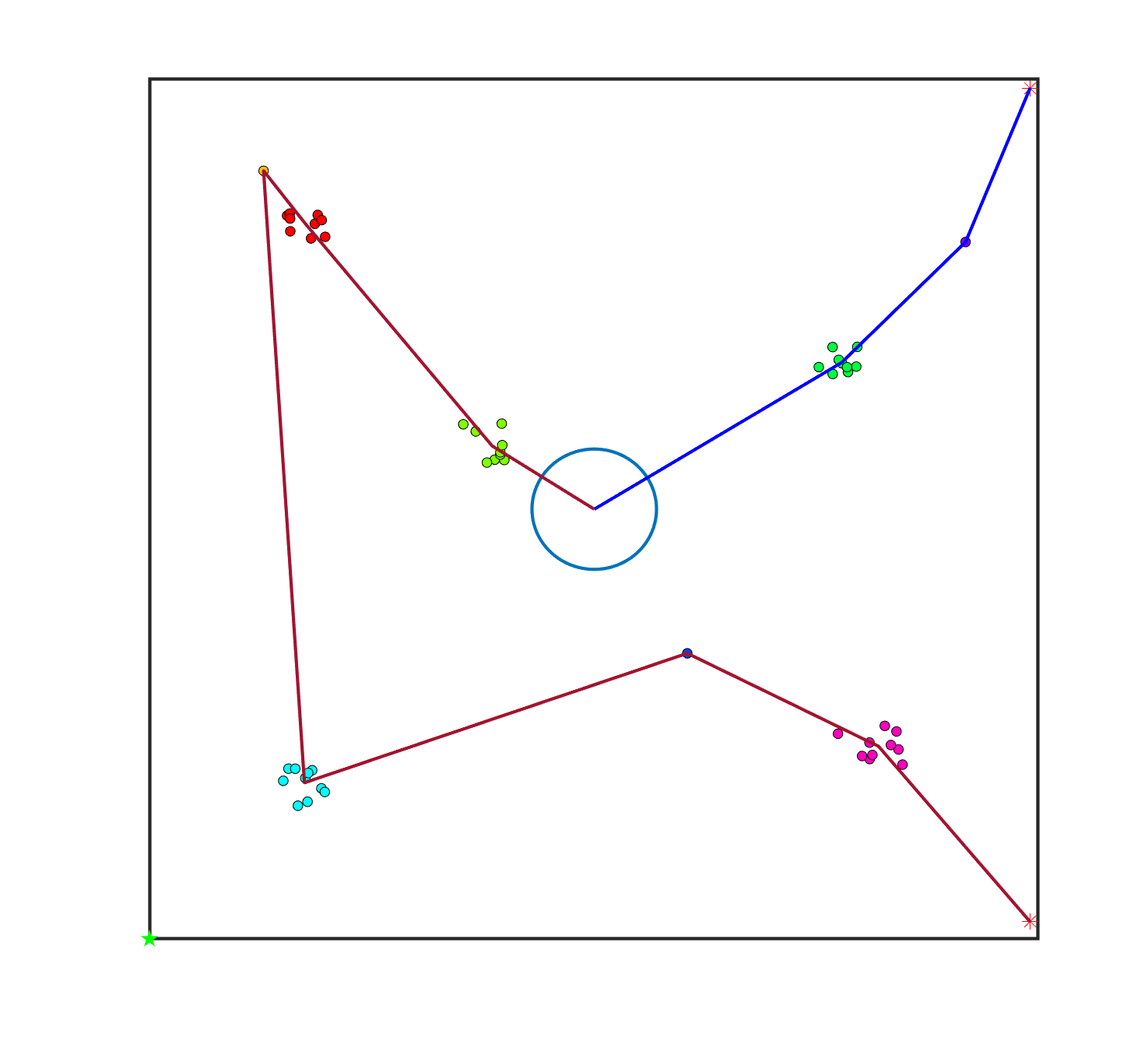}
    \label{fig_2sheepdog_planningresults_case8}
    }
    \subfloat[Case3\_trajectories\_Method1]{
    \includegraphics[width=0.23\textwidth,height=3.2cm]{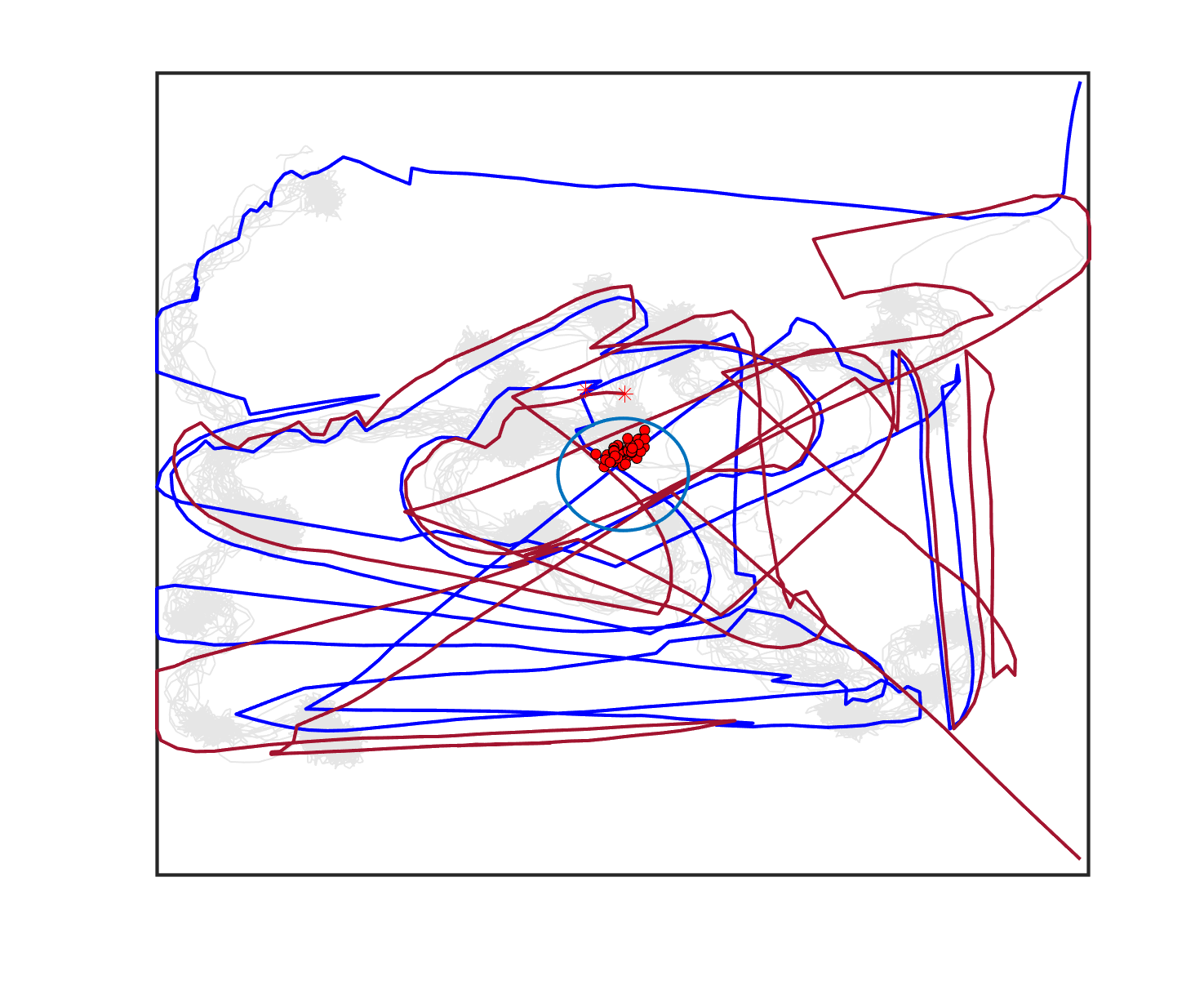}
    \label{fig_trajectory_case3_method4}
    }
    \subfloat[Case3\_trajectories\_Method2]{
    \includegraphics[width=0.23\textwidth,height=3.2cm]{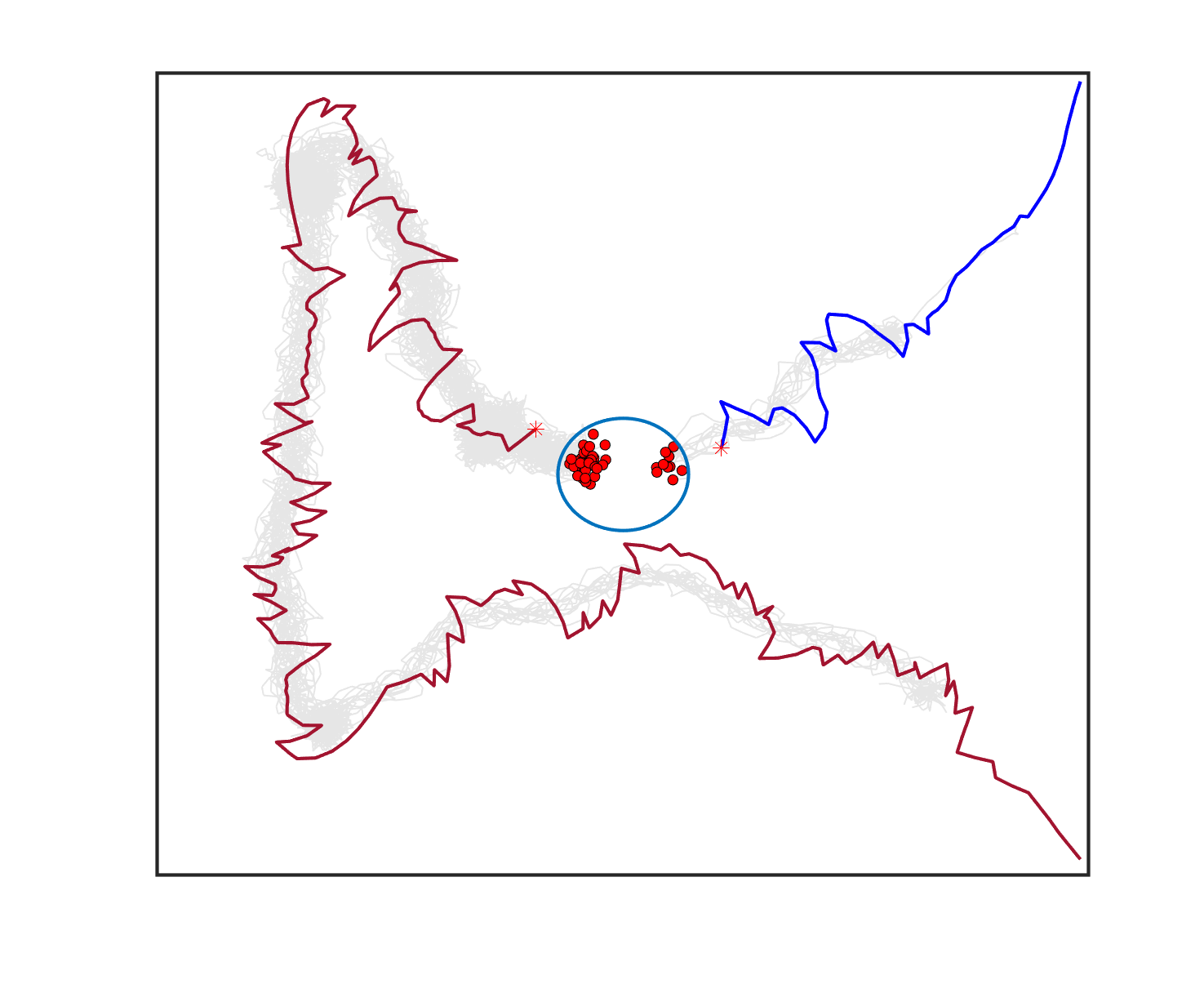}
    \label{fig_trajectory_case3_method5}
    }
    \subfloat[Case3\_trajectories\_Method3]{
    \includegraphics[width=0.23\textwidth,height=3.2cm]{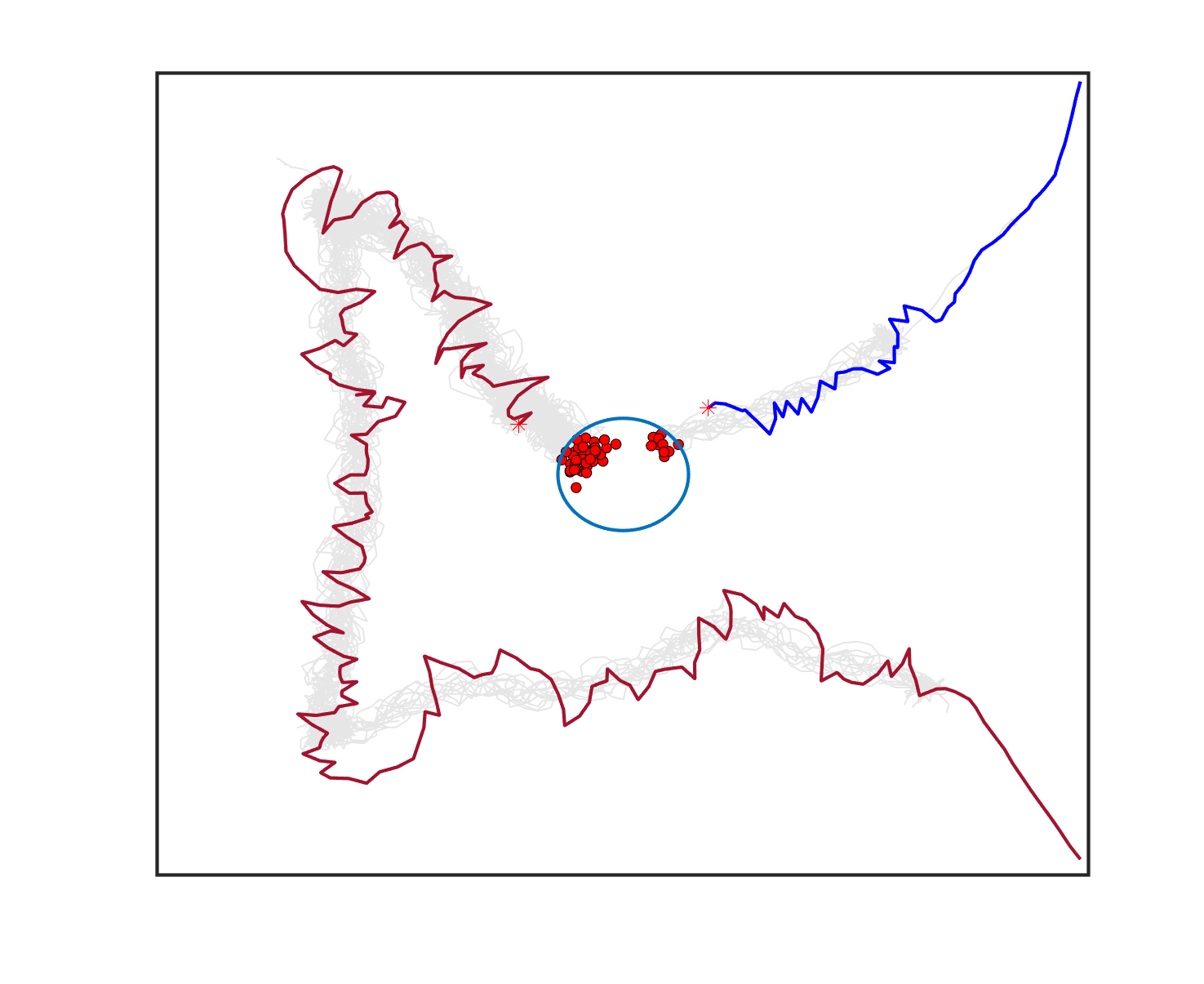}
    \label{fig_trajectory_case3_method6}
    }

     \subfloat[Case13\_planning results]{
    \includegraphics[width=0.23\textwidth,height=3.2cm]{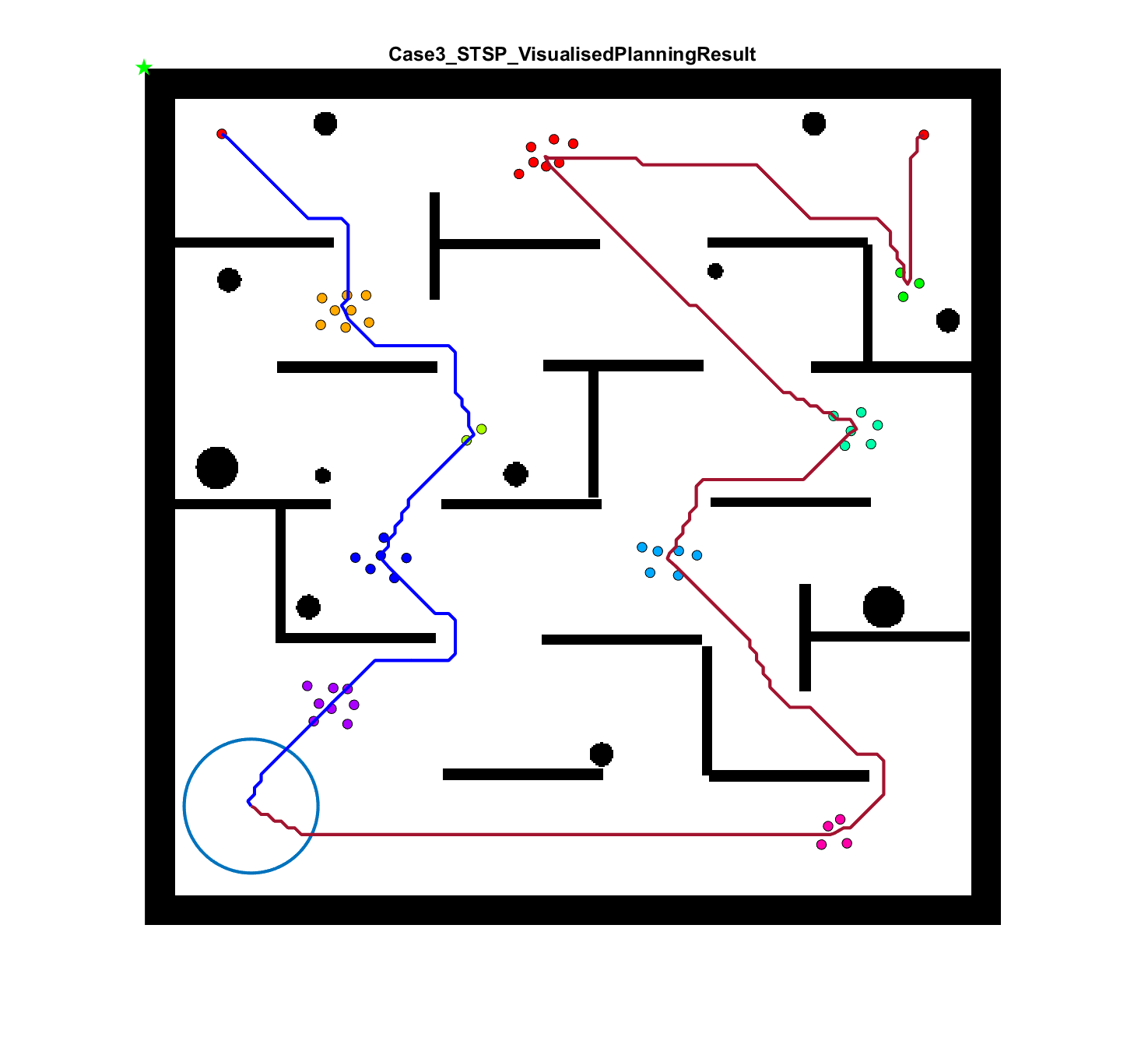}
    \label{fig_2sheepdog_planningresults_case13}
    }
    \subfloat[Case13\_trajectories\_Method1]{
    \includegraphics[width=0.23\textwidth,height=3.2cm]{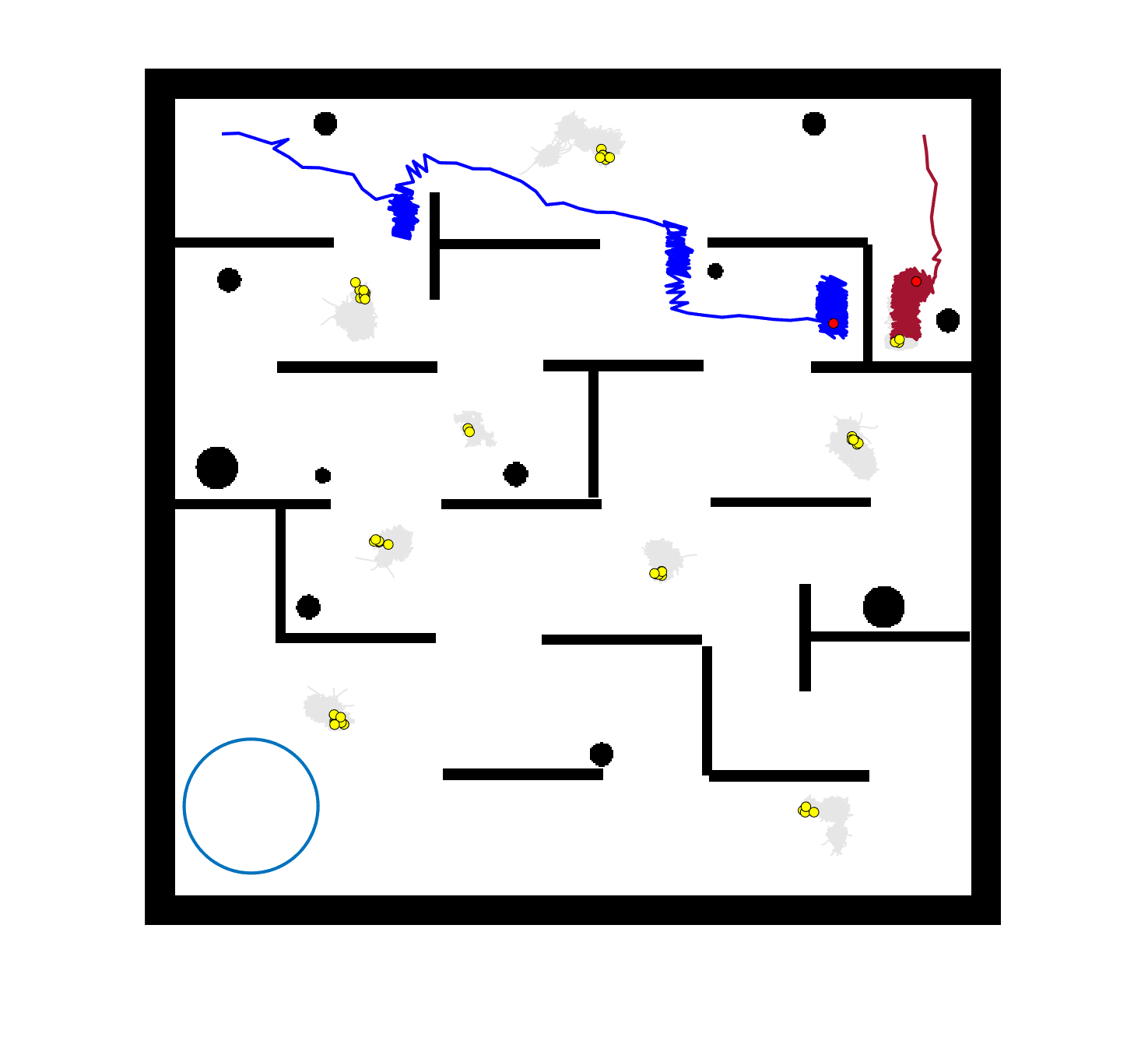}
    \label{fig_trajectory_case13_method4}
    }
    \subfloat[Case13\_trajectories\_Method2]{
    \includegraphics[width=0.23\textwidth,height=3.2cm]{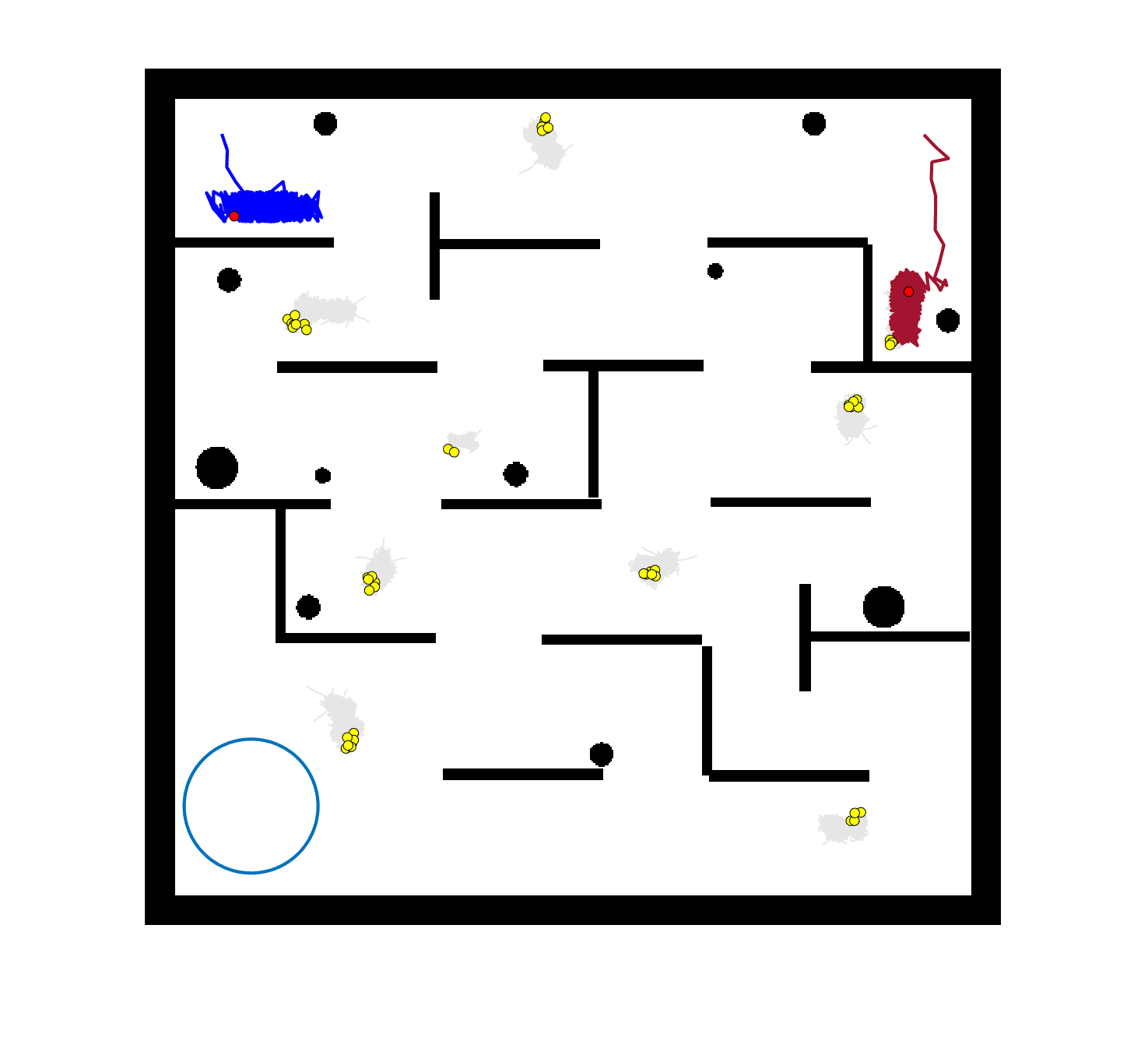}
    \label{fig_trajectory_case13_method5}
    }
    \subfloat[Case13\_trajectories\_Method3]{
    \includegraphics[width=0.23\textwidth,height=3.2cm]{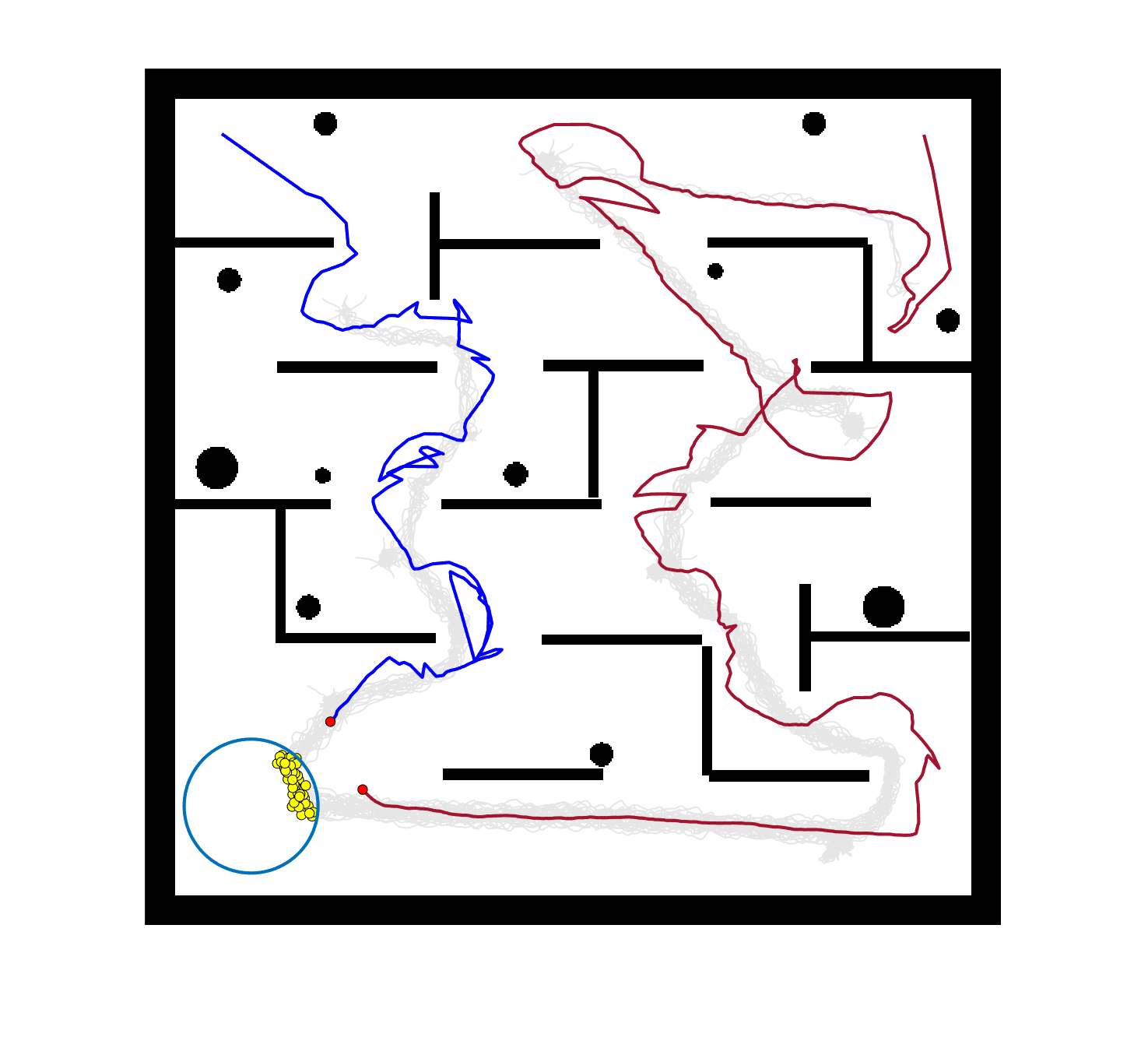}
    \label{fig_trajectory_case13_method6}
    }
    
    \subfloat[Case18\_planning results]{
    \includegraphics[width=0.23\textwidth,height=3.2cm]{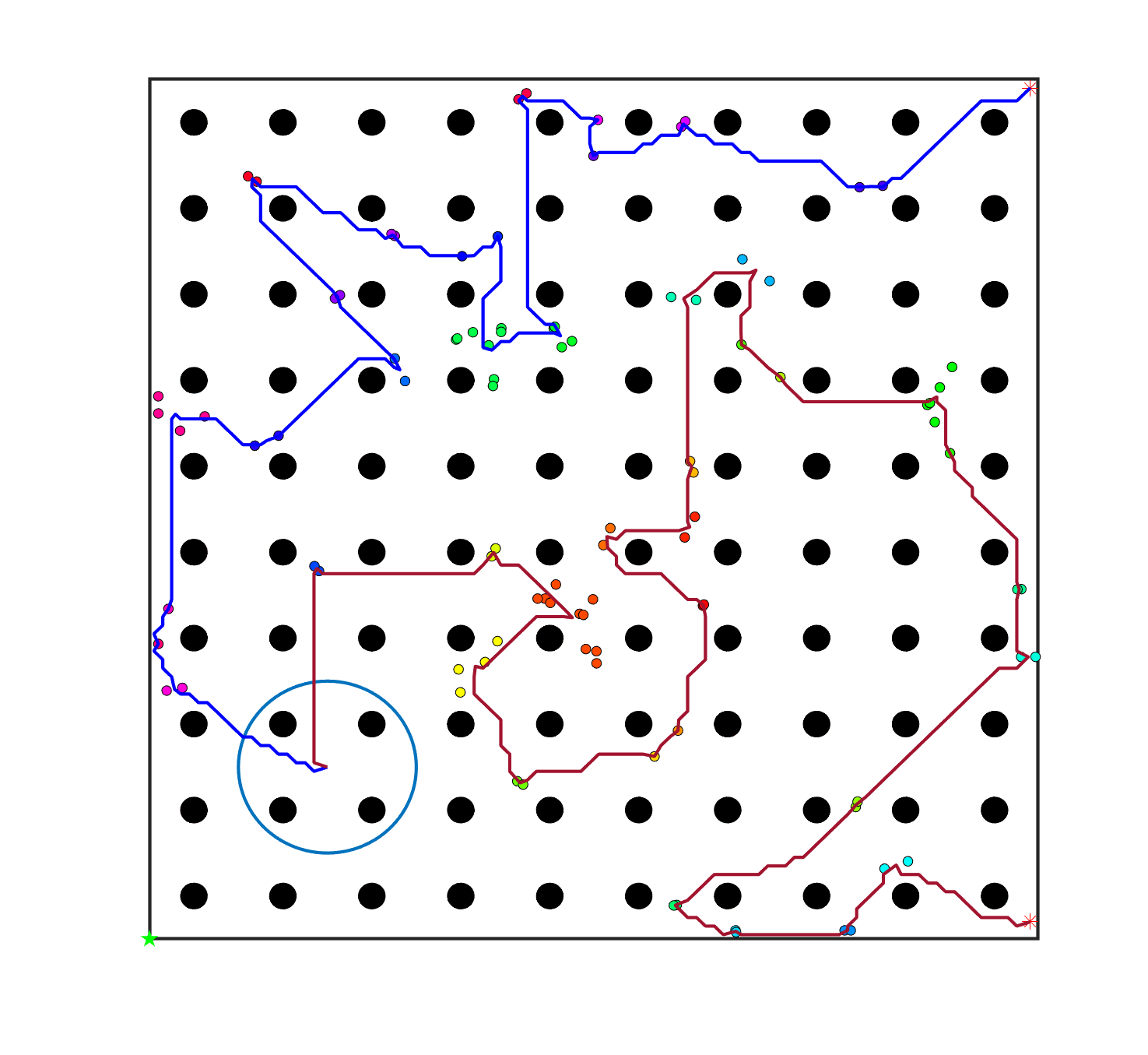}
    \label{fig_2sheepdog_planningresults_case18}
    }
    \subfloat[Case18\_trajectories\_Method1]{
    \includegraphics[width=0.23\textwidth,height=3.2cm]{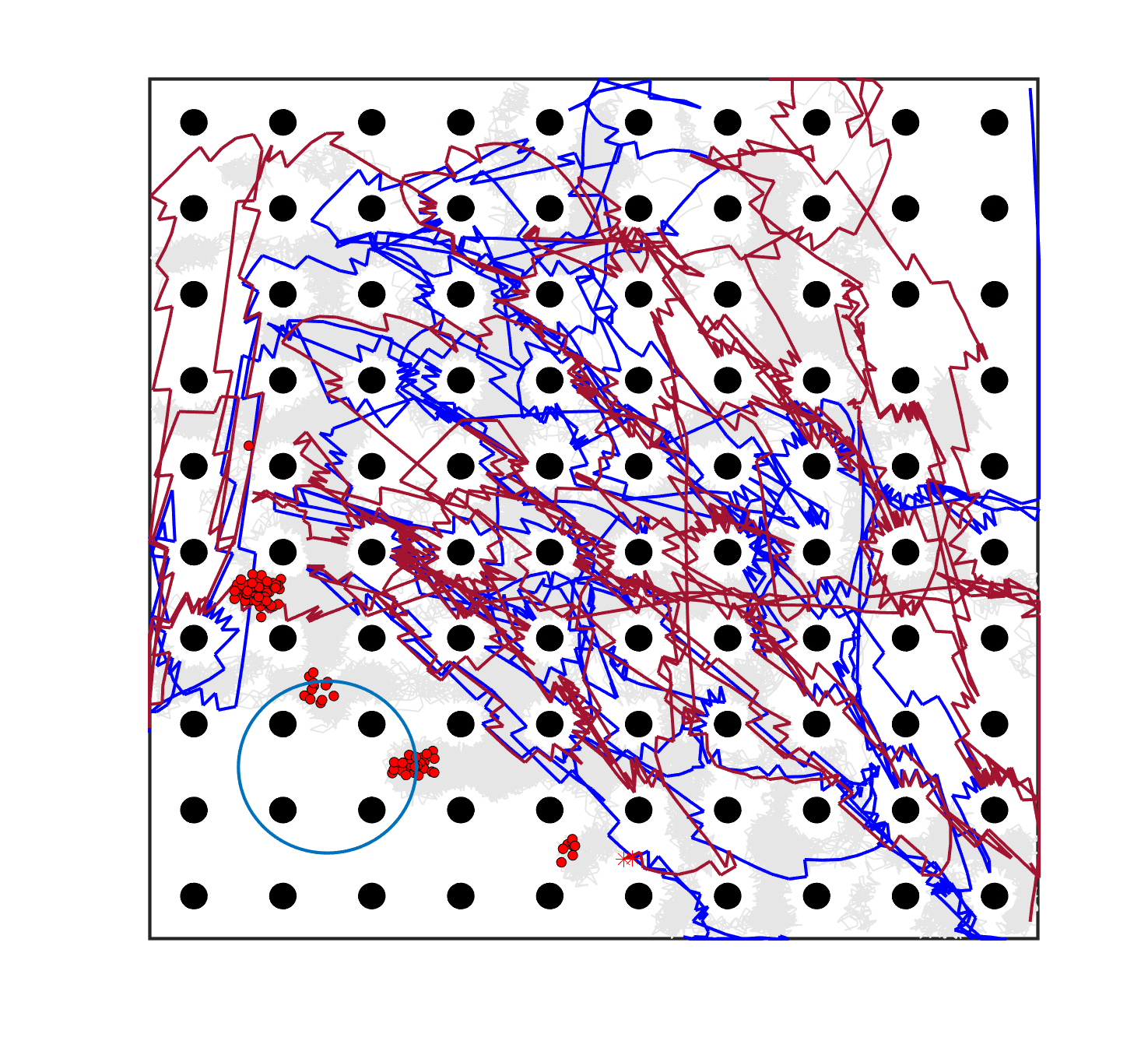}
    \label{fig_trajectory_case18_method4}
    }
    \subfloat[Case18\_trajectories\_Method2]{
    \includegraphics[width=0.23\textwidth,height=3.2cm]{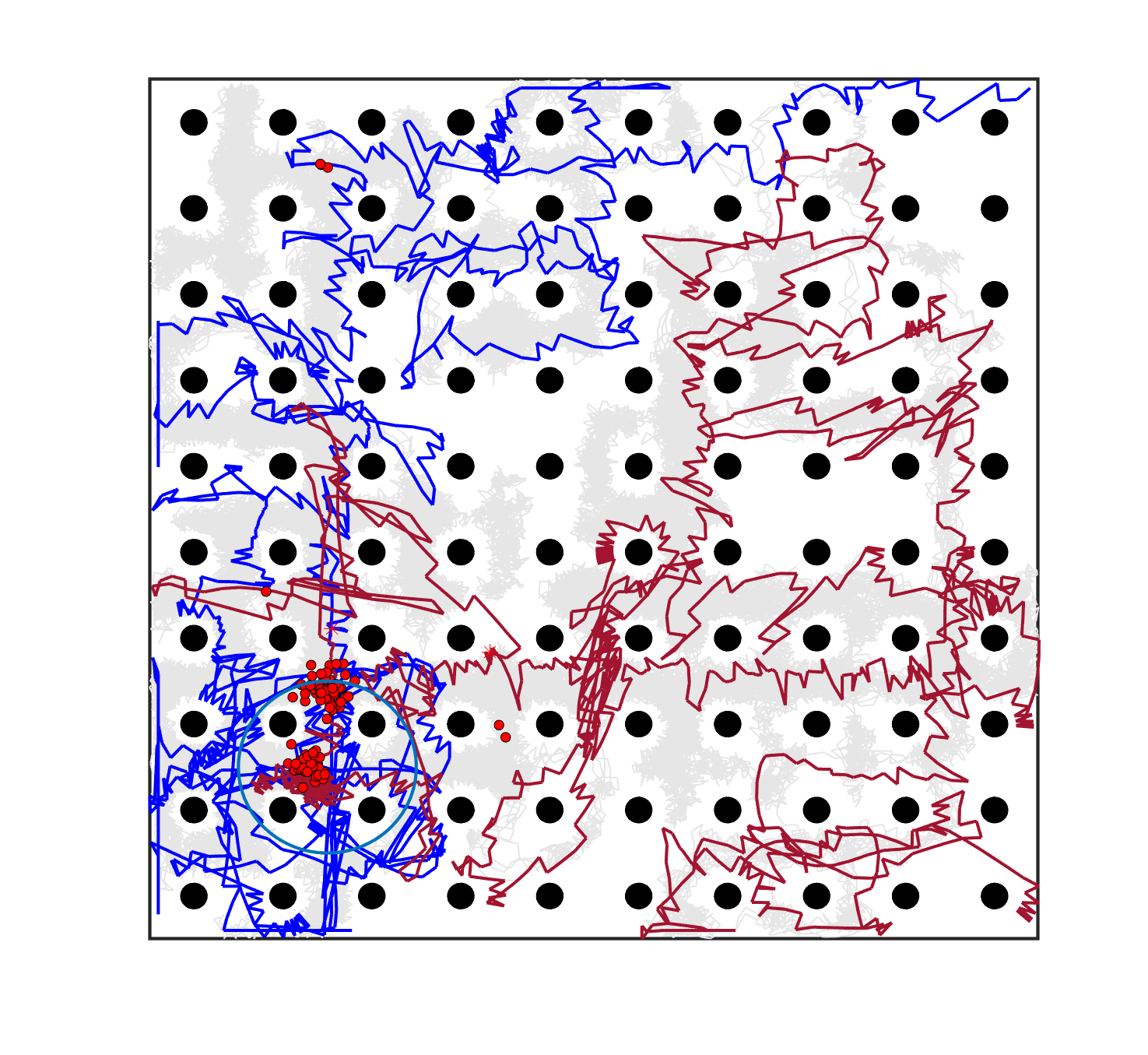}
    \label{fig_trajectory_case18_method5}
    }
    \subfloat[Case18\_trajectories\_Method3]{
    \includegraphics[width=0.23\textwidth,height=3.2cm]{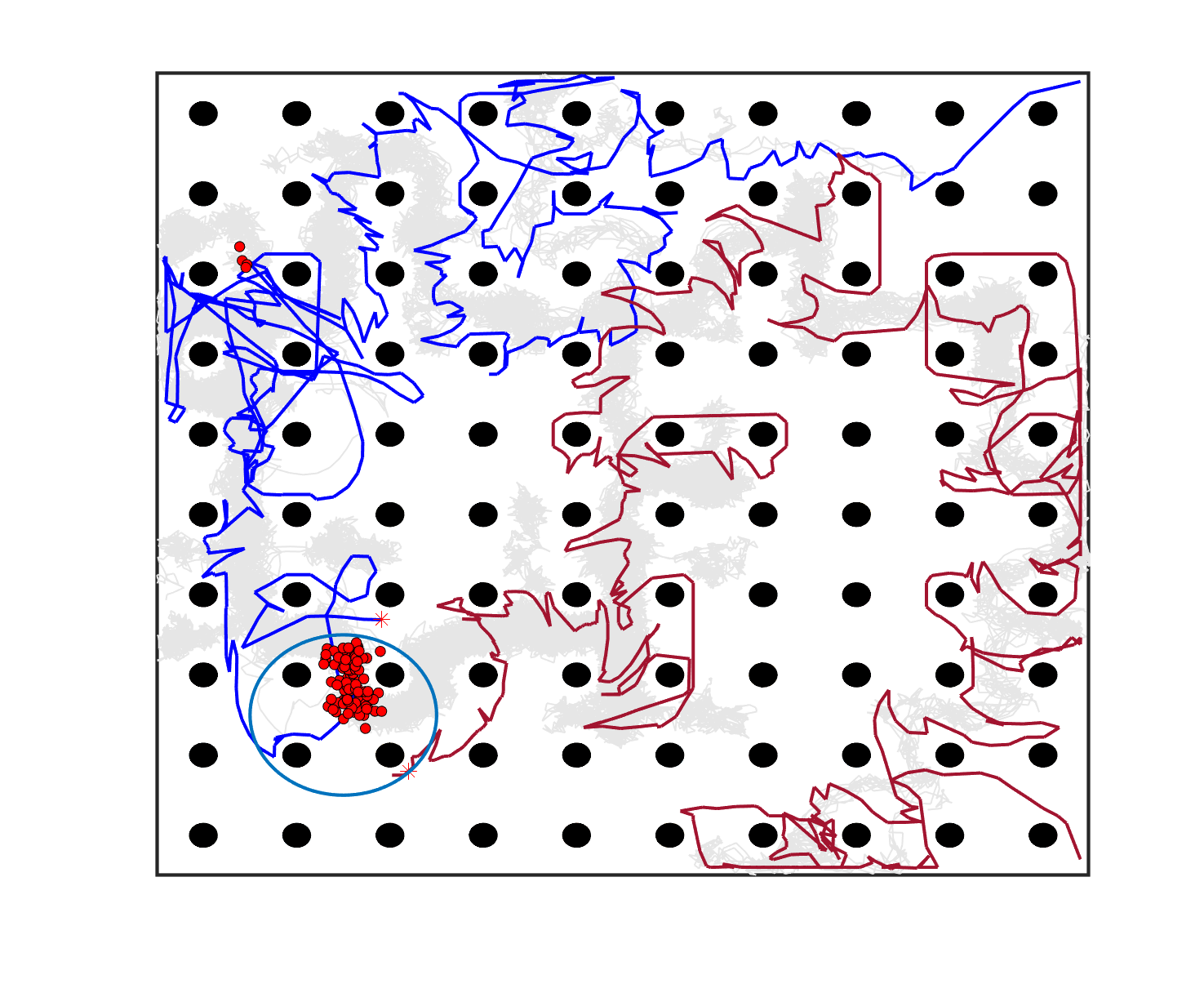}
    \label{fig_trajectory_case18_method6}
    }

    \caption{The visualised planning results and  trajectories of Case 3, 13 and 18 with bi-sheepdog}
    \label{fig_VisualisedShepherdingResults_bisheepdog}
 \end{figure*}

The failure of the planning-assisted shepherding with single-sheepdog in Cases 19 and 20 encourages the employment of multi-sheepdog in shepherding. We evaluated the performance of the three methods with two sheepdogs on the benchmark set, and Table~\ref{tab_results_bi_sheepdog} presents the results. When employing multiple agents, the mission completion time and the minimum cruising ability requirement are determined by the agent which consumes the most time and travels the longest distance, respectively. Therefore, the No. of steps and Path length presented in Table~\ref{tab_results_bi_sheepdog} are calculated based on the larger one between the values of the two sheepdogs.
The visualised planning results for the 3 representative cases are presented in Fig.~\ref{fig_2sheepdog_planningresults_case8},~\ref{fig_2sheepdog_planningresults_case13},~\ref{fig_2sheepdog_planningresults_case18} where the lines in different colours represent the optimal routes for sheepdogs. The trajectories of sheep (represented as lines in grey) and sheepdogs (represented as lines in blue and red) generated during the bio-sheepdog shepherding process based on different methods for these cases are visualised as other figures in Fig.~\ref{fig_VisualisedShepherdingResults_bisheepdog}.

\begin{table*}[!ht]
\begin{spacing}{0.9}
    \centering
    \caption{Comparative results of shepherding methods with bi-sheepdog }   \label{tab_results_bi_sheepdog}
    \resizebox{\textwidth}{!}{ 
    \begin{tabular}{|c|c|c|c|c|c|c|c|c|c|c|}
    \hline
     Case  & \multicolumn{3}{c|}{Method 1: Reactive shepherding }  & \multicolumn{3}{c|}{Method 2: Task planning-assisted shepherding}  & \multicolumn{3}{c|}{Method 3: Planning-assisted shepherding}  \\
     \hline
      & SR   & No. of steps &Path length  &  SR  & No. of steps &Path length & SR  & No. of steps &Path length  \\
    \hline
     C1&1.00 & 192.05$\pm$60.30*& 378.07$\pm$113.59*& 1.00 & 77.50$\pm$8.29& 153.00$\pm$13.99*& 1.00 & \textbf{76.85$\pm$7.94} & \textbf{132.49$\pm$15.92} \\
     	C2&0.95 & 412.84$\pm$136.45*& 811.45$\pm$264.00*& 1.00 & \textbf{132.45$\pm$7.34} & 250.01$\pm$13.10*& 1.00 & 133.10$\pm$7.37& \textbf{232.43$\pm$13.68} \\
     	C3&1.00 & 688.05$\pm$124.54*& 1392.42$\pm$242.39*& 1.00 & 289.40$\pm$20.38& 494.00$\pm$37.71*& 1.00 & \textbf{284.45$\pm$19.16} & \textbf{414.43$\pm$31.54} \\
     	C4&0.95 & 662.53$\pm$122.04*& 1322.85$\pm$243.08*& 1.00 & \textbf{195.65$\pm$10.52} & 374.66$\pm$20.84*& 1.00 & 198.30$\pm$8.89& \textbf{317.29$\pm$20.73} \\
     C5&1.00 & 768.90$\pm$158.46*& 1543.09$\pm$317.80*& 1.00 & 433.35$\pm$28.25& 424.59$\pm$22.60*& 1.00 & \textbf{431.90$\pm$49.90} & \textbf{365.22$\pm$28.34} \\
     	C6&1.00 & 1077.00$\pm$208.75*& 2128.58$\pm$420.37*& 1.00 & 263.65$\pm$15.46& 366.74$\pm$19.44*& 1.00 & \textbf{261.35$\pm$20.29} & \textbf{298.20$\pm$25.56} \\
     	\hline
     	C7&0.90 & 349.72$\pm$118.35*& 647.45$\pm$209.27*& 0.95 & 204.89$\pm$113.89& 357.89$\pm$146.12*& 1.00 & \textbf{157.40$\pm$36.19} & \textbf{282.79$\pm$67.60} \\
     	C8&0.85 & 463.94$\pm$113.64*& 854.12$\pm$194.77*& 1.00 & 174.30$\pm$15.00& 323.17$\pm$29.17*& 1.00 & \textbf{168.25$\pm$9.63} & \textbf{293.42$\pm$19.03} \\
     	C9&0.00 & ----& ----& 0.00 & ----& ----& 1.00 & \textbf{407.25$\pm$11.39} & \textbf{281.33$\pm$23.32} \\
     	C10&0.40 & 906.50$\pm$177.10*& 1556.23$\pm$267.23*& 0.00 & ----& ----& 1.00 & \textbf{249.65$\pm$11.50} & \textbf{383.25$\pm$24.44} \\
     		C11&0.00 & ----& ----& 0.70 & 755.50$\pm$267.97*& 1149.11$\pm$288.59*& 1.00 & \textbf{468.95$\pm$190.20} & \textbf{765.19$\pm$244.20} \\
     		C12&0.00 & ----& ----& 0.00 & ----& ----& 1.00 & \textbf{568.90$\pm$12.33} & \textbf{350.87$\pm$17.13} \\
     	C13&0.00 & ----& ----& 0.00 & ----& ----& 1.00 & \textbf{668.95$\pm$14.39} & \textbf{419.06$\pm$23.33} \\
     	\hline
     	C14&0.85 & 1138.35$\pm$336.01*& 1907.98$\pm$477.13*& 0.00 & ----& ----& 1.00 & \textbf{352.50$\pm$14.99} & \textbf{359.98$\pm$26.04} \\
     	C15&0.60 & 1656.17$\pm$394.33*& 2690.52$\pm$491.41*& 1.00 & 423.20$\pm$312.16*& 555.73$\pm$208.74*& 1.00 & \textbf{296.95$\pm$18.88} & \textbf{395.81$\pm$47.19} \\	
     	C16&0.10 & 1807.50$\pm$245.37*& 2469.13$\pm$34.49*& 0.75 & 1234.27$\pm$459.81*& 1111.22$\pm$453.89*& 1.00 & \textbf{605.45$\pm$209.93} & \textbf{564.11$\pm$116.75} \\
      C17&0.00 & ----& ----& 0.00 & ----& ----& 1.00 & \textbf{720.95$\pm$168.93} & \textbf{558.64$\pm$87.80} \\
     	C18&0.05 & 2077.00$\pm$0.00*& 2770.34$\pm$0.00*& 0.30 & 1421.00$\pm$600.48*& 1673.98$\pm$613.43*& 0.85 & \textbf{689.29$\pm$134.12} & \textbf{1028.32$\pm$195.86} \\
     	C19&0.10 & 1765.50$\pm$77.07*& 2892.37$\pm$71.84*& 0.80 & 1030.62$\pm$351.46*& 1436.14$\pm$471.77*& 1.00 & \textbf{527.25$\pm$48.09} & \textbf{618.39$\pm$156.38} \\
     	C20&0.00 & ----& ----& 0.00 & ----& ----& 1.00 & \textbf{1162.45$\pm$27.88} & \textbf{559.21$\pm$36.11} \\
        \hline
    \end{tabular}
    }
\end{spacing}
\end{table*}

\par We can observe from Table~\ref{tab_results_bi_sheepdog} and Fig.~\ref{fig_VisualisedShepherdingResults_bisheepdog} that the planning-assisted bi-sheepdog shepherding performed the best among these 3 methods and Method 2 performed better than Method 1, indicating the same findings from the above single-sheepdog shepherding: task planning and path planning can significantly improve the shepherding performance, especially for complex shepherding missions. Specifically, bi-sheepdog planning-assisted shepherding achieved 100\% SR on all cases except Case 18, while Method 1 and Method 2 with bi-sheepdog still completely failed in some cases.  Furthermore, compared to single-sheepdog shepherding, it is easy to find that the deployment of 2 sheepdogs, no matter based on which method, significantly improved the SR of addressing the complex shepherding tasks and reduced the number of time steps and the path length to complete the mission as shown in Table~\ref{tab_results_bi_sheepdog}. For example, bi-sheepdog planning-assisted shepherding increased the SR from 0 to 100\% on Cases 19 and 20 and obtained lower time steps and shorter path length compared to single-sheepdog planning-assisted shepherding in all cases. We can conclude that the deployment of multiple sheepdogs is an efficient way to reduce the completion time of the shepherding mission and the cruising ability requirement of the sheepdog. 

\par To further validate the effectiveness of bi-sheepdog shepherding, we also compare the total number of steps and the total path length of 2 sheepdogs obtained by planning-assisted shepherding to the  best results of single-sheepdog shepherding. The results are presented in Table~\ref{tab_reulsts_bisheepdog_single}, where the boldface denotes that the bi-sheepdog shepherding achieves better results than the single-sheepdog shepherding and `*' indicates the significant difference. It can be found that the planning-assisted bi-sheepdog shepherding still significantly outperformed the single-sheepdog shepherding in most of the cases in terms of the total values. This further demonstrates the efficiency of bi-sheepdog planning-assisted shepherding in terms of reducing the total time and energy consumption to complete the mission. 

\begin{table*}[!ht]
    \centering
    \caption{The comparison of bi-sheepdog shepherding to single sheepdog shepherding}  \label{tab_reulsts_bisheepdog_single}
    \resizebox{.35\textwidth}{!}{ 
    \begin{tabular}{|c|c|c|c|}
    \hline
     Case &  & \multicolumn{2}{c|}{The total} \\
     \hline
      & SR   & No. of steps &Path length   \\
    \hline
    C1&1.00 & \textbf{112.75$\pm$7.99}* & \textbf{195.62$\pm$15.68}* \\
     	C2&1.00 & \textbf{257.50$\pm$9.74} & \textbf{444.74$\pm$19.52} \\
     	C3&1.00 & \textbf{338.70$\pm$18.77} & \textbf{503.27$\pm$28.82} \\
     	C4&1.00 & \textbf{332.65$\pm$10.80}* & \textbf{531.97$\pm$22.48}* \\
     	C5&1.00 & \textbf{547.70$\pm$49.69} & \textbf{541.10$\pm$38.67} \\
     	C6&1.00 & \textbf{467.90$\pm$26.30}* & \textbf{548.44$\pm$35.06}* \\
     	\hline
     	C7&1.00 & \textbf{196.45$\pm$36.49}* & \textbf{347.06$\pm$67.72}* \\
     	C8&1.00 & \textbf{307.25$\pm$46.74}* & \textbf{537.17$\pm$85.49}* \\
     	C9&1.00 & 562.05$\pm$12.17* & 378.06$\pm$23.17* \\
     	C10&1.00 & \textbf{518.15$\pm$153.67}* & \textbf{840.70$\pm$283.54}* \\
     	C11&1.00 & \textbf{704.45$\pm$220.95}* & \textbf{1183.11$\pm$267.44}* \\
     	C12&1.00 & \textbf{866.65$\pm$15.21}* & \textbf{601.40$\pm$29.48}* \\
     	C13&1.00 & \textbf{897.10$\pm$16.47}* & \textbf{606.20$\pm$25.21}* \\
     	\hline
     	C14&1.00 & \textbf{572.30$\pm$69.50} & \textbf{704.61$\pm$178.78} \\
     	C15&1.00 & \textbf{503.55$\pm$38.45}* & \textbf{612.12$\pm$62.65}* \\
     		C16&1.00 & \textbf{687.70$\pm$85.22} & \textbf{900.81$\pm$161.92} \\
     	C16&1.00 & \textbf{827.45$\pm$242.41} & \textbf{947.27$\pm$209.32} \\
     	C17&0.85 & \textbf{1137.29$\pm$221.63} & 1728.28$\pm$234.88 \\
     	C19&1.00 & \textbf{836.45$\pm$122.29}* & \textbf{1039.61$\pm$218.38}* \\
     	C20&1.00 & \textbf{1628.65$\pm$39.02}* & \textbf{908.14$\pm$58.52}* \\
    \hline
    \multicolumn{4}{l}{ * represents the statistical significance} \\
    \end{tabular}
   }
\end{table*}

\section{Conclusion}  \label{Sec_conclusion}
This paper presents a planning-assisted context-sensitive swarm shepherding model and a hierarchical mission planning system for effectively herding a large flock of highly dispersed sheep to the destination in an environment with obstacles. In the proposed shepherding model, the sheep swarm is first grouped into some sheep sub-swarms, based on which the shepherding problem is transformed into a TSP to determine the optimal pushing sequence of sub-swarms by regarding each sub-swarm as a `city' to visit. Then the online path planning is integrated with a context-sensitive response model to find the optimal paths for the sheep sub-swarms to be pushed to the next `city' and the optimal paths for the sheepdogs to push the sheep sub-swarms. The hierarchical mission planning system is designed to solve the planning problems in the proposed shepherding model by combining a cohesion range-based method for grouping, ACO for TSP, and A*-PP for path planning.  

\par Experiments conducted on 20 shepherding cases consisting of three groups with different levels of complexity demonstrated the effectiveness of the planning-assisted swarm shepherding model in terms of increasing the success rate and reducing the time and energy consumption to complete the mission. The planning-assisted swarm shepherding model can also be extended for employing multiple sheepdogs, and experiments have also validated the performance improvements for bi-sheepdog shepherding. However, there remains more opportunities for extending this research. The employment of more than 2 sheepdogs for shepherding has not been studied in this work. Besides, when transforming the swarm shepherding problem into a TSP, the dynamics in shepherding are not considered. The travelling cost between each pair of cities does not consider the influence of the swarm size on the cost. Our future research will focus on how to model the multi-sheepdog swarm shepherding problem as a multiple dynamic TSP with a more accurate cost evaluation.  

\section{Funding}
This work is supported by a U.S. Office of Naval Research-Global (ONR-G) Grant and a Defence Science and Technology Group grant.

\section*{Declaration of Competing Interest}
None.

\bibliographystyle{elsarticle-num}
\bibliography{References}

\begin{thebibliography}{10}
\expandafter\ifx\csname url\endcsname\relax
  \def\url#1{\texttt{#1}}\fi
\expandafter\ifx\csname urlprefix\endcsname\relax\def\urlprefix{URL }\fi
\expandafter\ifx\csname href\endcsname\relax
  \def\href#1#2{#2} \def\path#1{#1}\fi

\bibitem{long2020comprehensive}
N.~K. Long, K.~Sammut, D.~Sgarioto, M.~Garratt, H.~A. Abbass, A comprehensive
  review of shepherding as a bio-inspired swarm-robotics guidance approach,
  IEEE Transactions on Emerging Topics in Computational Intelligence 4~(4)
  (2020) 523--537.

\bibitem{lien2009interactive}
J.-M. Lien, E.~Pratt, Interactive planning for shepherd motion., in: AAAI
  Spring Symposium: Agents that Learn from Human Teachers, 2009, pp. 95--102.

\bibitem{evered2014investigation}
M.~Evered, P.~Burling, M.~Trotter, et~al., An investigation of predator
  response in robotic herding of sheep, International Proceedings of Chemical,
  Biological and Environmental Engineering 63 (2014) 49--54.

\bibitem{strombom2018robot}
D.~Str{\"o}mbom, A.~J. King, Robot collection and transport of objects: A
  biomimetic process, Frontiers in Robotics and AI (2018) 48.

\bibitem{bat2017shepherding}
B.~Bat-Erdene, O.-E. Mandakh, Shepherding algorithm of multi-mobile robot
  system, in: 2017 First IEEE International Conference on Robotic Computing
  (IRC), IEEE, 2017, pp. 358--361.

\bibitem{paranjape2018robotic}
A.~A. Paranjape, S.-J. Chung, K.~Kim, D.~H. Shim, Robotic herding of a flock of
  birds using an unmanned aerial vehicle, IEEE Transactions on Robotics 34~(4)
  (2018) 901--915.

\bibitem{strombom2014solving}
D.~Str{\"o}mbom, R.~P. Mann, A.~M. Wilson, S.~Hailes, A.~J. Morton, D.~J.
  Sumpter, A.~J. King, Solving the shepherding problem: heuristics for herding
  autonomous, interacting agents, Journal of the royal society interface
  11~(100) (2014) 20140719.

\bibitem{singh2019modulation}
H.~Singh, B.~Campbell, S.~Elsayed, A.~Perry, R.~Hunjet, H.~Abbass, Modulation
  of force vectors for effective shepherding of a swarm: A bi-objective
  approach, in: 2019 IEEE Congress on Evolutionary Computation (CEC), IEEE,
  2019, pp. 2941--2948.

\bibitem{nguyen2020perceptron}
T.~Nguyen, J.~Liu, H.~Nguyen, K.~Kasmarik, S.~Anavatti, M.~Garratt, H.~Abbass,
  Perceptron-learning for scalable and transparent dynamic formation in
  swarm-on-swarm shepherding, in: 2020 International Joint Conference on Neural
  Networks (IJCNN), IEEE, 2020, pp. 1--8.

\bibitem{zhi2021learning}
J.~Zhi, J.-M. Lien, Learning to herd agents amongst obstacles: Training robust
  shepherding behaviors using deep reinforcement learning, IEEE Robotics and
  Automation Letters 6~(2) (2021) 4163--4168.

\bibitem{chipade2021multiagent}
V.~S. Chipade, D.~Panagou, Multiagent planning and control for swarm herding in
  2-d obstacle environments under bounded inputs, IEEE Transactions on Robotics
  37~(6) (2021) 1956--1972.

\bibitem{song2021herding}
H.~Song, A.~Varava, O.~Kravchenko, D.~Kragic, M.~Y. Wang, F.~T. Pokorny,
  K.~Hang, Herding by caging: a formation-based motion planning framework for
  guiding mobile agents, Autonomous Robots 45~(5) (2021) 613--631.

\bibitem{el2020limits}
H.~El-Fiqi, B.~Campbell, S.~Elsayed, A.~Perry, H.~K. Singh, R.~Hunjet, H.~A.
  Abbass, The limits of reactive shepherding approaches for swarm guidance,
  IEEE Access 8 (2020) 214658--214671.

\bibitem{hussein2022autonomous}
A.~Hussein, E.~Petraki, S.~Elsawah, H.~A. Abbass, Autonomous swarm shepherding
  using curriculum-based reinforcement learning., in: AAMAS, 2022, pp.
  633--641.

\bibitem{lavalle2006planning}
S.~M. LaValle, Planning algorithms, Cambridge university press, 2006.

\bibitem{zhao2021path}
Z.~Zhao, M.~Jin, E.~Lu, S.~X. Yang, Path planning of arbitrary shaped mobile
  robots with safety consideration, IEEE Transactions on Intelligent
  Transportation Systems (2021).

\bibitem{muller2022motion}
J.~M{\"u}ller, J.~Strohbeck, M.~Herrmann, M.~Buchholz, Motion planning for
  connected automated vehicles at occluded intersections with infrastructure
  sensors, IEEE Transactions on Intelligent Transportation Systems (2022).

\bibitem{liu2021mission}
J.~Liu, S.~Anavatti, M.~Garratt, H.~A. Abbass, Mission planning for shepherding
  a swarm of uninhabited aerial vehicles, Shepherding UxVs for Human-Swarm
  Teaming: An Artificial Intelligence Approach to Unmanned X Vehicles (2021)
  87--114.

\bibitem{lenstra1975some}
J.~K. Lenstra, A.~R. Kan, Some simple applications of the travelling salesman
  problem, Journal of the Operational Research Society 26~(4) (1975) 717--733.

\bibitem{elsayed2020path}
S.~Elsayed, H.~Singh, E.~Debie, A.~Perry, B.~Campbell, R.~Hunjel, H.~Abbass,
  Path planning for shepherding a swarm in a cluttered environment using
  differential evolution, in: 2020 IEEE Symposium Series on Computational
  Intelligence (SSCI), IEEE, 2020, pp. 2194--2201.

\bibitem{stutzle2000max}
T.~St{\"u}tzle, H.~H. Hoos, Max--min ant system, Future generation computer
  systems 16~(8) (2000) 889--914.

\bibitem{reynolds1987flocks}
C.~W. Reynolds, Flocks, herds and schools: A distributed behavioral model, in:
  Proceedings of the 14th annual conference on Computer graphics and
  interactive techniques, 1987, pp. 25--34.

\bibitem{miki2007effective}
T.~Miki, T.~Nakamura, An effective rule based shepherding algorithm by using
  reactive forces between individuals, International Journal of
  InnovativeComputing, Information and Control 3~(4) (2007) 813--823.

\bibitem{fujioka2018effective}
K.~Fujioka, Effective herding in shepherding problem in {V}-formation control,
  Transactions of the Institute of Systems, Control and Information Engineers
  31~(1) (2018) 21--27.

\bibitem{harrison2010scalable}
J.~F. Harrison, C.~Vo, J.-M. Lien, Scalable and robust shepherding via
  deformable shapes, in: International Conference on Motion in Games, Springer,
  2010, pp. 218--229.

\bibitem{hu2020occlusion}
J.~Hu, A.~E. Turgut, T.~Krajn{\'\i}k, B.~Lennox, F.~Arvin, Occlusion-based
  coordination protocol design for autonomous robotic shepherding tasks, IEEE
  Transactions on Cognitive and Developmental Systems (2020).

\bibitem{go2016reinforcement}
C.~K. Go, B.~Lao, J.~Yoshimoto, K.~Ikeda, A reinforcement learning approach to
  the shepherding task using sarsa, in: 2016 International Joint Conference on
  Neural Networks (IJCNN), IEEE, 2016, pp. 3833--3836.

\bibitem{nguyen2020continuous}
H.~T. Nguyen, T.~D. Nguyen, V.~P. Tran, M.~Garratt, K.~Kasmarik, S.~Anavatti,
  M.~Barlow, H.~A. Abbass, Continuous deep hierarchical reinforcement learning
  for ground-air swarm shepherding, arXiv preprint arXiv:2004.11543 (2020).

\bibitem{berczi2022efficient}
K.~B{\'e}rczi, M.~Mnich, R.~Vincze, Efficient approximations for many-visits
  multiple traveling salesman problems, arXiv preprint arXiv:2201.02054 (2022).

\bibitem{ayari2019acd3gpso}
A.~Ayari, S.~Bouamama, Acd3gpso: automatic clustering-based algorithm for
  multi-robot task allocation using dynamic distributed double-guided particle
  swarm optimization, Assembly Automation 40~(2) (2019) 235--247.

\bibitem{xie2022multiregional}
J.~Xie, J.~Chen, Multiregional coverage path planning for multiple energy
  constrained uavs, IEEE Transactions on Intelligent Transportation Systems
  (2022).

\bibitem{baniasadi2020transformation}
P.~Baniasadi, M.~Foumani, K.~Smith-Miles, V.~Ejov, A transformation technique
  for the clustered generalized traveling salesman problem with applications to
  logistics, European Journal of Operational Research 285~(2) (2020) 444--457.

\bibitem{khoufi2019survey}
I.~Khoufi, A.~Laouiti, C.~Adjih, A survey of recent extended variants of the
  traveling salesman and vehicle routing problems for unmanned aerial vehicles,
  Drones 3~(3) (2019) 66.

\bibitem{xu2021precedence}
X.~Xu, J.~Li, M.~Zhou, X.~Yu, Precedence-constrained colored traveling salesman
  problem: An augmented variable neighborhood search approach, IEEE
  Transactions on Cybernetics (2021).

\bibitem{mavrovouniotis2016ant}
M.~Mavrovouniotis, F.~M. M{\"u}ller, S.~Yang, Ant colony optimization with
  local search for dynamic traveling salesman problems, IEEE transactions on
  cybernetics 47~(7) (2016) 1743--1756.

\bibitem{ali2020novel}
I.~M. Ali, D.~Essam, K.~Kasmarik, A novel design of differential evolution for
  solving discrete traveling salesman problems, Swarm and Evolutionary
  Computation 52 (2020) 100607.

\bibitem{dorigo1992optimization}
M.~Dorigo, Optimization, learning and natural algorithms, PhD Thesis,
  Politecnico di Milano (1992).

\bibitem{dorigo1997ant}
M.~Dorigo, L.~M. Gambardella, Ant colony system: a cooperative learning
  approach to the traveling salesman problem, IEEE Transactions on evolutionary
  computation 1~(1) (1997) 53--66.

\bibitem{xiang2021pairwise}
X.~Xiang, Y.~Tian, X.~Zhang, J.~Xiao, Y.~Jin, A pairwise proximity
  learning-based ant colony algorithm for dynamic vehicle routing problems,
  IEEE Transactions on Intelligent Transportation Systems 23~(6) (2021)
  5275--5286.

\bibitem{cheikhrouhou2021comprehensive}
O.~Cheikhrouhou, I.~Khoufi, A comprehensive survey on the multiple traveling
  salesman problem: Applications, approaches and taxonomy, Computer Science
  Review 40 (2021) 100369.

\bibitem{oberlin2010today}
P.~Oberlin, S.~Rathinam, S.~Darbha, Today's traveling salesman problem, IEEE
  robotics \& automation magazine 17~(4) (2010) 70--77.

\bibitem{roberge2012comparison}
V.~Roberge, M.~Tarbouchi, G.~Labont{\'e}, Comparison of parallel genetic
  algorithm and particle swarm optimization for real-time {UAV} path planning,
  IEEE Transactions on Industrial Informatics 9~(1) (2012) 132--141.

\bibitem{yu2018aco}
X.~Yu, W.-N. Chen, T.~Gu, H.~Yuan, H.~Zhang, J.~Zhang, {{ACO-A}*: Ant colony
  optimization plus {A}* for 3-{D} traveling in environments with dense
  obstacles}, IEEE Transactions on Evolutionary Computation 23~(4) (2018)
  617--631.

\bibitem{hart1968formal}
P.~E. Hart, N.~J. Nilsson, B.~Raphael, A formal basis for the heuristic
  determination of minimum cost paths, IEEE transactions on Systems Science and
  Cybernetics 4~(2) (1968) 100--107.

\bibitem{yang2011anytime}
K.~Yang, Anytime synchronized-biased-greedy rapidly-exploring random tree path
  planning in two dimensional complex environments, International Journal of
  Control, Automation and Systems 9~(4) (2011) 750--758.

\end{thebibliography}
\end{document}